\documentclass[twocolumn]{article}
\usepackage{preprint}
\usepackage{xcolor}
\usepackage{float}
\usepackage{cite}
\usepackage{amsmath,amssymb,amsfonts}
\usepackage{amsthm}
\usepackage{algorithm}
\usepackage{algorithmic}
\usepackage{graphicx}
\usepackage{textcomp}
\usepackage{url}
\usepackage{xspace}
\usepackage{mathtools}
\usepackage{tabularx}
\usepackage{booktabs}
\usepackage{multirow}
\usepackage{enumitem}
\usepackage{bm}
\usepackage{upgreek}
\usepackage{placeins}
\usepackage{stfloats}
\usepackage[hidelinks,bookmarks=false]{hyperref}
\usepackage{fontawesome5}
\usepackage{cleveref}

\def\BibTeX{{\rm B\kern-.05em{\sc i\kern-.025em b}\kern-.08em
T\kern-.1667em\lower.7ex\hbox{E}\kern-.125emX}}

\newcommand{\eps}{{\varepsilon}}										
\newcommand{\E}[2][{}]{\mathbb E_{#1}\left[#2\right]}					
\newcommand{\PP}[1]{\mathbb P\left[#1\right]}							
\newcommand{\ra}{\rightarrow}											
\newcommand{\la}{\leftarrow}											

\newcommand{\Z}{{\mathbb{Z}}}											
\newcommand{\R}{{\mathbb{R}}}											
\newcommand{\state}{s}													
\newcommand{\State}{S}													
\newcommand{\states}{\mathbb S}											
\newcommand{\action}{a}													
\newcommand{\Action}{A}													
\newcommand{\actions}{\mathbb A}										
\newcommand{\policy}{\pi}												
\newcommand{\policies}{\Pi}												
\newcommand{\transit}{p}												
\newcommand{\reward}{r}													
\newcommand{\Value}{v}													
\newcommand{\G}{\ensuremath{\mathbb{G}}}								
\newcommand{\KL}{\ensuremath{\mathcal{KL}}\xspace}						
\newcommand{\Kinf}{\ensuremath{\mathcal{K}_{\infty}}\xspace}			

\newtheorem{dfn}{Definition}
\newtheorem{theorem}{Theorem}
\newtheorem{remark}{Remark}

\newtheorem{lemma}{Lemma}
\newtheorem{proposition}{Proposition}

\newcommand{\goaldist}[1][{}]{d_{\G#1}}
\newcommand{\basepolicy}{\policy_{b}}
\newcommand{\fallbackpolicy}{\policy_{f}}
\newcommand{\calfwpolicy}{\policy_{w}}
\newcommand{\Valuebase}{\hat{\Value}^{\basepolicy}}
\newcommand{\calfwrepo}{{\footnotesize\url{https://github.com/aidagroup/calf-wrapper}}\xspace}
\newcommand{\whiteqedsymbol}{\(\square\)}
{%
  \par\noindent\qquad\textit{Proof#1:}
}%
{%
  \hfill\whiteqedsymbol\par
}

\definecolor{orcidlogocol}{rgb}{0.65098,0.80784,0.22353}
\definecolor{secondbestblue}{rgb}{0.12,0.25,0.45}
\definecolor{thirdbestorange}{rgb}{0.62,0.30,0.04}
\crefname{figure}{Fig.}{Figs.}
\Crefname{figure}{Fig.}{Figs.}
\DeclareRobustCommand{\orcidicon}[1]{%
  \textsuperscript{\href{https://orcid.org/#1}{\textcolor{orcidlogocol}{\faIcon[brands]{orcid}}}}%
}

\makeatletter
\let\not\n@tch@r
\def\bstctlcite{\@ifnextchar[{\@bstctlcite}{\@bstctlcite[@auxout]}}
\def\@bstctlcite[#1]#2{\@bsphack
  \@for\@citeb:=#2\do{%
    \edef\@citeb{\expandafter\@firstofone\@citeb}%
    \if@filesw\immediate\write\csname #1\endcsname{\string\citation{\@citeb}}\fi}%
  \@esphack}
\makeatother

\begin{document}
\bstctlcite{IEEEexample:BSTcontrol}
\twocolumn[
  \begin{@twocolumnfalse}

\title{A Universal Policy Wrapper With Guarantees}

\author{
  Georgiy Malaniya\,\orcidicon{0000-0002-2936-6271}\thanks{The first two authors contributed equally.}\\
  \small Center for Engineering Systems and Sciences,\\
  \small 30 Bolshoy Boulevard, bld. 1, Moscow, 121205, Russia;\\
  \small Central University,\\
  \small 7 Gasheka St., bld. 1, Moscow, 123056, Russia\\
  \small \texttt{g.malaniya@rcdei.ru}
  \And
  Anton Bolychev\,\orcidicon{0000-0001-6327-7478}\footnotemark[1]\\
  \small Center for Engineering Systems and Sciences,\\
  \small 30 Bolshoy Boulevard, bld. 1, Moscow, 121205, Russia\\
  \small \texttt{bolychev.anton@gmail.com}
  \AND
  Pavel Osinenko\,\orcidicon{0000-0002-6184-3293}\thanks{Corresponding author.}\\
  \small Central University, 7 Gasheka St., bld. 1, Moscow, 123056, Russia;\\
  \small Center for Engineering Systems and Sciences, 30 Bolshoy Boulevard, bld. 1, Moscow, 121205, Russia;\\
  \small Sirius University of Science and Technology, 1 Olimpiyskiy Avenue, Sirius, 354340, Russia\\
  \small \texttt{p.osinenko@gmail.com}
}

\date{}

\maketitle

\begin{abstract}
  Many control tasks require driving the system
    into a prescribed goal set and subsequently keeping it there.
    The considered setting assumes that a working controller is available
    that provably reaches and maintains the goal, albeit with
    suboptimal performance.
    This controller is referred to as the \emph{fallback policy}.
    Conversely, although state-of-the-art reinforcement-learning (RL)
    algorithms can discover policies with impressive empirical
    performance, the neural networks at their core act as opaque black
    boxes learned through data-driven exploration and, as such, provide
    no theoretical mechanism to guarantee that the system will ever
    reach---and stay within---the prescribed goal set.
    This article presents CALF-Wrapper, a method that combines the
    fallback policy with a pretrained RL base policy and uses the critic
    estimate associated with the base policy to select between them
    during deployment.
    The theoretical analysis establishes that CALF-Wrapper inherits the
    goal-reaching property of the fallback policy under the stated
    assumptions and provides probability bounds for the settling time
    and the maximum distance from the goal set over a prescribed time interval.
    The experimental study evaluates CALF-Wrapper in four simulated
    control environments of different complexity.
    Comparisons include the base policy, fallback policy, and
    constrained reinforcement-learning methods formulated for the same objective.
    Sensitivity and ablation studies provide direct hyperparameter
    recommendations.
    The experiments show how CALF-Wrapper can provide effective task
    performance while attaining the maximum possible observed
    goal-reaching rate.
    CALF-Wrapper requires no additional dynamics model, online
    learning, or constrained optimization during execution.
\end{abstract}


\vspace{0.35cm}

  \end{@twocolumnfalse}
]

\begingroup
\renewcommand{\thefootnote}{\fnsymbol{footnote}}
\footnotetext[1]{The first two authors contributed equally.}
\footnotetext[2]{Corresponding author.}
\endgroup
\begingroup
\renewcommand{\thefootnote}{}
\footnotetext{This work was supported by the Russian Science Foundation
  (RSF) under Grant 25-21-00872.
  This article has been accepted for publication in IEEE Access.}
\endgroup

\section{Introduction}
\label{sec1}

Reinforcement learning (RL) has proved to be a powerful
framework for solving control problems in complex, high-dimensional settings.
Most RL algorithms optimize control policies through direct
interaction with a system (often called an environment), relying
solely on reward signals observed at each step, without requiring
explicit knowledge of system dynamics.
The remarkable successes of RL span applications such as robotic
manipulation~\cite{Akkaya2019Solvingrubiks, Kumar2016Optimalcontrol},
board games like Go, chess, and
shogi~\cite{Silver2018generalreinfor}, and complex real-time video
games~\cite{Vinyals2019Grandmasterlev}.

State-of-the-art deep-RL algorithms—including Twin-Delayed Deep
Deterministic Policy Gradient (TD3) \cite{Fujimoto2018}, Soft
Actor-Critic (SAC) \cite{Haarnoja2019}, and Proximal Policy
Optimization (PPO) \cite{Schulman2017ProximalPolicy}—learn effective
control policies in practice, yet they do not establish that a
learned policy satisfies a user-specified objective.
To make this requirement precise, this work considers a prescribed
goal set: a task-dependent subset of the state space that the agent
is required to reach and remain in.
The geometry of this set is task-dependent—for an inverted pendulum
it corresponds to upright configurations within a narrow angular
tolerance; in algorithmic trading it might be the region where
cumulative profit is positive; and in autonomous driving it can be a
small neighborhood surrounding the intended GPS destination.
Throughout this work, a nonempty goal set is assumed to be specified
in advance.
A policy that reaches the prescribed goal set and remains there is
said to satisfy the goal-reaching property formalized in
Section~\ref{sec_problem}.

Standard reinforcement-learning methods such as PPO and TD3 can
discover high-performing policies.
However, these methods provide no formal mechanism for establishing
whether the learned policy reaches the prescribed goal.
This behavior can therefore generally be assessed only empirically.
These methods are also often difficult to tune.
Rather than attempting to establish this property for opaque learned
policies, the formulation assumes the existence of a predefined
(possibly suboptimal) controller that already satisfies the
goal-reaching property---e.g., a provably stable yet
energy-inefficient impedance controller for a robotic arm; a
conservative, capital-preserving rule-based trading strategy; or a
heuristic vehicle-routing routine that ensures delivery irrespective
of fuel costs.
In the remainder of the paper, this controller is referred to as the
\emph{fallback controller} or \emph{fallback policy}.
The focus is to \emph{enhance} this fallback policy while preserving
its goal-reaching property.
The origin of the fallback controller—classical control (e.g., proportional--derivative (PD) control,
linear--quadratic regulator (LQR), or model predictive control (MPC)), domain heuristics, or engineering practice—is immaterial
to the analysis and problem formulation.
\textit{The sole objective is to improve the performance of the
fallback policy without losing its formal goal-reaching property.}

To improve the performance of the fallback policy, the method
introduces a second ingredient: a performance-seeking policy learned
by reinforcement learning, referred to as the \emph{base policy}.
Trained with any standard RL algorithm—TD3, Soft Actor-Critic, PPO,
or similar—the base policy is explicitly optimized for high
cumulative reward and therefore often attains stronger empirical returns.
However, it is not known to satisfy the goal-reaching property for
the prescribed goal set.

The approach presented in this paper treats the fallback controller
as an anchor for correctness and the learned RL policy as a source of
performance gains.
The approach is referred to as the \emph{CALF-Wrapper}.
It is designed as a lightweight decision-making mechanism that, at
each step, decides whether to execute the learned \emph{base policy}
or revert to the \emph{fallback policy}.
The wrapper can be applied after training to a broad class of RL
policies for which a corresponding value estimate is available.
This estimate may be provided directly by a critic or obtained by
evaluating an action-value critic at the action selected by the base policy.

\subsection{Related Work}
\label{sec:related_work}
The approach in~\cite{Osinenko2024CriticLyapunov}, known as
\textit{Critic as a Lyapunov Function} (CALF), is an \textbf{online}
reinforcement-learning algorithm that establishes the goal-reaching
property by repeatedly solving constrained optimization subproblems.
Although CALF inspired the proposed method, CALF-Wrapper differs from
the original CALF algorithm in three principal respects:

\begin{itemize}
  \item \textbf{No online learning or per-step optimization.} CALF
    continuously updates its policy during execution by solving a
    constrained optimization problem at each step.
    In contrast, CALF-Wrapper is a \textbf{deployment-time wrapper},
    requiring \textbf{no additional learning} and \textbf{no per-step
    optimization}.

  \item \textbf{Plug-and-play deployment.} CALF’s fallback policy is
    \textbf{tightly integrated into the training loop}, making the
    overall algorithm inseparable from the learning process.
    The proposed wrapper can be attached to an already-trained policy
    \emph{without} modifying the training code or the underlying
    policy network, yielding immediate applicability across a wide
    range of domains.

  \item \textbf{Adjustable trade-off between performance and the
    goal-reaching property.} CALF’s control law is strictly deterministic.
    CALF-Wrapper introduces a probabilistic decision-making mechanism
    that provides an adjustable trade-off between performance and the
    goal-reaching property.

\end{itemize}

{The differences between the original CALF algorithm and CALF-Wrapper
  are summarized in Table~\ref{tab:calf_algorithm_comparison}.
}

\begin{table}[H]
  \centering
  \caption{Comparison of the original CALF algorithm and CALF-Wrapper.}
  \label{tab:calf_algorithm_comparison}
  \scriptsize \setlength{\tabcolsep}{2pt}
  \begin{tabularx}{\columnwidth}{@{}p{0.20\columnwidth}p{0.36\columnwidth}X@{}}
    \toprule Aspect & Original CALF~\cite{Osinenko2024CriticLyapunov}
    & CALF-Wrapper \textup{(Algorithm~\ref{alg_calfwrapper})}
    \\ \midrule[\heavyrulewidth] Operation & Online learning &
    Deployment after training \\ \midrule[0.2pt] Policy & Actions are
    obtained by optimizing the critic over the action space or are
    generated by an actor model & Fixed pretrained base policy
    \\ \midrule[0.2pt] Critic & The critic model is updated online
    through constrained optimization & The critic is loaded from the
    same saved training checkpoint as the base policy and remains
    fixed during deployment \\ \midrule[0.2pt] Online optimization &
    Critic-loss minimization subject to a decrease constraint and
    critic bounds & No critic-weight or action optimization
    \\ \midrule[0.2pt] Decision at each step & The constrained
    critic-update problem is solved.
    The action generated by the learning algorithm is applied if a
    feasible solution is found; otherwise, the fallback-policy action
    is applied.
    & The decision requires only one critic forward pass, one uniform
    random-number sample, and scalar comparisons.
    No optimization problem is solved.
    \\ \midrule[0.2pt] RL training & Modifies the critic update &
    Does not modify training \\ \bottomrule
  \end{tabularx}
\end{table}

\begin{table*}[!b]
  \centering
  \caption{Conceptual comparison with representative safe-RL and
  runtime-intervention approaches.}
  \label{tab:related_work_comparison}
  \scriptsize \setlength{\tabcolsep}{3pt}
  \begin{tabularx}{\textwidth}{@{}p{0.18\textwidth}p{0.18\textwidth}p{0.20\textwidth}p{0.20\textwidth}X@{}}
    \toprule Approach & Objective or property & Operation during
    execution & Principal requirements & Relation to policy training
    \\ \midrule Simplex and
    shielding~\cite{seto1998,alshiekh2018shielding,kim2025continuousshields}
    & Formal safety-specification satisfaction & Switches controller
    or rejects or replaces the proposed action & Runtime monitor and
    certified backup controller, or model or abstraction and safety
    specification & May be constructed separately or incorporated
    into training \\ \specialrule{0.2pt}{0.2ex}{0.2ex} Predictive
    safety
    filters~\cite{wabersich2021filter,lavanakul2024safetyfilters} &
    State and input constraint satisfaction & Modifies the proposed
    action through a model-based feasibility or optimization problem
    & Dynamics model, constraints, and feasible set or certificate &
    Can wrap a fixed policy and solves an online problem
    \\ \specialrule{0.2pt}{0.2ex}{0.2ex} Recovery and intervention
    methods~\cite{thananjeyan2021recovery,wagener2021sailr,banerjee2024dmps}
    & Constraint satisfaction or reduced violation probability &
    Replaces the proposed action with a recovery, backup, or planned
    action & Safety critic and recovery or backup policy, or dynamics
    model and planner & Requires additional learning, online
    planning, or both \\ \specialrule{0.2pt}{0.2ex}{0.2ex}
    Constrained safe
    RL~\cite{altman1999constrained,achiam2017cpo,tessler2019rcpo} &
    Expected return subject to expected cumulative cost limits &
    Executes the trained policy directly & Reward, constraint-cost
    signal, and cost limit & Includes cost constraints in policy
    training \\ \specialrule{0.2pt}{0.2ex}{0.2ex} PPO-Lagrangian and
    TD3-Lagrangian implemented and used as baselines in this
    article~\cite{tessler2019rcpo,Schulman2017ProximalPolicy,Fujimoto2018}
    & Expected return subject to a bound on failure probability under
    the environment-specific goal-reaching criterion & Executes the
    trained policy directly & Goal-set predicate, episode horizon,
    binary cost, and cost limit & Adds cost estimation and
    Lagrange-multiplier updates to PPO and TD3 training
    \\ \specialrule{0.2pt}{0.2ex}{0.2ex} CALF-Wrapper & Goal-reaching
    property inherited from the fallback policy & Selects the
    base-policy or fallback-policy action at each step & Base policy,
    its fixed value-function estimate, and fallback policy satisfying
    the goal-reaching property & Keeps all components fixed and
    solves no online optimization problem \\ \bottomrule
  \end{tabularx}
\end{table*}

Achieving these contributions required a substantial reorganization
of the theoretical results.
The inheritance of the goal-reaching property from the fallback
policy—though conceptually similar to CALF—was the least demanding
part, yet it still required several techniques to accommodate the
probabilistic decision mechanism.
In contrast, the analysis of the uniform properties of the
goal-reaching behavior of CALF-Wrapper required substantially more
extensive theoretical development.
The analysis is presented in this article in two alternative forms.
Section~\ref{sec:uniform_lp_bounds} develops a fundamentally new
theoretical analysis under weaker assumptions on the fallback policy
and establishes uniform properties of the goal-reaching behavior of
CALF-Wrapper.
Specifically, it provides a single finite bound on the expected
distance from the goal set that holds at every time and gives
probability bounds on when the trajectory enters and subsequently
remains in a prescribed neighborhood of the goal set.
Appendix~\ref{sec:uniform_hitting_time} provides an alternative
analysis under stronger assumptions and establishes correspondingly
stronger claims.

{Beyond CALF, related work includes runtime assurance, shielding,
  safety filters, and methods that use intervention or recovery policies.
  The Simplex architecture switches from an unverified advanced
  controller to a certified backup controller when a runtime monitor
  predicts loss of safety~\cite{seto1998}.
  Shielding similarly rejects or replaces actions that violate a
  formal safety
  specification~\cite{alshiekh2018shielding,jansen2020probabilistic};
  recent work extends this principle to realizable shields in
  continuous state and action spaces~\cite{kim2025continuousshields}.
  Predictive safety filters instead modify proposed actions through a
  model-based feasibility or optimization problem so that state and
  input constraints remain
  satisfied~\cite{wabersich2021filter,lavanakul2024safetyfilters}.
  These approaches primarily prevent violations of formal safety
  specifications or state and input constraints.
  CALF-Wrapper instead addresses whether the system reaches a
  prescribed goal set and remains there.
}

{Several related methods replace an action proposed by a
  performance-seeking policy when that action may lead to a
  constraint violation.
  Recovery RL uses a task policy for performance and a recovery
  policy for constraint satisfaction.
  A safety critic, initialized from offline examples of constraint
  violations, estimates the discounted probability of a future
  violation for the action proposed by the task policy.
  If the estimate exceeds a threshold, the recovery-policy action is
  applied instead~\cite{thananjeyan2021recovery}.
  SAILR compares each action proposed by the policy being trained
  with the action selected by a predefined backup policy that is
  expected to satisfy the constraints.
  If the proposed action has a sufficiently higher estimated future
  constraint cost, it is replaced by the backup-policy action.
  SAILR also penalizes such proposed actions in a surrogate
  unconstrained problem, which trains the policy to avoid actions
  that would require replacement by the backup policy~\cite{wagener2021sailr}.
  Dynamic Model Predictive Shielding uses local planning when the
  action proposed by the learned policy may lead to a safety violation.
  The planner selects an alternative action that preserves safety
  while also improving task progress~\cite{banerjee2024dmps}.

  These methods generally require additional learning, system models,
  online planning, or a formal safety specification.
  CALF-Wrapper instead operates after the base policy and critic have
  been trained.
  It uses the base policy and critic stored at a selected training
  checkpoint together with a fixed fallback policy.
  During deployment, CALF-Wrapper updates none of these components
  and requires neither a system model nor the solution of a planning
  or constrained optimization problem.
  CALF-Wrapper addresses a different objective from shields and safety filters.
  It does not detect or prevent violations of state or input constraints.
  The predefined fallback policy is used so that the closed-loop
  trajectory reaches the prescribed goal set and remains there under
  the stated assumptions, rather than to prevent constraint violations.
  Consequently, the methods discussed above do not provide direct
  baselines for CALF-Wrapper in their original formulations because
  applying them to the present problem would require reformulating
  their objectives and requirements.
}

{However, the constrained reinforcement-learning framework can be
  formulated for the goal-reaching objective considered in this
  article and therefore provides a suitable baseline for the
  experimental evaluation of CALF-Wrapper in
  \Cref{sec:experiments}.
  This methodology is widely used in safe reinforcement learning,
  where undesirable outcomes are represented by cost functions and a
  policy is trained to maximize the expected cumulative reward
  subject to limits on expected cumulative
  costs~\cite{altman1999constrained,achiam2017cpo,tessler2019rcpo,ray2019benchmarking,stooke2020responsive,ha2021learning}.
  For the present comparison, failure under the environment-specific
  goal-reaching criterion is represented by a binary cost.

  A standard way to solve this constrained training problem is to
  introduce a nonnegative Lagrange multiplier for the cost limit and
  update both the policy and the multiplier during
  training~\cite{tessler2019rcpo,stooke2020responsive}.
  In this article, the resulting Lagrangian variants of PPO and TD3
  are referred to as PPO-Lagrangian and TD3-Lagrangian, respectively.
  Their constrained problems and implementations are described in
  Appendix~\ref{sec:lagrangian_baseline_implementation}.
  Using these baselines retains the reinforcement-learning algorithm
  used in each environment while changing whether the
  requirement on reaching the prescribed goal set is addressed during
  constrained training or by applying CALF-Wrapper after training.

  To conclude this section, Table~\ref{tab:related_work_comparison}
  summarizes the comparison with the related methods discussed above.
}

\subsection{Key Contributions}
\begin{itemize}
  \item A deployment-time policy wrapper that can be applied to a
    broad class of trained RL policies for which a corresponding
    value estimate is available.
    This estimate may be provided directly by a critic or obtained by
    evaluating an action-value critic at the action selected by the base policy.
    CALF-Wrapper combines the base policy with a predefined fallback
    policy, keeps all components fixed during deployment, and
    requires neither online learning nor online constrained optimization.

  \item A theoretical analysis establishing that CALF-Wrapper
    inherits the goal-reaching property of the fallback policy under
    the stated assumptions.
    The analysis of the uniform properties of the goal-reaching
    behavior of CALF-Wrapper is presented in two alternative forms.
    Section~\ref{sec:uniform_lp_bounds} establishes these properties
    using a uniform $L^1$ bound on the rate at which the expected
    distance from the goal set decreases under the fallback policy.
    It provides a single finite bound on the expected distance from
    the goal set that holds at every time and gives probability bounds
    on when the trajectory enters and subsequently remains in a
    prescribed neighborhood of the goal set.
    Appendix~\ref{sec:uniform_hitting_time} provides an alternative
    analysis using a stronger uniform almost-sure bound on the rate at
    which the distance from the goal set decreases under the fallback
    policy and establishes correspondingly stronger claims.

  \item An open-source implementation \calfwrepo that can be
    integrated with existing RL frameworks without modifying the
    policy-training procedure (e.g.,~\cite{stable-baselines3,huang2022cleanrl}).

  \item An empirical evaluation in four simulated control
    environments using PPO and TD3 base policies.
    The evaluation includes comparisons with PPO-Lagrangian and
    TD3-Lagrangian, sensitivity to the critic threshold, robustness
    to additive critic noise, execution-latency measurements, and
    statistical comparisons.
\end{itemize}

\section{Problem statement}\label{sec_problem}

This section establishes the formal reinforcement-learning framework
and mathematical foundations necessary to rigorously define the
goal-reaching property.

\subsection{Markov Decision Process}

A Markov decision process (MDP) is defined as
\begin{equation}
  \label{eq:mdp_definition}
  \mathcal{M}
  \;=\;\bigl(
    \states,\;\actions,\;\transit_0,\;\transit,\;\reward
  \bigr),
\end{equation}

\noindent where $\states$ is the state space (a finite-dimensional
Banach space), $\actions$ is the action space (a compact topological
space), $\transit_0: \states \to \R$ is the initial state probability
density function, $\transit(\bullet \mid \state,\action)$ is the
transition probability density, i.e., the probability density of the
next state $\State' \sim \transit(\bullet \mid \state, \action)$
given the current state $\state \in \states$ and action $\action \in
\actions$, and $\reward(\state,\action) \in \mathbb{R}$ is the
immediate reward of taking action $\action \in \actions$ in state
$\state \in \states$.
For the theoretical results, an upper semicontinuous function
$\bar{p} : \states \times \actions \to \mathbb{R}_{\geq 0}$ is
assumed to exist such that, for any $\state \in \states$ and $\action
\in \actions$, the next state $\State'$ sampled from
$\transit(\bullet \mid \state, \action)$ satisfies $\PP{\|\State'\|
\leq \bar{p}(\state, \action)} = 1$, effectively bounding the
system's one-step transition magnitude.

\subsection{Policies and Value Functions}

Two types of policies are considered:

1) \textit{Stationary policies} $\Pi_{\mathrm{stat}}$.
For any stationary policy $\pi \in \Pi_{\mathrm{stat}}$, $\pi(\action
\mid \state)$ represents the probability density of selecting action
$\action$ in state $\state$.

2) \textit{Non-stationary policies} $\policies_{\mathrm{nstat}}$.
For any non-stationary policy $\pi \in \policies_{\mathrm{nstat}}$,
$\pi(\action \mid \state, t)$ represents the probability density of
selecting action $\action$ in state $\state$ at time $t$.
By definition, \(\Pi_{\mathrm{stat}} \subset \Pi_{\mathrm{nstat}}\),
since any stationary policy can be viewed as a special case of a
non-stationary policy that remains constant across time steps.

For a discount factor $\gamma \in (0,1]$ and a policy $\policy \in
\policies_{\mathrm{nstat}}$, the \emph{value function}
$\Value^\policy : \states \to \mathbb{R}$ is
\begin{equation*}
  \label{eq:value_function}
  \Value^\policy(\state)
  =
  \E{
    \sum_{t=0}^{\infty}
    \gamma^t\,
    \reward(\State_t,\Action_t)
    \Bigm|
    \State_0 = \state
  },
\end{equation*}
where $\Action_t \sim \policy(\bullet \mid \State_t, t)$, and
$\State_{t+1} \sim \transit(\bullet \mid \State_t,\Action_t)$ for $t \ge 0$.

The corresponding \emph{action-value function}
$Q^\policy:\states\times\actions\to\mathbb{R}$ is defined by
\begin{equation*}
  Q^\policy(\state,\action)
  =
  \E{
    \sum_{t=0}^{\infty}
    \gamma^t\,
    \reward(\State_t,\Action_t)
    \Bigm|
    \State_0=\state,\Action_0=\action
  }.
\end{equation*}
For a stationary policy $\policy\in\policies_{\mathrm{stat}}$, the
state-value and action-value functions satisfy
\begin{equation*}
  \Value^\policy(\state)=\mathbb{E}[Q^\policy(\state,\Action)],
  \quad
  \text{for }\Action\sim\policy(\bullet\mid\state).
\end{equation*}

The standard RL objective is to maximize the expected discounted return:
\begin{equation}
  \label{eq:rl_objective}
  \E{\Value^\policy(\State_0)} = \E{\sum_{t = 0}^{\infty}\gamma^t
  \reward(\State_t, \Action_t)} \ra \max_{\policy\in\policies_{\mathrm{nstat}}},
\end{equation}
where $\State_0 \sim \transit_0(\bullet)$, $\Action_t \sim
\policy(\bullet \mid \State_t, t)$, and $\State_{t+1} \sim
\transit(\bullet \mid \State_t,\Action_t)$ for $t \ge 0$.

For a given policy $\policy$ and initial state $\state_0 \in
\states$, the state at time $t$ is denoted by $\State_t^{\policy}(\state_0)$.

\subsection{Goal Set and Goal-Reaching Property}

While traditional RL focuses on reward maximization, many control
problems require driving the system to a specific target region.
This requirement is formalized as follows:

\begin{dfn}[Goal set]\label{dfn:goal_set}
  A \emph{goal set} $\G \subset \states$ is a compact set
  representing the target region that the system must eventually
  reach and remain in.
  Typically, $\G$ is chosen as a compact neighborhood of a desired
  operating point.
  Without loss of generality, the desired operating point is assumed
  to be the origin, with $0 \in \G$.
\end{dfn}

The goal set may be specified directly as a target region or may be
related to the reward function.
For instance, when the reward is maximal at a target state or target region and
decreases away from it, a neighborhood of the reward maximizer can be
used as the goal set.
\begin{remark}[Reward function and goal set]
  \label{rem:reward_goal_set}
  Throughout this paper, the reward function is assumed to be bounded
  above: $\sup_{\state\in\states,\,\action\in\actions}
  \reward(\state,\action)<+\infty$.
  Without this condition, the return in \eqref{eq:rl_objective} may
  be unbounded above, and the optimization problem may not have a
  finite objective scale.
  Working with rewards that are not bounded above therefore requires
  additional conditions that exclude returns that diverge to $+\infty$.
  In addition, the function
  $\state\mapsto\sup_{\action\in\actions}\reward(\state,\action)$ is
  assumed to attain a local maximum at a state in $\G$.
  This condition reflects the intended relation between the reward
  function and the goal set.
  More generally, it is desirable that reaching the goal set be
  favorable under the return objective.
  Otherwise, return maximization and reaching the goal set may
  represent competing objectives.
\end{remark}

The distance from state $\state$ to the goal set is defined as: $
\goaldist(\state) := \inf_{\state' \in \G} \,\|\state - \state'\| $,
where $\goaldist(\state)=0$ if and only if $\state \in \G$.

The following two properties characterize policies that reliably
drive the system to the goal:

\begin{dfn}[$\eps$-improbable goal-reaching property]\label{dfn:eta_improbable}
  A policy $\policy \in \policies_{\mathrm{nstat}}$ satisfies the
  $\eps$-improbable goal-reaching property for some $\eps \in [0,1)$
  if, for all initial states $\state_0 \in \states$,
  \[
    \label{eqn_introstab}
    \PP{ \goaldist(\State_t^{\policy}(\state_0)) \,\xrightarrow{t \to
      \infty}\, 0
    }
    \ge 1 - \eps.
  \]
  In other words, the system reaches the goal set in the limit with
  probability at least \(1-\eps\).
\end{dfn}

Definition~\ref{dfn:eta_improbable} is intentionally asymptotic: it
does not assert finite-time entry into $\G$.
Properties concerning finite-time entry into prescribed neighborhoods
of $\G$ are studied in Section~\ref{sec:uniform_lp_bounds} and
Appendix~\ref{sec:uniform_hitting_time}.

More precise control over the convergence rate requires a stronger
condition.
The definition uses the standard notion of class-$\mathcal{KL}$ functions,
which are fundamental to control theory and provide the mathematical
framework for characterizing convergence behavior.

\begin{dfn}[Class-$\mathcal{KL}$ function]\label{dfn:kl_function}
  A continuous mapping \(
  \beta:\,[0,\infty)\times[0,\infty)\;\longrightarrow\;[0,\infty) \)
  is said to be of class $\mathcal{KL}$ if and only if the following
  two properties hold:

  \begin{enumerate}
    \item For every fixed time $t\ge 0$, the function $
      d\mapsto\beta(d,t) $ is of class $\mathcal{K}$; that is

      \begin{enumerate}
        \item $\beta(0,t)=0$ and $\beta(d,t)>0$ for all $d>0$,
        \item $d\mapsto\beta(d,t)$ is strictly increasing in $d$ and continuous.
      \end{enumerate}

    \item For every fixed $d\ge 0$, the function $
      t\mapsto\beta(d,t) $ is monotonically decreasing and satisfies $
      \lim_{t\to\infty} \beta(d,t) = 0 .
      $
  \end{enumerate}
\end{dfn}

\begin{dfn}[Uniform $L^1$ goal-reaching property]
    \label{dfn:uniform_lp_goal_reaching}
    A stationary policy $\policy\in\policies_{\mathrm{stat}}$
    satisfies the \emph{uniform $L^1$ goal-reaching property} if
    there exists a function $\beta\in\KL$ such that, for every
    initial state $\state_0\in\states$ and every $t\in\mathbb{Z}_{\geq0}$,
    \begin{equation}
      \E{
        \goaldist\bigl(\State_t^{\policy}(\state_0)\bigr)
      }
      \leq
      \beta\bigl(\goaldist(\state_0),t\bigr).
      \label{eq:uniform_lp_goal_reaching}
    \end{equation}
    Uniformity means that the same function $\beta$ applies to every
    initial state.
\end{dfn}

\begin{remark}
    The uniform $L^1$ goal-reaching property in
    Definition~\ref{dfn:uniform_lp_goal_reaching} is used to establish
    Theorem~\ref{thm:lp_first_hitting} in
    Section~\ref{sec:uniform_lp_bounds}.
    Appendix~\ref{sec:uniform_hitting_time} introduces the stronger
    uniform almost-sure goal-reaching property and establishes an
    analogue of Theorem~\ref{thm:lp_first_hitting} with stronger
    uniform bounds for CALF-Wrapper.
\end{remark}

\section{Proposed Approach}\label{sec:approach}

\emph{CALF-Wrapper} offers a general methodology for strengthening RL
policies through an adaptive policy-selection mechanism.
This approach dynamically switches between a performance-optimized
base RL policy and a fallback controller that satisfies the
goal-reaching property in Definition~\ref{dfn:eta_improbable} or
Definition~\ref{dfn:uniform_lp_goal_reaching}.
The selection mechanism uses value-function evaluations.

The setup in Algorithm~\ref{alg_calfwrapper} assumes access to

1) A \emph{base policy} $\basepolicy \in \policies_{\mathrm{stat}}$
parameterized by a deep neural network, learned via a standard RL
procedure to optimize~\eqref{eq:rl_objective}, aided by a value-function
estimate $\Valuebase$.

2) A \emph{fallback policy}
\(\fallbackpolicy\in\policies_{\mathrm{stat}}\) assumed to satisfy either the
$\eps$-improbable goal-reaching property in
Definition~\ref{dfn:eta_improbable} for some $\eps$ or the uniform
$L^1$ goal-reaching property in
Definition~\ref{dfn:uniform_lp_goal_reaching}.
The fallback policy need not be reward-optimal in terms
of~\eqref{eq:rl_objective}.

\begin{remark}
  \label{rem:value_estimate_construction}
  Not every RL algorithm provides a state-value estimate
  $\Valuebase(\state)$ directly.
  For example, deep deterministic policy gradient (DDPG), TD3, and SAC learn action-value
  critics~\cite{Lillicrap2016Continuouscont,Fujimoto2018,Haarnoja2019}.
  If an action-value estimate
  $\widehat{Q}^{\basepolicy}(\state,\action)$ is available, then, for
  a deterministic base policy, the simplest construction is \(
    \Valuebase(\state) := \widehat{Q}^{\basepolicy}(\state,\basepolicy(\state)).
  \) For a stochastic base policy, the corresponding expectation can
  be approximated by sampling actions, while a cheaper alternative is
  to evaluate the action-value estimate at a representative action,
  such as
  $\mathbb{E}_{\Action\sim\basepolicy(\bullet\mid\state)}[\Action]$.
  If computational efficiency is prioritized, a fixed representative
  action can also be used, such as the zero action when it belongs to
  the action space or the center of a bounded action space.
  This choice avoids evaluation of the base policy when constructing
  the critic estimate.
  These constructions need not equal the expectation over the action
  distribution.
  More generally, critics learned in practice are imperfect estimates
  of the theoretical quantities they are intended to
  represent~\cite{Fu2019DiagnosingBottlenecks}.
  CALF-Wrapper does not require a critic estimate to coincide exactly
  with its theoretical counterpart and can operate with such
  imperfect estimates.
  The experiments in Section~\ref{sec:critic_noise_robustness} show
  substantial degradation of the episode return only for sufficiently
  large fluctuations in the critic estimate.
\end{remark}

\subsection{Action Selection Mechanism and Its Benefits}

\subsubsection{Critic and Random Acceptance Conditions}
\label{sec:critic_based_decision_rule}

At each time step, CALF-Wrapper decides whether to continue using the
performance-seeking base policy or to apply the fallback policy.
This decision is guided by the base-policy critic.
A sufficiently large increase in the critic value indicates that
another action from the base policy may improve the accumulated return.
An insufficient increase causes CALF-Wrapper to prefer the fallback policy.

To identify such increases, CALF-Wrapper stores a reference critic
value $\Value_t^\dagger$, initialized as
$\Value_0^\dagger=\Valuebase(\State_0)$ in
line~\ref{lst:line:initialize} of Algorithm~\ref{alg_calfwrapper}, and
uses an improvement threshold $\nu>0$.
If the current critic value exceeds $\Value_t^\dagger$ by at least
$\nu$, the base policy is selected and the reference value is updated.
Otherwise, the reference value remains unchanged.
Because a learned critic is an imperfect estimate, CALF-Wrapper can
still select the base policy with probability $\rho_t(\State_t)$ when
the required increase is not observed.

The rule also allows the fallback policy to be selected unconditionally
at prescribed states.
Let $\states(\fallbackpolicy)\subseteq\states$ denote the set of these
states.
The default choice $\states(\fallbackpolicy)=\G$ selects the fallback
policy unconditionally on the goal set.
The choice $\states(\fallbackpolicy)=\varnothing$ introduces no
unconditional selections and leaves the remaining selection rule unchanged.

\begin{figure}[t]
  \centering
  \includegraphics[width=0.98\columnwidth]{
    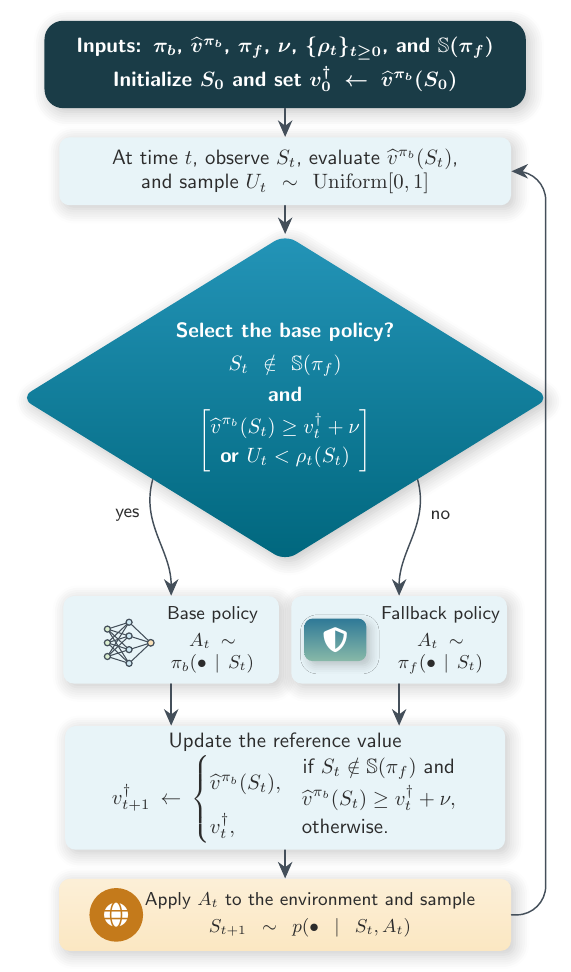
  }
  \caption{Diagram of Algorithm~\ref{alg_calfwrapper}.}
  \label{fig:calf_wrapper_algorithm_diagram}
\end{figure}

The complete decision rule is implemented in
lines~\ref{lst:line:selection_conditions}--%
\ref{lst:line:accept_fallback} of Algorithm~\ref{alg_calfwrapper} and illustrated in
\Cref{fig:calf_wrapper_algorithm_diagram}.
If $\State_t\in\states(\fallbackpolicy)$, CALF-Wrapper selects the
fallback policy.
Otherwise, either of two conditions can select the base policy.
The \emph{critic condition} is
\begin{equation*}
  \Valuebase(\State_t)\geq\Value_t^\dagger+\nu.
\end{equation*}
The \emph{random acceptance condition} is
\begin{equation*}
  U_t<\rho_t(\State_t),
\end{equation*}
where $U_t$ is sampled uniformly from $[0,1]$.
The random acceptance condition therefore holds with probability
$\rho_t(\State_t)$ and selects the base policy when it holds.
If neither condition holds, CALF-Wrapper selects the fallback policy.
The reference value is updated to $\Valuebase(\State_t)$ only when the
critic condition holds; it remains equal to $\Value_t^\dagger$ in
all other cases, as specified in
line~\ref{lst:line:benchmark_update}.

The functions $\rho_t:\states\to[0,1]$ may be selected subject to
\begin{equation*}
  \sum_{t=0}^{\infty}\sup_{\state\in\states}\rho_t(\state)<\infty.
\end{equation*}
This condition implies that the probability that the random acceptance
condition holds converges uniformly to zero.

\subsubsection{Intuition Behind the Mechanism}

The critic $\Valuebase$ serves as an indicator of ongoing
performance, so the rule checks whether the state's value remains
above the running benchmark $\Value_t^\dagger$.
If so, it indicates that the trajectory has improved markedly since
the most recent update of the running benchmark value $\Value_t^\dagger$.
Accordingly, the rule treats the trajectory as favorable and accepts
the calculated risk of executing the base-policy action in pursuit of
long-term gains.
Conversely, a value below the benchmark indicates that the trajectory
has deviated from the desired course, so the rule reverts to the
fallback policy $\fallbackpolicy$.
The random acceptance probability $\rho_{t}(\State_t)$ helps avoid rejecting
$\basepolicy$ prematurely due to minor fluctuations in $\Valuebase$,
acting like a simulated annealing mechanism that occasionally permits
risk-taking for potential long-term gains.

An analogy can be drawn with Lyapunov functions in control theory.
In control systems, provable stability is achieved by ensuring the
controller moves toward decreasing the Lyapunov function.
Similarly, outside
$\states(\fallbackpolicy)$, selecting the base policy $\basepolicy$
when $\Valuebase$ increases can be interpreted as steering the system
toward decreasing $-\Valuebase$, analogous to moving toward a
decrease of a Lyapunov function.
This interpretation motivates the name CALF-Wrapper (Critic as
Lyapunov Function Wrapper).

\subsubsection{Adjustable Trade-off Between Performance and the
Goal-Reaching Property} The implementation uses
$\bar\rho_t:=\lambda^t p_{\mathrm{relax}}$, where
$\lambda\in[0,1)$ and $p_{\mathrm{relax}}\in[0,1]$, as the summable
majorant of $\rho_t(\state)$.
The implementation sets $\rho_t(\state)=\bar\rho_t$ by default.
The hyperparameter $p_{\text{relax}}$ represents the initial trust in
the base policy, and $\lambda$ determines how quickly this trust
decays over time.
On finite horizons, setting $p_{\text{relax}} \approx 1$ and $\lambda
\approx 1$ makes the behavior nearly identical to the original base
policy, preserving its performance advantages.
By tuning $p_{\text{relax}}$ and $\lambda$, one can bias the policy
toward $\basepolicy$ (thus recovering its performance) or impose more
frequent switches to $\fallbackpolicy$.
In many cases, this approach can even outperform both $\basepolicy$
and $\fallbackpolicy$ in terms of cumulative reward (see
Section~\ref{sec:experiments}).

\subsubsection{Seamless Integration}
CALF-Wrapper is designed for seamless integration into existing RL pipelines.
It is implemented as a wrapper around the vectorized
environment that follows the Gymnasium wrapper
pattern~\cite{gym-wrapper}: at each step, it receives the action of the
base policy, replaces it with the action of the fallback policy when
required, and passes the resulting action to the environment.
It operates as a non-intrusive wrapper
around trained RL policies, requiring no modifications to the
original training algorithm.

\subsubsection{Rigorous Theoretical Analysis}
The theoretical analysis establishes how the goal-reaching properties
assumed for the fallback policy lead to corresponding properties of
CALF-Wrapper.
Theorem~\ref{thm:etagoalreaching} establishes that CALF-Wrapper
inherits the $\eps$-improbable goal-reaching property in
Definition~\ref{dfn:eta_improbable} from the fallback policy under the
stated assumptions.
Section~\ref{sec:uniform_lp_bounds} establishes uniform properties of
CALF-Wrapper when the fallback policy satisfies the uniform $L^1$
goal-reaching property in
Definition~\ref{dfn:uniform_lp_goal_reaching}.
Appendix~\ref{sec:uniform_hitting_time} establishes stronger claims
under a uniform almost-sure condition on the fallback policy.

\begin{algorithm}[t]
  \caption{Critic as Lyapunov Function Wrapper}\label{alg_calfwrapper}
  \begin{algorithmic}[1]

    \REQUIRE\label{lst:requirements} \text{}
    \begin{itemize}[leftmargin=0.4em]
      \item $\fallbackpolicy \in \policies_{\mathrm{stat}}$: Fallback
        policy satisfying Definition~\ref{dfn:eta_improbable} or
        Definition~\ref{dfn:uniform_lp_goal_reaching}.
      \item $\basepolicy\in \policies_{\mathrm{stat}}$: Base policy
        trained via an RL method.
      \item $\Valuebase$: A value-function estimate trained alongside
        $\basepolicy$ via the RL method.
      \item $\nu > 0$: Minimum improvement threshold for updating the
        best observed critic value $\Value_t^\dagger$.
      \item $\states(\fallbackpolicy)\subseteq\states$:
        States at which the fallback policy is selected unconditionally.
        The default choice is $\states(\fallbackpolicy)=\G$.
        The choice $\states(\fallbackpolicy)=\varnothing$ leaves the
        remaining selection rule unchanged.
      \item $\{\rho_{t}(\state)\}_{t \geq 0}$: Random acceptance
        probabilities.
        They are assumed to be bounded by a summable majorant
        $\{\bar{\rho}_t\}_{t\ge 0}\subset[0,1]$, meaning
        $\rho_t(\state)\le\bar{\rho}_t \; \forall t,\state$ and
        $\sum_{t=0}^\infty \bar{\rho}_t<\infty$.
    \end{itemize}

    \STATE\label{lst:line:initialize} \textbf{Initialize}:
    \begin{itemize}
      \item $\State_0 \sim \transit_0(\bullet)$ or set $\State_0 =
        \state_0$ for some $\state_0 \in \states$
      \item $\Value_0^\dagger := \Valuebase(\State_0)$
    \end{itemize}

    \FOR{$t = 0,1,2,\dots$}
    \STATE\label{lst:u_t} Sample $U_t$ independently from $\operatorname{Uniform}[0,1]$
    \IF{$\State_t\!\notin\!\states(\fallbackpolicy)\;\land\;
    [\Valuebase(\State_t)\!\geq\!\Value_t^\dagger\!+\!\nu
    \;\lor\; U_t\!<\!\rho_t(\State_t)]$}
    \label{lst:line:selection_conditions}
    \STATE\label{lst:line:accept_base} $\Action_t\la$ sampled from
    $\basepolicy(\bullet \mid \State_t)$
    \ELSE
    \STATE\label{lst:line:accept_fallback} $\Action_t\la$ sampled
    from $\fallbackpolicy(\bullet \mid \State_t)$
    \ENDIF
    \STATE\label{lst:line:benchmark_update}
    $
    \displaystyle
    \Value_{t+1}^\dagger :=
    \begin{cases}
      \Valuebase(\State_t),\ \text{if }
      \State_t\!\notin\!\states(\fallbackpolicy)\,\text{and}\,
      \Valuebase(\State_t)\!\geq\!\Value_t^\dagger\!+\!\nu \\
      \Value_{t}^\dagger, \text{ otherwise }
    \end{cases}
    $
    \STATE $\State_{t+1} \la$ sampled from $\transit(\bullet \mid
    \State_t, \Action_t)$
    \ENDFOR
  \end{algorithmic}
\end{algorithm}

\subsection{Computational Complexity}
\label{sec:computational_complexity}
Consider execution over $T$ time steps, and let $N_{\basepolicy}$ be
the number of time steps at which the base policy supplies the applied action.
By definition, $N_{\basepolicy}\leq T$.
The selection rule in
line~\ref{lst:line:selection_conditions} of
Algorithm~\ref{alg_calfwrapper} selects the base policy only when the
state lies outside $\states(\fallbackpolicy)$ and either the critic
condition or the random acceptance condition holds.
The number of such selections is $N_{\basepolicy}$.
Outside $\states(\fallbackpolicy)$, decreasing
$\rho_t(\State_t)$ reduces selections based on the random acceptance
condition but does not prevent selection of the base policy when the
critic condition holds.
Deployment of the base policy without CALF-Wrapper requires $T$
base-policy evaluations.
When $\Valuebase$ is provided directly by a state-value critic or
obtained by evaluating an action-value critic at one fixed
representative action, as described in
Remark~\ref{rem:value_estimate_construction}, CALF-Wrapper requires
\begin{equation*}
  \begin{aligned}
    &T\text{ critic evaluations} \\
    &\quad+N_{\basepolicy}\text{ base-policy evaluations} \\
    &\quad+(T-N_{\basepolicy})\text{ fallback-policy evaluations} \\
    &\quad+O(T)\text{ scalar operations}.
  \end{aligned}
\end{equation*}
The $O(T)$ scalar operations consist, at each time step, of one
uniform random-number sample, evaluation of $\rho_t(\State_t)$,
evaluation of whether
$\State_t\in\states(\fallbackpolicy)$, a constant number of
comparisons, and the update of $\Value_t^\dagger$.
In the reported implementation, membership in
$\states(\fallbackpolicy)$ is evaluated using the corresponding
environment-specific goal-neighborhood condition.
These operations require only scalar arithmetic and comparisons and
therefore have a small computational cost compared with evaluations
of the neural-network policies and critic.

When $\Valuebase(\state)$ is obtained from an action-value critic as
$\widehat{Q}^{\basepolicy}(\state,\basepolicy(\state))$ for a
deterministic base policy or as \( \widehat{Q}^{\basepolicy} \bigl(
    \state,
    \mathbb{E}_{\Action\sim\basepolicy(\bullet\mid\state)}[\Action]
\bigr) \) for a stochastic base policy, the corresponding base-policy
output must be computed before the critic is evaluated.
In this case, CALF-Wrapper requires
\begin{equation*}
  \begin{aligned}
    &T\text{ base-policy evaluations} \\
    &\quad+T\text{ critic evaluations} \\
    &\quad+(T-N_{\basepolicy})\text{ fallback-policy evaluations} \\
    &\quad+O(T)\text{ scalar operations}.
  \end{aligned}
\end{equation*}
The persistent memory required by the wrapper consists of the fixed
critic parameters and a constant number of scalar values, including
$\Value_t^{\dagger}$ and the current random acceptance probability.
Algorithm~\ref{alg_calfwrapper} stores no trajectory, replay buffer,
or sequence of previous critic values.

Consider a fully connected perceptron with $K$ layers and layer
widths $H_0,H_1,\ldots,H_K$, where $H_0$ is the input dimension and
$H_K$ is the output dimension.
One forward evaluation requires $O(\sum_{k=1}^{K}H_{k-1}H_k)$
arithmetic operations for the matrix multiplications.
The bias additions and elementwise activation functions require
$O(\sum_{k=1}^{K}H_k)$ additional operations and do not change this order.
If the base policy, fallback policy, and critic have this form, the
evaluation counts above give an overall time complexity of
$O(T\sum_{k=1}^{K}H_{k-1}H_k)$.
Thus, CALF-Wrapper increases the number of forward evaluations but
does not change the asymptotic time complexity with respect to $T$
and the layer widths.

\section{Goal-Reaching Property of CALF-Wrapper}
\label{sec:goal_reaching_property}

This section proves that CALF-Wrapper inherits the goal-reaching
property from the fallback policy under the stated assumptions.
The central result of this section is Theorem~\ref{thm:etagoalreaching}.
The superscript $\mathrm{P}$ in the assumption and claim labels in
this section denotes ``pathwise.''
Together with the superscript $\mathrm{U}$ introduced in
Section~\ref{sec:uniform_lp_bounds}, this convention makes every
assumption and claim label unique and identifies the analysis to
which it belongs.

The proof uses the following explicit form of the policy
$\calfwpolicy$ defined by \Cref{alg_calfwrapper}.

\begin{dfn}\label{dfn:calfwpolicy}
  Let $\calfwpolicy \in \policies_{\mathrm{nstat}}$ be the policy
  defined by \Cref{alg_calfwrapper}.
  The subscript $w$ indicates that $\calfwpolicy$ is the ``wrapped
  policy'' produced by CALF-Wrapper.
  Specifically, for every $t \in \Z_{\geq0}$, set
  \begin{equation*}
    \calfwpolicy(\bullet\mid\State_t,t) :=
    \begin{cases}
      \basepolicy(\bullet\mid\State_t), &
      \begin{aligned}
        &\text{if }\State_t\notin\states(\fallbackpolicy)
          \text{ and}\\[-0.2ex]
        &\bigl(\Valuebase(\State_t)\geq\Value_t^\dagger+\nu\\[-0.2ex]
        &\quad\text{or }U_t<\rho_t(\State_t)\bigr),
      \end{aligned}\\
      \fallbackpolicy(\bullet\mid\State_t), & \text{otherwise.}
    \end{cases}
  \end{equation*}
\end{dfn}

\begin{theorem}\label{thm:etagoalreaching}
  Consider \Cref{alg_calfwrapper}.

  \noindent Assume that:
  \begin{enumerate}[label={$\bigl(\mathrm{A}_{\arabic*}^{\mathrm{P}}\bigr)$},
      leftmargin=2.5em]
    \item\label{ass:finite_critic_supremum}
      $\bar{\Value}:=\sup_{\state\in\states}\Valuebase(\state)<\infty$.
    \item\label{ass:fallback_goal_reaching} The fallback policy
      $\fallbackpolicy$ satisfies the $\eps$-improbable goal-reaching
      property in Definition~\ref{dfn:eta_improbable}.
  \end{enumerate}

  \noindent Then the following claim holds:
  \begin{enumerate}[label={$\bigl(\mathrm{C}_{\arabic*}^{\mathrm{P}}\bigr)$},
      leftmargin=2.5em]
    \item\label{claim:improbable_goal_reaching} The policy
      $\calfwpolicy$ satisfies the $\eps$-improbable goal-reaching
      property in Definition~\ref{dfn:eta_improbable}.
  \end{enumerate}
\end{theorem}

\begin{proof}
  See Appendix~\ref{sec:proof_goal_reaching}.
\end{proof}

\label{sec:finite_critic_supremum_discussion}
Assumption~\ref{ass:finite_critic_supremum} requires
$\sup_{\state\in\states}\Valuebase(\state)<\infty$.
Remark~\ref{rem:reward_goal_set} assumes that
$\sup_{\state\in\states,\,\action\in\actions}
\reward(\state,\action)<+\infty$.
For $\gamma<1$, the exact value function of every policy
$\policy\in\policies_{\mathrm{nstat}}$ therefore satisfies, for every
$\state\in\states$,
\[
  \Value^{\policy}(\state)
  \leq
  \sum_{k=0}^{\infty}\gamma^k
  \sup_{\state'\in\states,\,\action\in\actions}
  \reward(\state',\action)
  <+\infty.
\]
Consequently, the exact value of the base policy does not exceed the
theoretical upper bound in the displayed inequality.
A learned critic does not necessarily satisfy
$\sup_{\state\in\states}\Valuebase(\state)<\infty$ unless this
condition is imposed on its output.
Using this theoretical upper bound as a clipping value for the
learned critic is a natural choice because critic outputs above this
value cannot represent an exact policy value attainable in the
underlying control problem.
The clipping can be applied when the critic is evaluated by
Algorithm~\ref{alg_calfwrapper} at deployment and does not restrict
how the critic is trained.
\begin{remark}
  For an infinite horizon with $\gamma=1$, the boundedness of the
  reward from above is not sufficient for this argument.
  If $\reward(\state,\action)\leq0$ for every $\state\in\states$ and
  $\action\in\actions$, then every partial return is nonpositive.
  If the value function exists as a finite real-valued function as in
  Section~\ref{sec_problem}, then $\Value^{\policy}(\state)\leq0$ for
  every $\state\in\states$.
  In this case, zero is a theoretical upper bound for the learned critic.
\end{remark}
All experiments reported in \Cref{sec:experiments} evaluate the
learned critic after clipping its output at the environment-specific
upper value in \Cref{tab:critic_upper_clipping_values}.
The reported results therefore use the critic that satisfies the
finite upper bound required by Assumption~\ref{ass:finite_critic_supremum}.
\begin{table}[H]
  \centering
  \caption{Upper clipping values used in the evaluations in
  Section~\ref{sec:experiments} for the environments described in
  Appendix~\ref{app:environment_descriptions}.}
  \label{tab:critic_upper_clipping_values}
  \begin{tabularx}{\columnwidth}{@{}Xc@{}}
    \toprule Environment & Upper clipping value \\ \midrule Pendulum
    & $0$ \\ CartPole Swing-Up & $0$ \\ Treasure-Collecting Robot &
    $50$ \\ Contaminated-Zone AUV Navigation & $0$ \\ \bottomrule
  \end{tabularx}
\end{table}
The value $0$ applies to Pendulum, CartPole Swing-Up, and
Contaminated-Zone Autonomous Underwater Vehicle (AUV) Navigation because their rewards are
nonpositive at every step, as specified in
Appendix~\ref{app:environment_descriptions}.
For the Treasure-Collecting Robot, the treasure-collection bonus is
$50$ and can be received at most once per episode.
The distance-to-goal term and every reward after the treasure has
been collected are nonpositive.
Therefore, the return of every episode is at most $50$.

The evaluations use upper clipping to impose a finite upper bound on
the critic output in the considered environments.

\newcommand{\theoremoneproofappendix}{%
\section{Proof of Theorem 1}
\label{sec:proof_goal_reaching}
\begin{proof}
  The key observation is that Algorithm~\ref{alg_calfwrapper}
  eventually uses only the fallback policy, thus inheriting its
  goal-reaching property.
  This follows because the total number of base-policy selections is
  almost surely finite.

  Let $N_{\basepolicy} := \sum_{t=0}^{\infty}
  \mathbf{1}\{\basepolicy\text{ is selected at time }t\}$ denote
  the total number of times the base policy is chosen.
  $\mathbf{1}\{\mathcal{E}\}$ denotes the
  indicator of an event $\mathcal{E}$; it equals one when
  $\mathcal{E}$ occurs and zero otherwise.
  By Definition~\ref{dfn:calfwpolicy}, the base
  policy can be selected only when
  $\State_t\notin\states(\fallbackpolicy)$.
  At such a state, the base policy is selected when either (i) the
  critic condition
  $\Valuebase(\State_t) \geq \Value_t^\dagger + \nu$ holds or (ii) the
  random acceptance condition $U_t < \rho_t(\State_t)$ holds.
  Therefore,
  \begin{equation*}
    N_{\basepolicy} \leq N_{\Value^\dagger} + N_{\rho}
  \end{equation*}
  where
  \begin{align*}
    N_{\Value^\dagger}
    &:=\sum_{t=0}^{\infty}
    \mathbf{1}\!\left\{
      \State_t\notin\states(\fallbackpolicy),
      \Valuebase(\State_t)\geq\Value_t^\dagger+\nu
    \right\},\\
    N_{\rho}
    &:=\sum_{t=0}^{\infty}
    \mathbf{1}\!\left\{
      \State_t\notin\states(\fallbackpolicy),
      U_t<\rho_t(\State_t)
    \right\}.
  \end{align*}
  The quantity $N_{\Value^\dagger}$ counts selections based on the
  critic condition, and $N_{\rho}$ counts selections based on the
  random acceptance condition.
  Both terms are finite:

  \emph{Finiteness of $N_{\Value^\dagger}$}: At
  every time counted by $N_{\Value^\dagger}$, the value
  $\Value_t^\dagger$ increases by at least $\nu$.
  By Assumption~\ref{ass:finite_critic_supremum}, $\Valuebase$ is
  bounded above by $\bar{\Value}$.
  Since $\Value_t^\dagger$ starts at $\Valuebase(\State_0)$, the
  value can increase only finitely many times.
  Thus, $N_{\Value^\dagger} < \infty$.

  \emph{Finiteness of $N_{\rho}$}: Since $\rho_t(\State_t) \leq
  \bar{\rho}_t$ for all $t$ and $\state$, and $\sum_{t=0}^\infty
  \bar{\rho}_t < \infty$ by assumption, the Borel--Cantelli
  lemma~\cite{billingsley1995probability} ensures that the event $U_t
  < \bar{\rho}_t$ occurs only finitely many times almost surely.
  Therefore, $N_{\rho} < \infty$ almost surely.

  Combining these results gives $N_{\basepolicy} \leq
  N_{\Value^\dagger} + N_{\rho} < \infty$ almost surely.
  For $k\in\mathbb{Z}_{\geq0}$, let
  \[
    \mathbb{F}_k :=\{\fallbackpolicy\text{ is selected at every time }
    t\geq k\}.
  \]
  Since the number of base-policy activations is almost surely
  finite, the events $\mathbb{F}_k$ are increasing and
  $\PP{\bigcup_{k=0}^{\infty}\mathbb{F}_k}=1$.
  Let $\mathbb{C}$ denote the event that the trajectory generated by
  CALF-Wrapper converges to $\G$.
  The event $\mathbb{C}^{c}$ is the complement of $\mathbb{C}$ and
  denotes the event that this trajectory does not converge to $\G$.
  For every deterministic $k$, the Markov property,
  Assumption~\ref{ass:fallback_goal_reaching}, and
  Definition~\ref{dfn:eta_improbable} give
  \begin{align*}
    &\PP{\goaldist\bigl(\State_j^{\fallbackpolicy}(\State_k)\bigr)
    \not\xrightarrow{j\to\infty}0}\\
    &\quad=
    \E{\PP{\goaldist\bigl(\State_j^{\fallbackpolicy}(\State_k)\bigr)
    \not\xrightarrow{j\to\infty}0\,\middle|\,\State_k}}
    \leq\eps.
  \end{align*}
  On $\mathbb{F}_k$, CALF-Wrapper follows the fallback policy from
  time $k$ onward, and hence $\PP{\mathbb{C}^{c}\cap\mathbb{F}_k}\leq\eps$.
  Therefore, continuity of probability from below yields
  \[
    \PP{\mathbb{C}^{c}}
    =\lim_{k\to\infty}\PP{\mathbb{C}^{c}\cap\mathbb{F}_k} \leq\eps.
  \]
  Thus, $\calfwpolicy$ satisfies the $\eps$-improbable goal-reaching property.
\end{proof}
}

\section{Uniform Bounds Under the $L^1$ Goal-Reaching Property}
  \label{sec:uniform_lp_bounds}

  This section establishes Theorem~\ref{thm:lp_first_hitting} when the
  fallback policy satisfies the uniform $L^1$ goal-reaching property
  in Definition~\ref{dfn:uniform_lp_goal_reaching}.
  The superscript $\mathrm{U}$ in the assumption and claim labels in
  this section denotes ``uniform.''
  Theorem~\ref{thm:lp_first_hitting} establishes three claims:
  \begin{itemize}
    \item Claim~\ref{claim:lp_uniform_moment} bounds the expected value of
          the distance from $\G$ uniformly over time and bounds the
          probability of exceeding a prescribed distance during a
          finite time interval.
    \item Claim~\ref{claim:lp_hitting_bound} gives a lower bound on the
          probability of entering a prescribed neighborhood of $\G$ by
          a specified time and establishes that the hitting time is
          finite almost surely.
    \item Under the additional assumptions,
          Claim~\ref{claim:lp_reaching_bound} gives a lower bound on the
          probability that the trajectory enters the prescribed
          neighborhood of $\G$ and remains there at every
          subsequent time, and establishes that the settling time is
          finite almost surely.
  \end{itemize}
  To state Theorem~\ref{thm:lp_first_hitting}, fix
  $d^{\circ}\in\mathbb{R}_{>0}$ and define the following quantities,
  whose interpretations are given in
  Remark~\ref{rem:lp_quantities_interpretation}.
  \begin{align}
    v_{\min}(d^{\circ})
    &:={\min}\left\{\Valuebase(\state):
    \goaldist(\state)\leq d^{\circ}\right\},
    \label{eq:lp_vmin}
    \displaybreak[0]\\
    \mathbb{V}^{\basepolicy}(d^{\circ})
    &:=\left\{\state\in\states:
    \Valuebase(\state)\geq v_{\min}(d^{\circ})\right\},
    \label{eq:lp_superlevel_set}
    \displaybreak[0]\\
    v_{\max}(d^{\circ})
    &:={\max}\left\{\Valuebase(\state):
    \state\in\mathbb{V}^{\basepolicy}(d^{\circ})\right\},
    \label{eq:lp_vmax}
    \displaybreak[0]\\
    \tau(d^{\circ})
    &:=\max\left\{1,
      \left\lfloor
      \frac{v_{\max}(d^{\circ})-v_{\min}(d^{\circ})}{\nu}
    \right\rfloor\right\},
    \label{eq:lp_critic_count}
    \displaybreak[0]\\
    d_{\bar p}(d^{\circ})
    &:=\sup\left\{\bar p(\state,\action):
      \state\in\mathbb{V}^{\basepolicy}(d^{\circ}),
    \ \action\in\actions\right\},
    \label{eq:lp_post_action_bound}
    \displaybreak[0]\\
    \delta(d^{\circ})
    &:=\max\left\{
      d^{\circ},
      d_{\bar p}(d^{\circ}),
      \sup_{\state\in\mathbb{V}^{\basepolicy}(d^{\circ})}
      \goaldist(\state)
    \right\}.
    \label{eq:lp_fallback_initial_bound}
  \end{align}
  For $t\in\mathbb{Z}_{\geq0}$ and $\ell\in\mathbb{Z}_{>0}$, define
  \begin{align}
    \mathcal{P}(t)
    &:=\prod_{k=t}^{\infty}(1-\bar\rho_k),
    \label{eq:lp_no_relaxation_probability}
    \\
    C(d^{\circ})
    &:=1+\tau(d^{\circ})+
    \sum_{k=0}^{\infty}\bar\rho_k,
    \label{eq:lp_expected_candidate_count}
    \\
    \Delta(d^{\circ})
    &:=\delta(d^{\circ})
    +C(d^{\circ})\beta(\delta(d^{\circ}),0),
    \label{eq:lp_uniform_distance_bound}
    \\
    H(d^{\circ},t,\ell)
    &:=\bigl(\tau(d^{\circ})+t+1\bigr)\ell.
    \label{eq:lp_block_horizon}
  \end{align}
  \begin{dfn}[Hitting and settling times]
    \label{dfn:lp_hitting_reaching_times}
    For an initial state $\state_0\in\states$ and $d^{*}>0$, define the
    following times, where the time indices $t$ and $k$ range over
    $\mathbb{Z}_{\geq0}$:
    \begin{equation}
      T_{\mathrm{hit}}(\state_0,d^{*})
      :=\inf\Bigl\{t\geq0:
      \goaldist\bigl(\State_t^{\calfwpolicy}(\state_0)\bigr)
      \leq d^{*}\Bigr\}.
      \label{eq:lp_hitting_time}
    \end{equation}
    \begin{equation}
      T_{\mathrm{set}}(\state_0,d^{*})
      :=\inf\{t\geq0:\sup_{k\geq t}
      \goaldist\bigl(\State_k^{\calfwpolicy}(\state_0)\bigr)
      \leq d^{*}\},
      \label{eq:lp_reaching_time}
    \end{equation}
    where $\inf\varnothing:=+\infty$.
  \end{dfn}

  Consider Algorithm~\ref{alg_calfwrapper} initialized at any
  $\state_0$ satisfying $\goaldist(\state_0)\leq d^{\circ}$.
  The following assumptions are used:
  \begin{enumerate}[label={$\bigl(\mathrm{A}_{\arabic*}^{\mathrm{U}}\bigr)$},
      leftmargin=2.5em]
    \item\label{ass:lp_acceptance} For every initial state $\state_0$
      satisfying $\goaldist(\state_0)\leq d^{\circ}$,
      $\rho_t(\state)=0$ whenever $\Valuebase(\state)<\Valuebase(\state_0)$.
    \item\label{ass:lp_critic} The function $\Valuebase$ is
      continuous, and the set in
      \eqref{eq:lp_superlevel_set} is bounded.
    \item\label{ass:lp_uniform_fallback} The fallback policy
      satisfies the uniform $L^1$ goal-reaching property in
      Definition~\ref{dfn:uniform_lp_goal_reaching} with
      $\beta\in\KL$.
  \end{enumerate}

  The conclusions concerning $T_{\mathrm{set}}(\state_0,d^{*})$
  for a prescribed $d^{*}\in(0,d^{\circ})$ use the following
  additional assumptions:
  \begin{enumerate}[label={$\bigl(\mathrm{A}_{\arabic*}^{\mathrm{U}}\bigr)$},
      leftmargin=2.5em,start=4]
    \item\label{ass:lp_fallback_locking}
      The set $\states(\fallbackpolicy)$ satisfies
      $\{\state\in\states:\goaldist(\state)\leq d^{*}\}
      \subseteq\states(\fallbackpolicy)$.
    \item\label{ass:lp_summable_tail}
      The series satisfies
      $\sum_{k=0}^{\infty}\beta(d^{*},k)<+\infty$.
  \end{enumerate}

  \begin{remark}[Interpretation]
    \label{rem:lp_quantities_interpretation}
    The quantities in
    \eqref{eq:lp_vmin}--\eqref{eq:lp_block_horizon} have
    the following engineering interpretations.
    \begin{itemize}
      \item The quantity $d^{\circ}$ is the radius of the prescribed
        initial neighborhood of $\G$.
        The bounds in Theorem~\ref{thm:lp_first_hitting} hold uniformly
        over every initial state $\state_0$ satisfying
        $\goaldist(\state_0)\leq d^{\circ}$.
      \item The quantity $d^{*}\in(0,d^{\circ})$ is the radius of the
        prescribed target neighborhood of $\G$.
        The quantity $T_{\mathrm{hit}}(\state_0,d^{*})$ is the first
        time at which the trajectory initialized at $\state_0$ enters
        this neighborhood.
        The quantity $T_{\mathrm{set}}(\state_0,d^{*})$ is the first
        time after which the trajectory remains in this neighborhood
        at every subsequent time.
      \item The set $\mathbb{V}^{\basepolicy}(d^{\circ})
        =\{\state\in\states:\Valuebase(\state)\geq
        v_{\min}(d^{\circ})\}$ is the critic superlevel set determined
        by the smallest critic value in the prescribed
        $d^{\circ}$-neighborhood of $\G$.
      \item The quantity $\tau(d^{\circ})$ bounds how many increases of
        the stored critic value by at least $\nu$ can occur.
        It bounds the number of selections of the base policy based on
        the critic condition and does not bound the time of their final
        occurrence.
      \item The quantity $d_{\bar p}(d^{\circ})$ bounds the norm of
        every state that can occur after one transition from
        $\mathbb{V}^{\basepolicy}(d^{\circ})$ under an admissible action.
      \item The quantity $\delta(d^{\circ})$ bounds the prescribed
        initial distance from $\G$, the distance at every state at
        which the base policy can be selected, and the distance after
        an action supplied by the base policy.
      \item The quantity $\mathcal P(t)$ is the probability that the
        random acceptance condition does not hold at or after time $t$
        when each
        random acceptance probability is replaced with its upper bound
        $\bar\rho_k$.
        For $\bar\rho_k=\lambda^k p_{\mathrm{relax}}$, this probability
        has the explicit positive lower bound stated in
        Proposition~\ref{prop:relaxation_product_bound} in
        Appendix~\ref{sec:explicit_probability_bound}.
      \item The quantity $C(d^{\circ})-1$ bounds the expected total
        number of base-policy selections.
        The quantity $C(d^{\circ})$ bounds the expected number of
        states consisting of the initial state and the states
        immediately after base-policy actions.
      \item On the event that the random acceptance condition does not hold at or after
        time $t$, the first $H(d^{\circ},t,\ell)$ actions, where
        $H(d^{\circ},t,\ell)=\bigl(\tau(d^{\circ})+t+1\bigr)\ell$,
        contain at least $\ell$ consecutive actions supplied by the
        fallback policy.
      \item The quantity $\Delta(d^{\circ})$ bounds the expected
        distance from $\G$ uniformly over time.
        Claim~\ref{claim:lp_uniform_moment} also uses this quantity to
        bound the probability that the distance from $\G$ exceeds a
        prescribed value during a finite time interval.
    \end{itemize}
  \end{remark}

  \begin{remark}[Practical computation]
    The quantities $v_{\min}(d^{\circ})$, $v_{\max}(d^{\circ})$,
    and $\tau(d^{\circ})$ are required only to evaluate the bounds in
    Theorem~\ref{thm:lp_first_hitting}.
    As a rule of thumb, the computation of $v_{\max}(d^{\circ})$
    can be simplified by replacing its exact value in
    \eqref{eq:lp_critic_count} with the theoretical upper value used to
    clip the critic, as described in
    Section~\ref{sec:finite_critic_supremum_discussion}.
    The quantity $v_{\min}(d^{\circ})$ can be estimated by projected
    gradient descent from multiple initial states with projection onto
    $\{\state\in\states:\goaldist(\state)\leq d^{\circ}\}$, but this
    procedure does not certify the global minimum of a nonconvex critic.
    Among the available neural network verification solvers, the
    open-source $\alpha,\beta$-CROWN implementation supports
    $L_q$-norm input sets for $q\in\{1,2,\infty\}$ and computes
    certified output bounds~\cite{AlphaBetaCROWNGitHub}.
    Its bound-propagation methods include CROWN, auto\_LiRPA,
    $\alpha$-CROWN, and $\beta$-CROWN
    \cite{Zhang2018CROWN,Xu2020AutoLiRPA,Xu2021AlphaCROWN,Wang2021BetaCROWN}.
    For a critic with rectified linear unit (ReLU) activations, $\beta$-CROWN provides branch
    and bound with ReLU split constraints, while GenBaB extends branch
    and bound to general nonlinearities, including
    $\tanh$~\cite{Wang2021BetaCROWN,Shi2025GenBaB}.
        This remark is provided for practical reference, since the
    experiments in \Cref{sec:experiments} do not require these quantities.
  \end{remark}

  \begin{theorem}[Uniform bounds for CALF-Wrapper]
    \label{thm:lp_first_hitting}
    Consider Algorithm~\ref{alg_calfwrapper} initialized at any
    $\state_0$ satisfying $\goaldist(\state_0)\leq d^{\circ}$.
    Under Assumptions~\ref{ass:lp_acceptance}--%
    \ref{ass:lp_uniform_fallback},
    Claims~\ref{claim:lp_uniform_moment} and
    \ref{claim:lp_hitting_bound} hold.
    If Assumptions~\ref{ass:lp_fallback_locking}
    and~\ref{ass:lp_summable_tail} also hold, then
    Claim~\ref{claim:lp_reaching_bound} holds.
    \begin{enumerate}[label={$\bigl(\mathrm{C}_{\arabic*}^{\mathrm{U}}\bigr)$},
        leftmargin=2.5em]
      \item\label{claim:lp_uniform_moment} The distance from $\G$ satisfies
        \begin{equation}
          \sup_{t\in\mathbb{Z}_{\geq0}}
          \E{
            \goaldist\bigl(\State_t^{\calfwpolicy}(\state_0)\bigr)
          }
          \leq\Delta(d^{\circ}).
          \label{eq:lp_uniform_moment_bound}
        \end{equation}
        Consequently, for every $T\in\mathbb{Z}_{\geq0}$ and $R>0$,
        \begin{equation}
          \PP{
            \max_{0\leq k\leq T}
            \goaldist\bigl(\State_k^{\calfwpolicy}(\state_0)\bigr)>R
          }
          \leq
          (T+1)\left(\frac{\Delta(d^{\circ})}{R}\right).
          \label{eq:lp_prefix_overshoot_bound}
        \end{equation}

      \item\label{claim:lp_hitting_bound} For every
        $d^{*}\in(0,d^{\circ})$, $t\in\mathbb{Z}_{\geq0}$, and
        $\ell\in\mathbb{Z}_{>0}$,
        \begin{multline}
          \PP{T_{\mathrm{hit}}(\state_0,d^{*})
          \leq H(d^{\circ},t,\ell)}
          \\
          \geq
          \mathcal{P}(t)
            -C(d^{\circ})
            \frac{\beta(\delta(d^{\circ}),\ell)}{d^{*}}.
          \label{eq:lp_first_hitting_bound}
        \end{multline}
        In particular,
        \begin{equation}
          \PP{T_{\mathrm{hit}}(\state_0,d^{*})<+\infty}=1.
          \label{eq:lp_finite_hitting_time}
        \end{equation}

      \item\label{claim:lp_reaching_bound} For every
        $\underline{d}^{*}\in(0,d^{*}]$,
        $t\in\mathbb{Z}_{\geq0}$, and $\ell\in\mathbb{Z}_{>0}$,
        \begin{multline}
          \PP{T_{\mathrm{set}}(\state_0,d^{*})
          \leq H(d^{\circ},t,\ell)}
          \geq\\
          \mathcal{P}(t)
            -C(d^{\circ})\left[
              \frac{\beta(\delta(d^{\circ}),\ell)}{\underline{d}^{*}}
              +\frac{1}{d^{*}}
              \sum_{k=0}^{\infty}\beta(\underline{d}^{*},k)
            \right].
          \label{eq:lp_reaching_time_bound}
        \end{multline}
        In particular,
        $\PP{T_{\mathrm{set}}(\state_0,d^{*})<+\infty}=1$.

    \end{enumerate}
  \end{theorem}
  \begin{proof}
    See Appendix~\ref{sec:proof_uniform_lp_bounds}.
  \end{proof}

  \begin{remark}
    The right-hand sides of
    Claims~\ref{claim:lp_uniform_moment}--%
    \ref{claim:lp_reaching_bound} do not depend on the particular
    initial state $\state_0$.
    The same bounds therefore hold for every initial state satisfying
    $\goaldist(\state_0)\leq d^{\circ}$.
  \end{remark}

  \subsection{Discussion of the Assumptions}
  \label{sec:uniform_lp_assumptions}

  \subsubsection{Assumptions~\ref{ass:lp_acceptance} and
  \ref{ass:lp_fallback_locking}}
  Assumption~\ref{ass:lp_acceptance} restricts the region in which the
  selection mechanism can select the base policy.
  Its feasibility follows directly from the implementation choice of
  $\rho_t(\state)$, which can be set to zero whenever
  $\Valuebase(\state)<\Valuebase(\state_0)$.
  Assumption~\ref{ass:lp_fallback_locking} is directly feasible by
  choosing $\states(\fallbackpolicy)$ to contain the prescribed
  $d^{*}$-neighborhood of $\G$.

  \subsubsection{Assumptions~\ref{ass:lp_uniform_fallback} and
  \ref{ass:lp_summable_tail}}
  Assumptions~\ref{ass:lp_uniform_fallback} and
  \ref{ass:lp_summable_tail} specify the rate at which the expected
  distance from \(\G\) converges to zero when the fallback policy is applied.
  Since $\beta(\bullet,k)$ is increasing and $\beta(0,k)=0$, for every
  $\underline{d}^{*}\in(0,d^{*}]$,
  \begin{equation*}
    0\leq\beta(\underline{d}^{*},k)\leq\beta(d^{*},k)
    \quad\text{and}\quad
    \beta(\underline{d}^{*},k)\xrightarrow{\underline{d}^{*}\downarrow0}0.
  \end{equation*}
  Assumption~\ref{ass:lp_summable_tail} and the dominated convergence
  theorem for series therefore imply
  \begin{equation*}
    \sum_{k=0}^{\infty}\beta(\underline{d}^{*},k)<+\infty
    \quad\text{and}\quad
    \sum_{k=0}^{\infty}\beta(\underline{d}^{*},k)
    \xrightarrow{\underline{d}^{*}\downarrow0}0.
  \end{equation*}
  To obtain a sufficient condition for
  Assumption~\ref{ass:lp_summable_tail}, consider a standard
  decomposition of class-$\mathcal{KL}$ functions
  (see~\cite[Lemma~8]{Sontag1998Comments}).
  To provide insight into how this decomposition is obtained,
  Appendix~\ref{sec:kl_function_decomposition} gives a proof and
  constructs the functions $\kappa$ and $\xi$ using the notation of
  this paper.
  \begin{dfn}[Class-$\mathcal K_{\infty}$ function]
    \label{dfn:kinf_function}
    A function $\kappa:\R_{\geq0}\to\R_{\geq0}$ belongs to
    $\mathcal K_{\infty}$ if it is continuous, strictly increasing,
    satisfies $\kappa(0)=0$, and satisfies
    $\kappa(x)\to+\infty$ as $x\to+\infty$.
  \end{dfn}
  This decomposition gives $\kappa,\xi\in\mathcal K_{\infty}$
  (see Definition~\ref{dfn:kinf_function}) such that
  \begin{equation*}
    \beta(d,t)\leq\kappa(d)\xi(e^{-t})
    \qquad\text{for all }d\geq0\text{ and }t\geq0.
  \end{equation*}
  For the $L^1$ property considered here, a sufficient condition for
  Assumption~\ref{ass:lp_summable_tail} is the existence of constants
  $c>0$, $a>0$, and $x_0>0$ such that
  \begin{equation*}
    \xi(x)\leq c x^a
    \qquad\text{for every }x\in[0,x_0].
  \end{equation*}
  Under this condition, $\xi(e^{-k})\leq c e^{-ak}$ for all
  sufficiently large $k$, and therefore
  $\sum_{k=0}^{\infty}\xi(e^{-k})<+\infty$.
  The inequality
  $\beta(d,k)\leq\kappa(d)\xi(e^{-k})$ then gives
  \begin{equation*}
    \sum_{k=0}^{\infty}\beta(d,k)
    \leq
    \kappa(d)\sum_{k=0}^{\infty}\xi(e^{-k})<+\infty.
  \end{equation*}
  Since $\kappa(d)\to0$ as $d\downarrow0$, the same bound also gives
  \begin{equation*}
    \sum_{k=0}^{\infty}\beta(d,k)\xrightarrow{d\downarrow0}0.
  \end{equation*}

  \subsubsection{Assumption~\ref{ass:lp_critic}}
  \label{sec:valuebase_assumption_discussion}
  Assumption~\ref{ass:lp_critic} is primarily technical.
  It consists of two requirements: $\Valuebase$ must be continuous and
  the set $\mathbb{V}^{\basepolicy}(d^{\circ})$ defined in
  Assumption~\ref{ass:lp_critic} must be bounded.

  The continuity requirement is considered first.
  This requirement is satisfied by standard feedforward neural critics.
  A neural network composed of affine layers and continuous activation
  functions is continuous as a composition of continuous mappings.
  Networks with smooth activation functions are differentiable and
  therefore continuous.
  Networks with ReLU activations are continuous piecewise-affine functions.
  Output clipping and finite minima of continuous critic estimates are
  also continuous.
  Discontinuous neural critics require explicitly discontinuous
  components and are uncommon in RL practice.
  Consequently, continuity is a standard property of the neural critics
  used in RL implementations.

  The second requirement in Assumption~\ref{ass:lp_critic} is that
  $\mathbb{V}^{\basepolicy}(d^{\circ})$ be bounded.
  The set over which $v_{\min}(d^{\circ})$ is defined is compact
  because $\G$ is compact and $\states$ is finite-dimensional.
  Continuity of $\Valuebase$ therefore implies that $v_{\min}(d^{\circ})$ exists.
  Continuity also implies that $\mathbb{V}^{\basepolicy}(d^{\circ})$ is closed.
  Assumption~\ref{ass:lp_critic} requires this specific set to be bounded.

  There are several ways to ensure that
  $\mathbb{V}^{\basepolicy}(d^{\circ})$ is bounded.
  One way is to require $\Valuebase$ to have bounded superlevel sets.
  \begin{dfn}\label{dfn:bounded_superlevel_sets}
    Let \(X\) be a metric space and \(f \colon X \to \R\).
    The function \(f\) has \emph{bounded superlevel sets} if, for every
    \(a\in f(X)=\{f(x)\mid x\in X\}\), the set \(\{x\in X \mid f(x)\ge
    a\}\) is bounded in \(X\).
  \end{dfn}
  Although functions with bounded superlevel sets do not appear as a
  separate concept in the literature, this condition is closely related
  to radial unboundedness.
  \begin{dfn}[Radially unbounded function]
    A function $f:\R^n\to\R$ is said to be radially unbounded if
    $f(x)\to+\infty$ as $\|x\|\to+\infty$.
  \end{dfn}
  The relation is stated precisely in the following proposition.
  \begin{proposition}[Connection to radial unboundedness]
    \label{prop:radial_unboundedness}
    A function $f:\R^n\to\R$ has bounded superlevel sets if and only if
    either of the following conditions holds:
    \begin{enumerate}[label=(\roman*)]
      \item\label{item:radial_unboundedness} $-f$ is radially unbounded;
      \item\label{item:log_unboundedness} there exists $b\in\R$ such
        that $f(x)>b$ for every $x\in\R^n$ and $-\log(f(x)-b)$ is
        radially unbounded.
    \end{enumerate}
  \end{proposition}
  \begin{proof}
    See Appendix~\ref{sec:bounded_superlevel_sets}.
  \end{proof}

  Lyapunov critics and learned Lyapunov functions are used in safe RL
  to impose stability, reachability, or safety conditions on a learned
  policy~\cite{Han2020ActorCriticStability,Du2023SafeRobotCLBF}.
  The conditions imposed on these functions commonly include
  positivity, a decrease condition, and lower and upper comparison bounds.
  For example, the Lyapunov critics in
  \cite{Han2020ActorCriticStability,Du2023SafeRobotCLBF} are bounded
  below and above by positive multiples of a state cost.
  Some works require lower and upper bounds given by functions
  $\kappa_1,\kappa_2\in\mathcal{K}_{\infty}$
  (see Definition~\ref{dfn:kinf_function}).
  For example, the original CALF formulation requires
  $\kappa_1(\|s\|)\leq\hat Q_w(s,a)\leq\kappa_2(\|s\|)$ for its critic
  $\hat Q_w$~\cite{Osinenko2024CriticLyapunov}.
  Such lower and upper bounds imply that the critic is radially unbounded.

  CALF-Wrapper does not require the critic to satisfy these lower and
  upper bounds.
  It is sufficient for $\Valuebase$ to have bounded superlevel sets,
  which, by condition~\ref{item:log_unboundedness} of
  Proposition~\ref{prop:radial_unboundedness}, also permits critics
  that are bounded from below, as is the case, for example, when the
  rewards are uniformly nonnegative.
  The critic can also be modified only outside a prescribed bounded
  region, as stated in the following proposition.
  \begin{proposition}
    \label{prop:bounded_superlevel_sets_extension}
    Let $f:\R^n\to\R$ be continuous, and let $R\in\R_{>0}$.
    For every $b\in\left(-\infty,\min_{\|x\|\leq
    R}f(x)\right)\cup\{-\infty\}$, there exists a continuous function
    that equals $f$ on $\{x\mid\|x\|\leq R\}$, has bounded superlevel
    sets, and has infimum equal to $b$.
  \end{proposition}
  \begin{proof}
    See Appendix~\ref{sec:bounded_superlevel_sets}.
    The appendix gives explicit modifications of the critic in
    \eqref{eq:bounded_superlevel_extension_unbounded_infimum} for
    $b=-\infty$ and in
    \eqref{eq:bounded_superlevel_extension_finite_infimum} for finite $b$.
  \end{proof}

  The radius $R$ in
  Proposition~\ref{prop:bounded_superlevel_sets_extension} can be
  chosen from the data used to train the base policy and its critic.
  In practice, a finite training run can produce only finitely many
  sampled states for critic training.
  These sampled states form a bounded set.
  It is therefore natural to preserve the critic on a region that
  contains the states represented in the training data.
  The critic values at states outside this region are extrapolations
  that are not directly supported by sampled training states.
  Accordingly, $R$ can be chosen so that the ball $\{s\mid\|s\|\leq
  R\}$ contains every sampled training state.
  Outside this ball, the critic can be replaced by one of the
  continuous extensions given in
  \eqref{eq:bounded_superlevel_extension_unbounded_infimum} and
  \eqref{eq:bounded_superlevel_extension_finite_infimum}.
  Proposition~\ref{prop:bounded_superlevel_sets_extension} shows that
  the resulting critic agrees with the trained critic throughout the
  prescribed ball and has bounded superlevel sets.
  This modification does not require retraining the critic or changing
  the base policy.

  In some control environments, this modification is not needed.
  Such environments are constructed so that every possible trajectory
  lies entirely in a closed and bounded subset
  $\underline{\states}\subset\states$.
  Because $\states$ is finite-dimensional, $\underline{\states}$ is compact.
  In this case, the theorem can be formulated with
  $\underline{\states}$ as the state space instead of $\states$.
  This replacement does not change any step of the proof.
  The set $\mathbb{V}^{\basepolicy}(d^{\circ})$ is closed by continuity
  of $\Valuebase$ and is a subset of the compact state space
  $\underline{\states}$.
  Therefore, $\mathbb{V}^{\basepolicy}(d^{\circ})$ is bounded, so the
  second requirement in Assumption~\ref{ass:lp_critic} holds automatically.

  \gdef\uniformlqproofappendix{%
  \section{Proof of Theorem 2}
  \label{sec:proof_uniform_lp_bounds}

  The set
  \begin{equation*}
    \left\{\state\in\states:\goaldist(\state)\leq d^{\circ}\right\}
  \end{equation*}
  is nonempty and compact because $\G$ is compact and $\states$ is
  finite-dimensional.
  Continuity of $\Valuebase$ therefore implies that
  $v_{\min}(d^{\circ})$ exists and is finite.
  The set $\mathbb{V}^{\basepolicy}(d^{\circ})$ is nonempty and closed
  by continuity of $\Valuebase$, and it is bounded by
  Assumption~\ref{ass:lp_critic}.
  It is therefore compact.
  Consequently, $v_{\max}(d^{\circ})$ exists and is finite, and
  \begin{equation*}
    \sup_{\state\in\mathbb{V}^{\basepolicy}(d^{\circ})}
    \goaldist(\state)<+\infty.
  \end{equation*}
  The function $\bar p$ is upper semicontinuous, and $\actions$ is
  compact.
  Hence, $d_{\bar p}(d^{\circ})$ is finite.
  The definitions in \eqref{eq:lp_critic_count} and
  \eqref{eq:lp_fallback_initial_bound} then show that
  $\tau(d^{\circ})$ and $\delta(d^{\circ})$ are finite.

  The series $\sum_{k=0}^{\infty}\bar\rho_k$ is finite by the
  conditions of Algorithm~\ref{alg_calfwrapper}.
  Therefore, the product defining $\mathcal P(t)$ exists, and
  $\mathcal P(t)\in[0,1]$.
  The quantities $C(d^{\circ})$ and $H(d^{\circ},t,\ell)$ are also
  finite.
  Assumption~\ref{ass:lp_uniform_fallback} implies that
  $\beta(\delta(d^{\circ}),0)$ is finite.
  Thus, $\Delta(d^{\circ})$ is finite.
  The convention $\inf\varnothing=+\infty$ makes
  $T_{\mathrm{hit}}(\state_0,d^{*})$ and
  $T_{\mathrm{set}}(\state_0,d^{*})$ well defined with values in
  $[0,+\infty]$ for every initial state $\state_0\in\states$.
  The nonnegative series
  $\sum_{k=0}^{\infty}\beta(\underline{d}^{*},k)$ is well defined
  with a value in $[0,+\infty]$.
  Assumption~\ref{ass:lp_summable_tail} and monotonicity of
  $\beta(\bullet,k)$ make this series finite for every
  $\underline{d}^{*}\in(0,d^{*}]$.

  \begin{lemma}[States at which the base policy is selected]
    \label{lem:lp_base_policy_states}
    Suppose that Assumption~\ref{ass:lp_acceptance} holds.
    Whenever Algorithm~\ref{alg_calfwrapper} selects the base policy
    at time $t$,
    \begin{equation}
      \State_t\in\mathbb{V}^{\basepolicy}(d^{\circ}),
      \qquad
      \goaldist(\State_t)\leq\delta(d^{\circ}),
      \label{eq:lp_base_policy_state_bound}
    \end{equation}
    and
    \begin{equation}
      \goaldist(\State_{t+1})\leq\delta(d^{\circ})
      \quad\text{almost surely}.
      \label{eq:lp_base_policy_next_state_bound}
    \end{equation}
  \end{lemma}

  \begin{proof}
    If the base policy is selected because the critic condition holds, then
    \begin{equation*}
      \Valuebase(\State_t)
      \geq\Value_t^{\dagger}+\nu
      \geq\Valuebase(\state_0)
      \geq v_{\min}(d^{\circ}).
    \end{equation*}
    If the base policy is selected because the random acceptance
    condition holds, then
    $U_t<\rho_t(\State_t)$.
    Assumption~\ref{ass:lp_acceptance} then implies
    $\Valuebase(\State_t)\geq\Valuebase(\state_0)$.
    Thus, $\State_t\in\mathbb{V}^{\basepolicy}(d^{\circ})$ in both cases.
    Equation~\eqref{eq:lp_base_policy_state_bound} follows from the
    definition of $\delta(d^{\circ})$.
    The one-step transition bound gives
    \begin{equation*}
      \goaldist(\State_{t+1})
      \leq\|\State_{t+1}\|
      \leq\bar p(\State_t,\Action_t)
      \leq d_{\bar p}(d^{\circ})
      \leq\delta(d^{\circ})
    \end{equation*}
    almost surely, which proves \eqref{eq:lp_base_policy_next_state_bound}.
  \end{proof}

  \begin{lemma}[Number of selections based on the critic condition]
    \label{lem:lp_critic_selection_count}
    Suppose that Assumption~\ref{ass:lp_critic} holds.
    The number of times at which the base policy is selected because
    the critic condition holds satisfies
    \begin{equation}
      N_{\mathrm{crit}}
      :=\sum_{t=0}^{\infty}
      \mathbf{1}\!\left\{
        \State_t\notin\states(\fallbackpolicy),
        \Valuebase(\State_t)\geq\Value_t^{\dagger}+\nu
      \right\}
      \leq\tau(d^{\circ}).
      \label{eq:lp_critic_selection_count_bound}
    \end{equation}
  \end{lemma}

  \begin{proof}
    Every selection based on the critic condition counted by
    $N_{\mathrm{crit}}$ increases $\Value_t^{\dagger}$ by at least
    $\nu$.
    At every such time, $\Valuebase(\State_t)\geq\Valuebase(\state_0)
    \geq v_{\min}(d^{\circ})$.
    Hence, the corresponding state belongs to
    $\mathbb{V}^{\basepolicy}(d^{\circ})$, where the critic value is
    at most $v_{\max}(d^{\circ})$.
    The stored value is initially at least $v_{\min}(d^{\circ})$.
    Therefore,
    \begin{equation*}
      N_{\mathrm{crit}}
      \leq\tau(d^{\circ})
      =\max\left\{1,
        \left\lfloor
        \frac{v_{\max}(d^{\circ})-v_{\min}(d^{\circ})}{\nu}
        \right\rfloor
      \right\}.
    \end{equation*}
  \end{proof}

  \begin{lemma}[Total number of base-policy selections]
    \label{lem:lp_base_selection_count}
    Let
    \begin{equation*}
      N_{\basepolicy}
      :=\sum_{t=0}^{\infty}
      \mathbf 1\{\basepolicy\text{ is selected at time }t\}.
    \end{equation*}
    Suppose that Assumptions~\ref{ass:lp_acceptance}
    and~\ref{ass:lp_critic} hold.
    Then
    \begin{equation}
      \E{N_{\basepolicy}}
      \leq C(d^{\circ})-1,
      \qquad
      \PP{N_{\basepolicy}<+\infty}=1.
      \label{eq:lp_base_selection_count_bound}
    \end{equation}
  \end{lemma}

  \begin{proof}
    Define
    \begin{equation*}
      N_{\rho}:=\sum_{t=0}^{\infty}
      \mathbf{1}\!\left\{
        \State_t\notin\states(\fallbackpolicy),
        U_t<\rho_t(\State_t)
      \right\}.
    \end{equation*}
    Every base-policy selection satisfies at least one of the two
    conditions in line~\ref{lst:line:selection_conditions} of
    Algorithm~\ref{alg_calfwrapper}.
    Therefore,
    \begin{equation*}
      N_{\basepolicy}\leq N_{\mathrm{crit}}+N_{\rho}.
    \end{equation*}
    Since $\rho_t(\State_t)\leq\bar\rho_t$,
    \begin{equation*}
      \E{N_{\rho}}
      =\sum_{t=0}^{\infty}
      \PP{\State_t\notin\states(\fallbackpolicy),
      U_t<\rho_t(\State_t)}
      \leq\sum_{t=0}^{\infty}\bar\rho_t.
    \end{equation*}
    Lemma~\ref{lem:lp_critic_selection_count} and
    \eqref{eq:lp_expected_candidate_count} give
    \begin{equation*}
      \E{N_{\basepolicy}}
      \leq\tau(d^{\circ})+\sum_{t=0}^{\infty}\bar\rho_t
      =C(d^{\circ})-1<+\infty.
    \end{equation*}
    A nonnegative integer-valued random variable with finite
    expectation is finite almost surely.
    This proves \eqref{eq:lp_base_selection_count_bound}.
  \end{proof}

  \begin{lemma}[A consecutive sequence of fallback-policy selections]
    \label{lem:lp_fallback_block}
    Suppose that Assumption~\ref{ass:lp_critic} holds.
    By Lemma~\ref{lem:lp_critic_selection_count}, the number of
    selections of the base policy based on the critic condition is at
    most $\tau(d^{\circ})$.
    On the event
    \begin{equation*}
      U_k\geq\bar\rho_k
      \quad\text{for every }k\geq t,
    \end{equation*}
    the first $H(d^{\circ},t,\ell)$ actions, where
    $H(d^{\circ},t,\ell)=\bigl(\tau(d^{\circ})+t+1\bigr)\ell$,
    contain at least $\ell$ consecutive actions supplied by the
    fallback policy.
  \end{lemma}

  \begin{proof}
    Before time $t$, there are at most $t$ action positions.
    After time $t$, the stated event excludes every selection based on
    the random acceptance condition because
    $\rho_k(\State_k)\leq\bar\rho_k$.
    There are at most $\tau(d^{\circ})$ selections of the base policy
    based on the critic condition.
    Thus, the first $H(d^{\circ},t,\ell)$ actions contain at most
    $m:=\tau(d^{\circ})+t$ actions supplied by the base policy.
    The at most $m$ actions supplied by the base policy divide the
    remaining actions into at most $m+1$ consecutive sequences of
    actions supplied by the fallback policy.
    By the pigeonhole principle, at least one of these sequences
    contains at least $\ell$ actions.
    Otherwise, the total number of actions would be at most
    $m+(m+1)(\ell-1)=(m+1)\ell-1$, which is less than
    $(m+1)\ell$.
    Substitution of $m=\tau(d^{\circ})+t$ proves the claim.
  \end{proof}

  \begin{lemma}[Expected-distance bound during consecutive fallback-policy selections]
    \label{lem:lp_fallback_sequence_bound}
    Suppose that Assumption~\ref{ass:lp_uniform_fallback} holds.
    Let $\sigma$ be a stopping time at which
    $\goaldist(\State_{\sigma})\leq\delta(d^{\circ})$.
    Consider the state process generated by the fallback policy from
    $\State_{\sigma}$.
    For every $\ell\in\mathbb{Z}_{\geq0}$,
    \begin{equation}
      \E{
        \goaldist\bigl(
          \State_{\ell}^{\fallbackpolicy}(\State_{\sigma})
        \bigr)
        \mid\State_{\sigma}
      }
      \leq\beta(\delta(d^{\circ}),\ell).
      \label{eq:lp_fallback_sequence_moment_bound}
    \end{equation}
    Consequently, for every $\underline{d}^{*}>0$,
    \begin{equation}
      \PP{
        \goaldist\bigl(
          \State_{\ell}^{\fallbackpolicy}(\State_{\sigma})
        \bigr)>\underline{d}^{*}
        \mid\State_{\sigma}
      }
      \leq
      \left(
        \frac{\beta(\delta(d^{\circ}),\ell)}{\underline{d}^{*}}
      \right).
      \label{eq:lp_fallback_sequence_probability_bound}
    \end{equation}
    If Algorithm~\ref{alg_calfwrapper} selects the fallback policy at
    times $\sigma,\ldots,\sigma+\ell-1$, then its state at time
    $\sigma+\ell$ coincides with the state in
    \eqref{eq:lp_fallback_sequence_moment_bound} under the same
    realizations of the fallback-policy actions and system transitions.
  \end{lemma}

  \begin{proof}
    The strong Markov property,
    Definition~\ref{dfn:uniform_lp_goal_reaching}, and monotonicity
    of $\beta$ in its first argument give
    \eqref{eq:lp_fallback_sequence_moment_bound}.
    Markov's inequality gives \eqref{eq:lp_fallback_sequence_probability_bound}.
    The final statement follows because both state processes apply
    the fallback policy for the stated $\ell$ time steps.
  \end{proof}

  \begin{lemma}[Probability of a subsequent exit under the fallback policy]
    \label{lem:lp_local_fallback_exit}
    Suppose that Assumption~\ref{ass:lp_uniform_fallback} holds.
    Let $0<\underline{d}^{*}\leq d^{*}$, and let $\sigma$ be a
    stopping time such that
    $\goaldist(\State_{\sigma})\leq\underline{d}^{*}$.
    For the state process generated by the fallback policy from
    $\State_{\sigma}$,
    \begin{multline}
      \PP{
        \goaldist\bigl(
          \State_k^{\fallbackpolicy}(\State_{\sigma})
        \bigr)>d^{*}
        \text{ for some }k\in\mathbb{Z}_{\geq0}
        \mid\State_{\sigma}
      }
      \\
      \leq
      \frac{1}{d^{*}}
      \sum_{k=0}^{\infty}\beta(\underline{d}^{*},k).
      \label{eq:lp_local_fallback_exit_bound}
    \end{multline}
  \end{lemma}

  \begin{proof}
    By the strong Markov property, conditionally on
    $\State_{\sigma}$, the state process after time $\sigma$ has the
    same transition law as the state process generated by the fallback
    policy and initialized at $\State_{\sigma}$.
    Definition~\ref{dfn:uniform_lp_goal_reaching} therefore gives
    \begin{equation*}
      \E{
        \goaldist\bigl(
          \State_k^{\fallbackpolicy}(\State_{\sigma})
        \bigr)
        \mid\State_{\sigma}
      }
      \leq\beta\bigl(\goaldist(\State_{\sigma}),k\bigr)
      \leq\beta(\underline{d}^{*},k)
    \end{equation*}
    for every $k\in\mathbb{Z}_{\geq0}$, where the second inequality
    follows from
    $\goaldist(\State_{\sigma})\leq\underline{d}^{*}$ and
    monotonicity of $\beta$ in its first argument.
    Conditional Markov's inequality then gives
    \begin{multline*}
      \PP{
        \goaldist\bigl(
          \State_k^{\fallbackpolicy}(\State_{\sigma})
        \bigr)>d^{*}
        \mid\State_{\sigma}
      }
      \\
      \leq
      \frac{1}{d^{*}}
      \E{
        \goaldist\bigl(
          \State_k^{\fallbackpolicy}(\State_{\sigma})
        \bigr)
        \mid\State_{\sigma}
      }
      \leq\frac{\beta(\underline{d}^{*},k)}{d^{*}}.
    \end{multline*}
    Finally, the countable union bound gives
    \begin{align*}
      &\PP{
        \goaldist\bigl(
          \State_k^{\fallbackpolicy}(\State_{\sigma})
        \bigr)>d^{*}
        \text{ for some }k\in\mathbb{Z}_{\geq0}
        \mid\State_{\sigma}
      }
      \\
      &\quad\leq
      \sum_{k=0}^{\infty}
      \PP{
        \goaldist\bigl(
          \State_k^{\fallbackpolicy}(\State_{\sigma})
        \bigr)>d^{*}
        \mid\State_{\sigma}
      }
      \\
      &\quad\leq
      \frac{1}{d^{*}}
      \sum_{k=0}^{\infty}\beta(\underline{d}^{*},k),
    \end{align*}
    which proves \eqref{eq:lp_local_fallback_exit_bound}.
  \end{proof}

  \begin{proof}
    Let $\mathcal F_t$ contain the history through the state
    $\State_t$ and before the policy is selected at time $t$.
    The initial-state condition gives
    $\goaldist(\state_0)\leq\delta(d^{\circ})$.
    Lemma~\ref{lem:lp_base_policy_states} gives
    \begin{equation}
      \goaldist(\State_{k+1})\leq\delta(d^{\circ})
      \label{eq:lp_proof_fallback_start_bound}
    \end{equation}
    almost surely whenever Algorithm~\ref{alg_calfwrapper} selects
    the base policy at time $k$.
    Thus, the initial state and every state immediately following an
    action supplied by the base policy satisfy the same bound.

    \textit{Claim~\ref{claim:lp_uniform_moment}.}
    If $\fallbackpolicy$ is selected at time $t$, then the current
    consecutive selections of the fallback policy begin either at
    time zero or immediately after an earlier selection of the base
    policy.
    Partitioning according to the most recent time before $t$ at
    which $\basepolicy$ is selected gives
    \begin{align*}
      \mathbf{1}\!\left\{
          \substack{
            \fallbackpolicy\text{ is selected}\\
            \text{at time }t
          }
        \right\}
      &=
        \mathbf{1}\!\left\{
          \substack{
            \fallbackpolicy\text{ is selected at every time }j\\
            0\leq j\leq t
          }
        \right\}
      \\
      &\quad+
      \sum_{k=0}^{t-1}
        \mathbf{1}\!\left\{
          \substack{
            \basepolicy\text{ is selected at time }k,\\
            \fallbackpolicy\text{ is selected at every time }j,\\
            k<j\leq t
          }
        \right\}.
    \end{align*}
    Multiplying this identity by
    $\goaldist(\State_t^{\calfwpolicy}(\state_0))$ and taking
    expectations gives the corresponding decomposition of the
    expectation.
    On the first event, the state at time $t$ is generated by $t$
    actions supplied by $\fallbackpolicy$ from $\state_0$.
    On every event in the sum, the state at time $t$ is generated by
    $t-k-1$ actions supplied by $\fallbackpolicy$ from $\State_{k+1}$.
    Dropping the restrictions on the policy selections after time
    $k$ and using nonnegativity give
    \begin{align*}
      &\E{
        \goaldist\bigl(\State_t^{\calfwpolicy}(\state_0)\bigr)
        \mathbf{1}\!\left\{
          \substack{
            \fallbackpolicy\text{ is selected}\\
            \text{at time }t
          }
        \right\}
      }
      \\
      &\quad\leq
      \E{
        \goaldist\bigl(
          \State_t^{\fallbackpolicy}(\state_0)
        \bigr)
      }
      \\
      &\qquad+
      \sum_{k=0}^{t-1}
      \E{
        \begin{aligned}
          &\mathbf{1}\!\left\{
            \substack{
              \basepolicy\text{ is selected}\\
              \text{at time }k
            }
          \right\}
          \\
          &\quad\times
          \E{
            \goaldist\bigl(
              \State_{t-k-1}^{\fallbackpolicy}(\State_{k+1})
            \bigr)
            \mid\mathcal F_{k+1}
          }
        \end{aligned}
      }.
    \end{align*}
    Lemma~\ref{lem:lp_fallback_sequence_bound},
    \eqref{eq:lp_proof_fallback_start_bound}, and monotonicity of
    $\beta$ in its first argument then give
    \begin{align*}
      &\E{
        \goaldist\bigl(\State_t^{\calfwpolicy}(\state_0)\bigr)
        \mathbf{1}\!\left\{
          \substack{
            \fallbackpolicy\text{ is selected}\\
            \text{at time }t
          }
        \right\}
      }
      \\
      &\quad\leq
      \beta(\delta(d^{\circ}),t)
      \\
      &\qquad+
      \E{
        \sum_{k=0}^{t-1}
        \mathbf{1}\!\left\{
          \substack{
            \basepolicy\text{ is selected}\\
            \text{at time }k
          }
        \right\}
        \beta(\delta(d^{\circ}),t-k-1)
      }
      \\
      &\quad\leq
      \beta(\delta(d^{\circ}),0)
      \left(
        1+
        \E{
          \sum_{k=0}^{t-1}
          \mathbf{1}\!\left\{
            \substack{
              \basepolicy\text{ is selected}\\
              \text{at time }k
            }
          \right\}
        }
      \right)
      \\
      &\quad\leq
      C(d^{\circ})\beta(\delta(d^{\circ}),0).
    \end{align*}
    For the complementary policy-selection case,
    Lemma~\ref{lem:lp_base_policy_states} gives
    \begin{equation*}
      \E{
        \goaldist\bigl(\State_t^{\calfwpolicy}(\state_0)\bigr)
        \mathbf{1}\{\basepolicy\text{ is selected at time }t\}
      }
      \leq\delta(d^{\circ}).
    \end{equation*}
    Consequently,
    \begin{equation*}
      \E{
        \goaldist\bigl(\State_t^{\calfwpolicy}(\state_0)\bigr)
      }
      \leq
      \delta(d^{\circ})
      +C(d^{\circ})\beta(\delta(d^{\circ}),0)
      =\Delta(d^{\circ}).
    \end{equation*}
    Taking the supremum over $t$ proves
    \eqref{eq:lp_uniform_moment_bound}.
    Moreover,
    \begin{align*}
      &\PP{
        \max_{0\leq k\leq T}
        \goaldist\bigl(\State_k^{\calfwpolicy}(\state_0)\bigr)>R
      }
      \leq
      \sum_{k=0}^{T}
      \PP{
        \goaldist\bigl(\State_k^{\calfwpolicy}(\state_0)\bigr)>R
      }
      \\
      &\leq
      \frac{1}{R}
      \sum_{k=0}^{T}
      \E{
        \goaldist\bigl(\State_k^{\calfwpolicy}(\state_0)\bigr)
      }
      \leq
      (T+1)\frac{\Delta(d^{\circ})}{R},
    \end{align*}
    which proves \eqref{eq:lp_prefix_overshoot_bound}.

    \textit{Claim~\ref{claim:lp_hitting_bound}.}
    Define
    \begin{equation*}
      \mathbb{U}_t:=\bigcap_{k=t}^{\infty}\{U_k\geq\bar\rho_k\}.
    \end{equation*}
    Independence of the uniform random variables gives
    \begin{equation}
      \PP{\mathbb{U}_t}
      =\prod_{k=t}^{\infty}(1-\bar\rho_k)
      =\mathcal P(t).
      \label{eq:lp_proof_no_random_acceptance}
    \end{equation}
    For any state $\state$ satisfying
    $\goaldist(\state)\leq\delta(d^{\circ})$, define
    \begin{equation*}
      \mathbb{D}_{\ell}(\state)
      :=
      \left\{
        \goaldist\bigl(\State_{\ell}^{\fallbackpolicy}(\state)\bigr)
        >d^{*}
      \right\}.
    \end{equation*}
    Equation~\eqref{eq:lp_fallback_sequence_probability_bound} gives
    \begin{equation*}
      \PP{\mathbb{D}_{\ell}(\state_0)}
      \leq
      \frac{\beta(\delta(d^{\circ}),\ell)}{d^{*}}.
    \end{equation*}
    For every $k\geq0$, the same equation and
    \eqref{eq:lp_proof_fallback_start_bound} give
    \begin{align*}
      &\E{
        \begin{aligned}
          &\mathbf{1}\{\basepolicy\text{ is selected at time }k\}
          \\
          &\quad\times
          \PP{
            \mathbb{D}_{\ell}(\State_{k+1})
            \mid\mathcal F_{k+1}
          }
        \end{aligned}
      }
      \\
      &\leq
      \PP{\basepolicy\text{ is selected at time }k}
      \frac{\beta(\delta(d^{\circ}),\ell)}{d^{*}}
    \end{align*}
    Define the event
    \begin{multline*}
      \mathbb{D}_{\ell}
      :=
      \mathbb{D}_{\ell}(\state_0)
      \\
      {}\cup
      \bigcup_{k=0}^{\infty}
      \left(
        \{\basepolicy\text{ is selected at time }k\}
        \cap
        \mathbb{D}_{\ell}(\State_{k+1})
      \right).
    \end{multline*}
    The event that the base policy is selected at time $k$ is
    $\mathcal F_{k+1}$-measurable.
    Therefore, for every $k\geq0$,
    \begin{multline*}
      \PP{
        \{\basepolicy\text{ is selected at time }k\}
        \cap
        \mathbb{D}_{\ell}(\State_{k+1})
      }
      \\
      =
      \E{
        \mathbf 1\{\basepolicy\text{ is selected at time }k\}
        \PP{
          \mathbb{D}_{\ell}(\State_{k+1})
          \mid\mathcal F_{k+1}
        }
      }.
    \end{multline*}
    The union bound, the conditional-expectation identity, and
    \eqref{eq:lp_base_selection_count_bound} therefore give
    \begin{equation}
    \begin{aligned}
      &\PP{\mathbb{D}_{\ell}}
      \\
      &\quad\leq
      \PP{\mathbb{D}_{\ell}(\state_0)}
      \\
      &\quad+
      \sum_{k=0}^{\infty}
      \E{
        \begin{aligned}
          &\mathbf 1\{\basepolicy\text{ is selected at time }k\}
          \\
          &\quad\times
          \PP{
            \mathbb{D}_{\ell}(\State_{k+1})
            \mid\mathcal F_{k+1}
          }
        \end{aligned}
      }
      \\
      &\quad\leq
      C(d^{\circ})
      \frac{\beta(\delta(d^{\circ}),\ell)}{d^{*}}.
      \label{eq:lp_proof_candidate_hitting_failure}
    \end{aligned}
    \end{equation}
    On $\mathbb{U}_t$, no selection of the base policy based on the
    random acceptance condition occurs at or after time $t$.
    Lemmas~\ref{lem:lp_critic_selection_count}
    and~\ref{lem:lp_fallback_block} therefore show that the first
    $H(d^{\circ},t,\ell)$ actions contain at least $\ell$ consecutive
    actions supplied by the fallback policy.
    These consecutive actions begin either at the initial state or at
    a state immediately following a selection of the base policy.
    If $T_{\mathrm{hit}}(\state_0,d^{*})>H(d^{\circ},t,\ell)$ also
    holds, then the state after these $\ell$ consecutive actions has
    distance greater than $d^{*}$ from $\G$.
    Consequently, the event
    \begin{equation*}
      \mathbb{U}_t
      \cap
      \{T_{\mathrm{hit}}(\state_0,d^{*})>H(d^{\circ},t,\ell)\}
    \end{equation*}
    is contained in $\mathbb{D}_{\ell}$.
    Hence,
    \begin{align*}
      &\PP{T_{\mathrm{hit}}(\state_0,d^{*})
      \leq H(d^{\circ},t,\ell)}
      \\
      &\quad\geq
      \PP{
        \mathbb{U}_t
        \cap
        \{T_{\mathrm{hit}}(\state_0,d^{*})
        \leq H(d^{\circ},t,\ell)\}
      }
      \\
      &\quad=
      \PP{\mathbb{U}_t}
      -
      \PP{
        \mathbb{U}_t
        \cap
        \{T_{\mathrm{hit}}(\state_0,d^{*})
        >H(d^{\circ},t,\ell)\}
      }
      \\
      &\quad\geq
      \mathcal P(t)
      -C(d^{\circ})
      \frac{\beta(\delta(d^{\circ}),\ell)}{d^{*}},
    \end{align*}
    which proves \eqref{eq:lp_first_hitting_bound}.
    Since
    \begin{equation*}
      \mathcal P(t)\xrightarrow{t\to\infty}1,
      \qquad
      \beta(\delta(d^{\circ}),\ell)\xrightarrow{\ell\to\infty}0,
    \end{equation*}
    \begin{multline*}
      \PP{T_{\mathrm{hit}}(\state_0,d^{*})<+\infty}
      \\
      \geq
      \sup_{\substack{t\in\mathbb Z_{\geq0}\\
      \ell\in\mathbb Z_{>0}}}
      \left[
        \mathcal P(t)
        -C(d^{\circ})
        \frac{\beta(\delta(d^{\circ}),\ell)}{d^{*}}
      \right]
      =1.
    \end{multline*}
    This proves \eqref{eq:lp_finite_hitting_time}.

    \textit{Claim~\ref{claim:lp_reaching_bound}.}
    For any state $\state$ satisfying
    $\goaldist(\state)\leq\delta(d^{\circ})$, define
    \begin{multline*}
      \mathbb{D}_{\ell}^{*}(\state)
      :=
      \{\goaldist(\State_{\ell}^{\fallbackpolicy}(\state))
      >\underline{d}^{*}\}
      \\
      {}\cup
      \left(
        \{\goaldist(\State_{\ell}^{\fallbackpolicy}(\state))
        \leq\underline{d}^{*}\}
        \cap
        \left\{
          \sup_{j\geq\ell}
          \goaldist(\State_j^{\fallbackpolicy}(\state))>d^{*}
        \right\}
      \right).
    \end{multline*}
    Lemmas~\ref{lem:lp_fallback_sequence_bound}
    and~\ref{lem:lp_local_fallback_exit} give
    \begin{multline*}
      \PP{\mathbb{D}_{\ell}^{*}(\state)}
      \leq
      \frac{\beta(\delta(d^{\circ}),\ell)}{\underline{d}^{*}}
      +\frac{1}{d^{*}}
      \sum_{k=0}^{\infty}\beta(\underline{d}^{*},k).
    \end{multline*}
    Lemma~\ref{lem:lp_base_policy_states} gives
    $\goaldist(\State_{k+1})\leq\delta(d^{\circ})$ almost surely on
    the event that the base policy is selected at time $k$.
    Conditional application of the preceding bound therefore gives,
    on this event,
    \begin{equation*}
      \PP{\mathbb{D}_{\ell}^{*}(\State_{k+1})\mid\mathcal F_{k+1}}
      \leq
      \frac{\beta(\delta(d^{\circ}),\ell)}{\underline{d}^{*}}
      +\frac{1}{d^{*}}
      \sum_{j=0}^{\infty}\beta(\underline{d}^{*},j).
    \end{equation*}
    Define the event
    \begin{multline*}
      \mathbb{D}_{\ell}^{*}
      :=
      \mathbb{D}_{\ell}^{*}(\state_0)
      \\
      {}\cup
      \bigcup_{k=0}^{\infty}
      \left(
        \{\basepolicy\text{ is selected at time }k\}
        \cap
        \mathbb{D}_{\ell}^{*}(\State_{k+1})
      \right).
    \end{multline*}
    The same conditional-expectation identity used for
    $\mathbb{D}_{\ell}$ and the union bound give
    \begin{equation*}
    \begin{aligned}
      &\PP{\mathbb{D}_{\ell}^{*}}
      \\
      &\quad\leq
      \PP{\mathbb{D}_{\ell}^{*}(\state_0)}
      \\
      &\quad+
      \sum_{k=0}^{\infty}
      \E{
        \begin{aligned}
          &\mathbf 1\{\basepolicy\text{ is selected at time }k\}
          \\
          &\quad\times
          \PP{
            \mathbb{D}_{\ell}^{*}(\State_{k+1})
            \mid\mathcal F_{k+1}
          }
        \end{aligned}
      }
      \\
      &\quad\leq
      \left(
        1+
        \sum_{k=0}^{\infty}
        \E{\mathbf 1\{\basepolicy\text{ is selected at time }k\}}
      \right)
      \\
      &\qquad\times
      \left[
        \frac{\beta(\delta(d^{\circ}),\ell)}{\underline{d}^{*}}
        +\frac{1}{d^{*}}
        \sum_{j=0}^{\infty}\beta(\underline{d}^{*},j)
      \right].
    \end{aligned}
    \end{equation*}
    The definition of $N_{\basepolicy}$ and
    \eqref{eq:lp_base_selection_count_bound} give
    \begin{equation}
    \begin{aligned}
      &\left(
        1+
        \sum_{k=0}^{\infty}
        \E{\mathbf 1\{\basepolicy\text{ is selected at time }k\}}
      \right)
      \\
      &\qquad\times
      \left[
        \frac{\beta(\delta(d^{\circ}),\ell)}{\underline{d}^{*}}
        +\frac{1}{d^{*}}
        \sum_{j=0}^{\infty}\beta(\underline{d}^{*},j)
      \right]
      \\
      &\quad=
      \left(1+\E{N_{\basepolicy}}\right)
      \left[
        \frac{\beta(\delta(d^{\circ}),\ell)}{\underline{d}^{*}}
        +\frac{1}{d^{*}}
        \sum_{j=0}^{\infty}\beta(\underline{d}^{*},j)
      \right]
      \\
      &\quad\leq
      C(d^{\circ})
      \left[
        \frac{\beta(\delta(d^{\circ}),\ell)}{\underline{d}^{*}}
        +\frac{1}{d^{*}}
        \sum_{j=0}^{\infty}\beta(\underline{d}^{*},j)
      \right].
      \label{eq:lp_proof_candidate_reaching_failure}
    \end{aligned}
    \end{equation}
    Fix $\underline{d}^{*}\in(0,d^{*}]$ and suppose that
    Assumption~\ref{ass:lp_fallback_locking} holds.
    On $\mathbb{U}_t$, Lemmas~\ref{lem:lp_critic_selection_count}
    and~\ref{lem:lp_fallback_block} provide at least $\ell$
    consecutive actions supplied by the fallback policy among the
    first $H(d^{\circ},t,\ell)$ actions.
    These actions begin either at the initial state or at a state
    immediately following a selection of the base policy.
    If $\mathbb{D}_{\ell}^{*}$ does not occur, the state after these
    $\ell$ actions has distance at most $\underline{d}^{*}$ from
    $\G$, and the trajectory generated by the fallback policy from
    that state remains within distance $d^{*}$ of $\G$.
    Since $\underline{d}^{*}\leq d^{*}$,
    Assumption~\ref{ass:lp_fallback_locking} requires
    Algorithm~\ref{alg_calfwrapper} to continue selecting the
    fallback policy.
    The trajectory generated by CALF-Wrapper therefore remains
    within distance $d^{*}$ of $\G$ from that time onward.
    Consequently,
    \begin{equation*}
      \mathbb{U}_t
      \cap
      \{T_{\mathrm{set}}(\state_0,d^{*})>H(d^{\circ},t,\ell)\}
      \subseteq
      \mathbb{D}_{\ell}^{*}.
    \end{equation*}
    Equations~\eqref{eq:lp_proof_no_random_acceptance}
    and~\eqref{eq:lp_proof_candidate_reaching_failure} then give
    \begin{align*}
      &\PP{T_{\mathrm{set}}(\state_0,d^{*})
      \leq H(d^{\circ},t,\ell)}
      \\
      &\quad\geq
      \PP{
        \mathbb{U}_t
        \cap
        \{T_{\mathrm{set}}(\state_0,d^{*})
        \leq H(d^{\circ},t,\ell)\}
      }
      \\
      &\quad=
      \PP{\mathbb{U}_t}
      -
      \PP{
        \mathbb{U}_t
        \cap
        \{T_{\mathrm{set}}(\state_0,d^{*})
        >H(d^{\circ},t,\ell)\}
      }
      \\
      &\quad\geq
      \mathcal P(t)-\PP{\mathbb{D}_{\ell}^{*}}
      \\
      &\quad\geq
      \mathcal P(t)
      -C(d^{\circ})\left[
        \frac{\beta(\delta(d^{\circ}),\ell)}{\underline{d}^{*}}
        +\frac{1}{d^{*}}
        \sum_{k=0}^{\infty}\beta(\underline{d}^{*},k)
      \right],
    \end{align*}
    which proves \eqref{eq:lp_reaching_time_bound}.
    Let $\epsilon>0$.
    Assumption~\ref{ass:lp_summable_tail} permits the successive
    selection of $\underline{d}^{*}$, $\ell$, and $t$ such that
    \begin{equation*}
      \frac{C(d^{\circ})}{d^{*}}
      \sum_{k=0}^{\infty}\beta(\underline{d}^{*},k)
      <\frac{\epsilon}{3},
    \end{equation*}
    \begin{equation*}
      C(d^{\circ})
      \frac{\beta(\delta(d^{\circ}),\ell)}{\underline{d}^{*}}
      <\frac{\epsilon}{3},
      \qquad
      1-\mathcal P(t)<\frac{\epsilon}{3}.
    \end{equation*}
    Substitution into \eqref{eq:lp_reaching_time_bound} gives
    \begin{equation*}
      \PP{T_{\mathrm{set}}(\state_0,d^{*})
      \leq H(d^{\circ},t,\ell)}>1-\epsilon.
    \end{equation*}
    Since $\epsilon$ is arbitrary,
    $\PP{T_{\mathrm{set}}(\state_0,d^{*})<+\infty}=1$.
    This proves the final statement of
    Claim~\ref{claim:lp_reaching_bound}.

  \end{proof}

  \subsection{Explicit Probability Bound for Geometrically Decaying
  Random Acceptance Probabilities}
  \label{sec:explicit_probability_bound}
  The probability bounds in Claims~\ref{claim:lp_hitting_bound} and
  \ref{claim:lp_reaching_bound} contain
  $\mathcal P(t)=\prod_{k=t}^{\infty}(1-\bar\rho_k)$.
  When $\bar\rho_k=\lambda^k p_{\mathrm{relax}}$, the following
  proposition gives an explicit lower bound on $\mathcal P(t)$.

  \begin{proposition}
    \label{prop:relaxation_product_bound}
    Suppose that $\bar\rho_k=\lambda^k p_{\mathrm{relax}}$ for every
    $k\in\mathbb{Z}_{\geq0}$, where $\lambda\in(0,1)$ and
    $p_{\mathrm{relax}}\in[0,1]$.
    Then, for every $t\in\mathbb{Z}_{\geq1}$,
    \begin{equation}
      \mathcal P(t)
      \geq
      \exp\!\left(
        \frac{-\lambda^t p_{\mathrm{relax}}}
        {(1-\lambda)(1-\lambda^t p_{\mathrm{relax}})}
      \right).
      \label{eq:relaxation_product_lower_bound}
    \end{equation}
  \end{proposition}

  \begin{proof}
    Under the stated choice of $\bar\rho_k$,
    \begin{equation*}
      \mathcal P(t)
      =\exp\left(
        \sum_{k=t}^{\infty}
        \log(1-\lambda^k p_{\mathrm{relax}})
      \right).
    \end{equation*}
    For every $x\in[0,1)$, $\log(1-x)\geq-x/(1-x)$.
    The function $\log(1-x)+x/(1-x)$ is zero at $x=0$ and has a
    nonnegative derivative on $[0,1)$.
    Therefore, for every $t\in\mathbb{Z}_{\geq1}$,
    \begin{align*}
      \sum_{k=t}^{\infty}\log(1-\lambda^k p_{\mathrm{relax}})
      &\geq
      \sum_{k=t}^{\infty}
      \frac{-\lambda^k p_{\mathrm{relax}}}
      {1-\lambda^k p_{\mathrm{relax}}}\\
      &\geq
      \frac{-\lambda^t p_{\mathrm{relax}}}
      {(1-\lambda)(1-\lambda^t p_{\mathrm{relax}})}.
    \end{align*}
    Exponentiating this inequality proves
    \eqref{eq:relaxation_product_lower_bound}.
  \end{proof}
  }

\newcommand{\boundedsuperlevelsetsappendix}{%
  \section{Technical Preliminary Results for Functions with
  Bounded Superlevel Sets}
  \label{sec:bounded_superlevel_sets}
  \begin{proposition}\label{prop:bounded_superlevel_sets}
    $f: \R^n \to \R$ has bounded superlevel sets if and only if both
    of the following hold:
    \begin{enumerate}[label=(\roman*)]
      \item\label{item:liminf} $\displaystyle\lim_{\lVert x\rVert \to
        \infty} f(x) = \inf_{x'\in\R^n} f(x') \in [-\infty, +\infty)$
      \item\label{item:infimum} $\forall x\in\R^n$, $\displaystyle
        f(x) > \inf_{x'\in\R^n} f(x')$
    \end{enumerate}
  \end{proposition}

  \begin{proof}
    Set \(\displaystyle b:=\inf_{x\in\R^{n}}f(x)\in[-\infty,\infty)\)
    and treat two mutually exclusive cases.

    \paragraph{Case 1}\label{case:1}
    \(b=-\infty\).

    \smallskip\noindent \textbf{($\Longrightarrow$)} Suppose \(f\)
    has bounded superlevel sets.
    Fix an arbitrary \(M>0\).
    Because \(b=-\infty\), there exists \(a\in f(\R^{n})\) with \(a\le -M\).
    Boundedness of the \(a\)-superlevel set gives an \(R>0\) such that
    \[
      \|x\|>R\;\Longrightarrow\;f(x)<a\le -M.
    \]
    Thus, \ref{item:liminf} holds.
    Condition \ref{item:infimum} is automatic for $b=-\infty$.

    \smallskip\noindent \textbf{($\Longleftarrow$)} Conversely,
    assume \(\displaystyle\lim_{\|x\|\to\infty}f(x)=-\infty\).
    Given \(a\in f(\R^{n})\), choose \(R>0\) so that
    \(\|x\|>R\Rightarrow f(x)<a\).
    Then \(\{x\mid f(x)\ge a\}\subseteq\{x\mid\|x\|\le R\}\), proving
    boundedness.

    \paragraph{Case 2}\label{case:2}
    \(b>-\infty\).

    \smallskip \noindent\textbf{($\Longrightarrow$)} Assume \(f\) has
    bounded superlevel sets.

    \emph{Step 1: show \ref{item:infimum}.} If \(b\in f(\R^{n})\),
    the \(b\)-superlevel set would equal \(\R^{n}\), contradicting boundedness.
    Thus \(f(x)>b\) for all \(x\), establishing \ref{item:infimum}.

    \emph{Step 2: show \ref{item:liminf}.} Let $a\in f(\mathbb{R}^n)$
    be arbitrary with $b<a$.
    Because $f$ has bounded superlevel sets, there is an $R>0$ such
    that $ f(x)\ge a \;\Longrightarrow\; \lVert x\rVert\le R, $ or
    equivalently, $ \lVert x\rVert>R \;\Longrightarrow\; f(x)<a.
    $ Because $b=\inf f(\mathbb{R}^n)$ and $f(x)>b$ for every
    $x\in\mathbb{R}^n$, one can pick $a$ arbitrarily close to $b$;
    this completes the proof of~\ref{item:liminf}.

    \smallskip \noindent\textbf{($\Longleftarrow$)} Now assume
    \ref{item:liminf} and \ref{item:infimum}.
    From \ref{item:infimum} it follows that $f(\R^n) \subset (b, +\infty)$.
    If $f(x) \to b$ as $\lVert x\rVert \to \infty$, then for any $a >
    b$ there exists $R>0$ such that
    \[
      \lVert x\rVert > R \Longrightarrow f(x) < a.
    \]
    Thus, for all $a > b$ (and,
    consequently, for all $a \in f(\R^n)$) the superlevel set $\{x
    \mid f(x) \ge a\}$ lies within the ball $\{x \mid \lVert x\rVert
    \leq R\}$ and is therefore bounded.

  \end{proof}

  \begin{proof}[Proof of Proposition~\ref{prop:radial_unboundedness}]
    The result follows directly from
    Proposition~\ref{prop:bounded_superlevel_sets} by examining the two cases:

    \textbf{(i):} Property \ref{item:radial_unboundedness}
    corresponds to \hyperref[case:1]{Case 1} in the proof of
    Proposition~\ref{prop:bounded_superlevel_sets} where $b = -\infty$.
    In this case, $f$ has bounded superlevel sets if and only if
    $\lim_{\|x\|\to\infty}f(x) = -\infty$, which is equivalent to the
    negative of $f$ being radially unbounded.

    \textbf{(ii):} Property \ref{item:log_unboundedness} corresponds
    to \hyperref[case:2]{Case 2} in the proof of
    Proposition~\ref{prop:bounded_superlevel_sets} where $b > -\infty$.
    The function $f$ has bounded superlevel sets if and only if $f(x)
    \to b$ as $\|x\| \to \infty$ and $f(x) > b$ for all $x$, which
    for finite $b$ is equivalent to
    \[
      \lim_{\|x\|\to\infty}-\log(f(x) - b) = +\infty
      \bigwedge \forall x\in\R^n, f(x) > b.
    \]
  \end{proof}

  The function in
  Proposition~\ref{prop:bounded_superlevel_sets_extension} can be
  constructed as follows.

  \textbf{Case 1:} If $b = -\infty$, then
  \begin{equation}
    \label{eq:bounded_superlevel_extension_unbounded_infimum}
    f_{\mathrm{corr},1}(x) =
    \begin{cases}
      f(x)                                              & \text{if }
      \|x\| < R    \\
      f\left(\frac{xR}{\|x\|} \right) + g(R) - g(\|x\|) & \text{if }
      \|x\| \geq R
    \end{cases}
  \end{equation}
  where $g:\R_{>0} \to \R$ is an arbitrary continuous function with
  $\lim_{x\to+\infty} g(x) = +\infty$.

  \textbf{Case 2:} If $-\infty < b < \min_{\|x\| \leq R} f(x)$, then
  \begin{equation}
    \label{eq:bounded_superlevel_extension_finite_infimum}
    \begin{aligned}
      f_{\mathrm{corr},2}(x)
      & =
      \begin{cases}
        f(x)
        & \text{if } \|x\| < R    \\
        h(x)+b
        & \text{if } \|x\| \geq R,
      \end{cases} \\
      h(x)
      & =
      \left(f\left(\frac{xR}{\|x\|} \right) - b\right)
      \exp\left(g(R)-g(\|x\|)\right).
    \end{aligned}
  \end{equation}
  where $g:\R_{>0} \to \R$ is an arbitrary continuous function with
  $\lim_{x\to+\infty} g(x) = +\infty$.
  \begin{proof}[Proof of
    Proposition~\ref{prop:bounded_superlevel_sets_extension}]
    Continuity can be checked directly by substituting an arbitrary
    $x$ with $\|x\| = R$ and verifying that $f_{\mathrm{corr},1}(x) =
    f_{\mathrm{corr},2}(x) = f(x)$.

    Moreover, both functions have infimum equal to $b$ since
    $f\left(\frac{xR}{\|x\|}\right)$ is bounded on $\R^n$, is never
    equal to $b$, and $\lim_{\|x\|\to\infty} f_{\mathrm{corr},1}(x) =
    \lim_{\|x\|\to\infty} f_{\mathrm{corr},2}(x) = b$.
    Therefore, by Proposition~\ref{prop:bounded_superlevel_sets},
    both functions have bounded superlevel sets.
  \end{proof}
}

\newcommand{\klfunctiondecompositionappendix}{%
  \section{A Decomposition of Class-$\mathcal{KL}$ Functions}
  \label{sec:kl_function_decomposition}

  The following lemma proves the decomposition used in
  Section~\ref{sec:uniform_lp_assumptions}.
  The proof follows the construction in
  \cite[Lemma~8]{Sontag1998Comments} and states the construction in
  the notation used in this paper.

  \begin{lemma}
    \label{lem:kl_product_decomposition}
    For every $\beta\in\KL$, there exist $\kappa,\xi\in\Kinf$ such that
    \begin{equation}
      \beta(d,t)\leq\kappa(d)\xi(e^{-t})
      \qquad
      \text{for all }d\geq0\text{ and }t\geq0.
      \label{eq:kl_product_decomposition_appendix}
    \end{equation}
  \end{lemma}

  \begin{proof}
    Define
    \begin{equation*}
      \bar\beta(d,t):=\beta(d,t)+de^{-t}.
    \end{equation*}
    Then $\beta(d,t)\leq\bar\beta(d,t)$, $\bar\beta\in\KL$, and
    $d\mapsto\bar\beta(d,t)$ belongs to $\Kinf$ for every fixed $t\geq0$.
    It is therefore sufficient to establish the decomposition for $\bar\beta$.

    For $d>0$, define
    \begin{equation*}
      h_d(t):=
      \frac{\bar\beta(d,t)}
      {\bar\beta(d,0)^{1/2}+\bar\beta(d,0)^2},
      \qquad
      h(t):=\sup_{d>0}h_d(t).
    \end{equation*}
    The inequality $\bar\beta(d,0)^{1/2}+\bar\beta(d,0)^2
    \geq\bar\beta(d,0)$ shows that $0\leq h(t)\leq1$.
    The function $h$ is nonincreasing because
    $t\mapsto\bar\beta(d,t)$ is nonincreasing for every fixed $d$.

    It remains to show that $h(t)\to0$ as $t\to\infty$.
    Fix $\epsilon>0$.
    Since
    \begin{equation*}
      \frac{x}{x^{1/2}+x^2}\longrightarrow0
      \quad\text{as }x\to0^+\text{ and as }x\to+\infty,
    \end{equation*}
    and $d\mapsto\bar\beta(d,0)$ belongs to $\Kinf$, there exist
    $0<a<b<\infty$ such that $h_d(0)<\epsilon$ whenever $d\notin[a,b]$.
    Hence, $h_d(t)<\epsilon$ for every $t\geq0$ and every $d\notin[a,b]$.
    For $d\in[a,b]$,
    \begin{equation*}
      h_d(t)
      \leq
      \frac{\bar\beta(b,t)}
      {\bar\beta(a,0)^{1/2}+\bar\beta(a,0)^2}.
    \end{equation*}
    The right-hand side converges to zero as $t\to\infty$.
    Thus, it is smaller than $\epsilon$ for all sufficiently large
    $t$, uniformly for $d\in[a,b]$.
    Consequently, $h(t)\to0$.

    Choose a continuous function $\vartheta:[0,\infty)\to(0,\infty)$
    that is strictly decreasing, converges to zero, and satisfies
    $h(t)\leq\vartheta(t)$ for every $t\geq0$.
    Such a function can be constructed by choosing an increasing
    sequence $\{T_n\}_{n\geq0}$ such that $h(t)\leq2^{-n}$ for $t\geq
    T_n$ and linearly interpolating the values $\vartheta(T_n)=2^{-n+1}$.
    Define
    \begin{equation*}
      \kappa(d):=
      \bar\beta(d,0)^{1/2}+\bar\beta(d,0)^2.
    \end{equation*}
    Then $\kappa\in\Kinf$ and
    \begin{equation*}
      \bar\beta(d,t)\leq\kappa(d)\vartheta(t).
    \end{equation*}

    Define $\xi_0(0):=0$ and
    \begin{equation*}
      \xi_0(r):=\vartheta(-\log r),
      \qquad r\in(0,1].
    \end{equation*}
    The function $\xi_0$ is continuous and strictly increasing on $[0,1]$.
    Extend it to $[0,\infty)$ by setting
    \begin{equation*}
      \xi(r):=
      \begin{cases}
        \xi_0(r), & 0\leq r\leq1,\\
        \xi_0(1)+r-1, & r>1.
      \end{cases}
    \end{equation*}
    Then $\xi\in\Kinf$ and $\vartheta(t)=\xi(e^{-t})$ for every $t\geq0$.
    Therefore,
    \begin{equation*}
      \beta(d,t)\leq\bar\beta(d,t)\leq\kappa(d)\xi(e^{-t}),
    \end{equation*}
    which proves \eqref{eq:kl_product_decomposition_appendix}.
  \end{proof}

  The factor $\kappa(d)$ contains the dependence on the initial distance $d$.
  The factor $\xi(e^{-t})$ contains the dependence on time and
  converges to zero because $e^{-t}\to0$ and $\xi(0)=0$.
  The decomposition does not require $\beta(d,t)$ itself to decrease
  exponentially in $t$.
  The function $\xi$ transforms the exponential quantity $e^{-t}$
  into the time dependence required to bound $\beta$.
}

\gdef\uniformasappendix{%
\section{Uniform Almost-Sure Bounds for CALF-Wrapper}
\label{sec:uniform_hitting_time}

The following assumption is an alternative to
Assumption~\ref{ass:lp_uniform_fallback}.
The superscript $\mathrm{U}\prime$ distinguishes the alternative
uniform assumption and claims in this appendix from those in
Section~\ref{sec:uniform_lp_bounds}.
\begin{enumerate}[label={$\bigl(\mathrm{A}_{3}^{\mathrm{U}\prime}\bigr)$},
    leftmargin=3.2em]
  \item\label{ass:lp_uniform_fallback_as}
    The fallback policy satisfies the uniform almost-sure
    goal-reaching property: there exists a function $\beta\in\KL$
    such that, for every initial state $\state_0\in\states$,
    \begin{equation}
      \PP{
        \begin{gathered}
          \goaldist\bigl(\State_k^{\fallbackpolicy}(\state_0)\bigr)
          \leq\beta\bigl(\goaldist(\state_0),k\bigr)\\
          \text{for every }k\in\mathbb Z_{\geq0}
        \end{gathered}
      }
      =1.
      \label{eq:uniform_as_goal_reaching}
    \end{equation}
    The same function $\beta$ applies to every initial state.
\end{enumerate}

Assumption~\ref{ass:lp_uniform_fallback_as} is stronger than
Assumption~\ref{ass:lp_uniform_fallback} when both assumptions use
the same function $\beta$.
Indeed, \eqref{eq:uniform_as_goal_reaching} directly implies the
expected-distance bound in
Definition~\ref{dfn:uniform_lp_goal_reaching}.
For the statement below, define
\[
  \Delta^{\prime}(d^{\circ})
  :=\max\left\{\delta(d^{\circ}),\beta(\delta(d^{\circ}),0)\right\}.
\]
Let $\kappa,\xi\in\Kinf$ satisfy
$\beta(d,t)\leq\kappa(d)\xi(e^{-t})$ for every $d\geq0$ and
$t\geq0$, as provided by the class-$\KL$ decomposition in
\cite[Lemma~8]{Sontag1998Comments} and
Appendix~\ref{sec:kl_function_decomposition}.
For every $d^{*}>0$, define
\begin{equation*}
  \ell(d^{\circ},d^{*})
  :=\max\left\{1,
    \left\lceil
      -\log\left(
        \xi^{-1}\left(
          \frac{d^{*}}{\kappa(\delta(d^{\circ}))}
        \right)
      \right)
    \right\rceil
  \right\}.
\end{equation*}
This construction gives $\ell(d^{\circ},d^{*})\in\mathbb Z_{>0}$ and
\begin{equation*}
  \beta\bigl(\delta(d^{\circ}),\ell(d^{\circ},d^{*})\bigr)\leq d^{*}.
\end{equation*}

\begin{theorem}[Uniform almost-sure bounds for CALF-Wrapper]
  \label{thm:unform_goal_reaching}
  Consider Algorithm~\ref{alg_calfwrapper} initialized at any
  $\state_0$ satisfying
  $\goaldist(\state_0)\leq d^{\circ}$.
  Under Assumptions~\ref{ass:lp_acceptance},
  \ref{ass:lp_critic}, and~\ref{ass:lp_uniform_fallback_as}, the
  following claims hold.
  \begin{enumerate}[label={$\bigl(\mathrm{C}_{\arabic*}^{\mathrm{U}\prime}\bigr)$},
      leftmargin=3.2em]
    \item\label{claim:uniform_overshoot_bound}
      The distance from $\G$ satisfies
      \begin{equation}
        \PP{
          \sup_{k\in\mathbb Z_{\geq0}}
          \goaldist\bigl(
            \State_k^{\calfwpolicy}(\state_0)
          \bigr)
          \leq\Delta^{\prime}(d^{\circ})
        }
        =1.
        \label{eq:uniform_overshoot_bound}
      \end{equation}

    \item\label{claim:hitting_time_distribution}
      For every $d^{*}\in(0,d^{\circ})$,
      $t\in\mathbb Z_{\geq0}$, and the integer
      $\ell(d^{\circ},d^{*})$ constructed above,
      \begin{equation}
        \PP{
          T_{\mathrm{hit}}(\state_0,d^{*})
          \leq H(d^{\circ},t,\ell(d^{\circ},d^{*}))
        }
        \geq\mathcal P(t).
        \label{eq:hitting_time_distribution}
      \end{equation}
      In particular,
      \begin{equation}
        \PP{T_{\mathrm{hit}}(\state_0,d^{*})<+\infty}=1.
        \label{eq:uniform_as_finite_hitting_time}
      \end{equation}

    \item\label{claim:reaching_time_distribution}
      For every $d^{*}\in(0,d^{\circ})$, suppose additionally that
      Assumption~\ref{ass:lp_fallback_locking} holds for this
      prescribed value of $d^{*}$.
      There exists $\underline{d}^{*}\in(0,d^{*}]$ such that
      \begin{equation*}
        \beta(\underline{d}^{*},0)\leq d^{*}.
      \end{equation*}
      For every $t\in\mathbb Z_{\geq0}$ and the integer
      $\ell(d^{\circ},\underline{d}^{*})$ constructed above,
      \begin{equation}
        \PP{
          T_{\mathrm{set}}(\state_0,d^{*})
          \leq H(d^{\circ},t,\ell(d^{\circ},\underline{d}^{*}))
        }
        \geq\mathcal P(t).
        \label{eq:reaching_time_distribution}
      \end{equation}
      In particular,
      \begin{equation}
        \PP{T_{\mathrm{set}}(\state_0,d^{*})<+\infty}=1.
        \label{eq:uniform_as_finite_reaching_time}
      \end{equation}
  \end{enumerate}
\end{theorem}

\begin{lemma}
  \label{lem:uniform_as_fallback_sequence}
  Suppose that
  Assumption~\ref{ass:lp_uniform_fallback_as} holds.
  Let $\sigma$ be a stopping time such that
  $\goaldist(\State_{\sigma})\leq\delta(d^{\circ})$.
  For the state process generated by the fallback policy from
  $\State_{\sigma}$,
  \begin{equation}
    \PP{
      \goaldist\bigl(
        \State_k^{\fallbackpolicy}(\State_{\sigma})
      \bigr)
      \leq\beta(\delta(d^{\circ}),k)
      \text{ for every }k\in\mathbb Z_{\geq0}
      \mid\State_{\sigma}
    }
    =1.
    \label{eq:uniform_as_fallback_sequence}
  \end{equation}
  If Algorithm~\ref{alg_calfwrapper} selects the fallback policy at
  times $\sigma,\ldots,\sigma+\ell-1$, then its state at time
  $\sigma+\ell$ coincides with the state in
  \eqref{eq:uniform_as_fallback_sequence} under the same realizations
  of the fallback-policy actions and system transitions.
\end{lemma}

\begin{proof}
  By the strong Markov property, conditionally on
  $\State_{\sigma}$, the state process after time $\sigma$ has the
  same transition law as the state process generated by the fallback
  policy and initialized at $\State_{\sigma}$.
  Assumption~\ref{ass:lp_uniform_fallback_as} and monotonicity of
  $\beta$ in its first argument give
  \eqref{eq:uniform_as_fallback_sequence}.
  The final statement follows because both state processes apply the
  fallback policy for the stated $\ell$ time steps.
\end{proof}

\begin{proof}[Proof of Theorem~\ref{thm:unform_goal_reaching}]
  Lemma~\ref{lem:lp_base_policy_states} shows that the initial state
  and every state immediately following an action supplied by the
  base policy are within distance $\delta(d^{\circ})$ of $\G$.
  Lemma~\ref{lem:uniform_as_fallback_sequence} shows that, during
  every consecutive sequence of actions supplied by the fallback
  policy, the distance from $\G$ is at most
  $\beta(\delta(d^{\circ}),0)$.
  The time indices at which such sequences can begin are countable.
  The two bounds therefore hold simultaneously almost surely at
  every time and prove Claim~\ref{claim:uniform_overshoot_bound}.

  Fix $d^{*}\in(0,d^{\circ})$ and $t\in\mathbb Z_{\geq0}$.
  The construction preceding the theorem gives
  \begin{equation*}
    \beta\bigl(
      \delta(d^{\circ}),\ell(d^{\circ},d^{*})
    \bigr)
    \leq d^{*}.
  \end{equation*}
  Define
  \begin{equation*}
    \mathbb{U}_t
    :=\bigcap_{k=t}^{\infty}\{U_k\geq\bar\rho_k\}.
  \end{equation*}
  Independence of the uniform random variables gives
  \begin{equation*}
    \PP{\mathbb{U}_t}
    =\prod_{k=t}^{\infty}(1-\bar\rho_k)
    =\mathcal P(t).
  \end{equation*}
  Lemma~\ref{lem:lp_critic_selection_count} and
  Lemma~\ref{lem:lp_fallback_block} show that, on
  $\mathbb{U}_t$, the first
  $H(d^{\circ},t,\ell(d^{\circ},d^{*}))$ actions contain at least
  $\ell(d^{\circ},d^{*})$ consecutive actions supplied by the
  fallback policy.
  This sequence is contained in a maximal consecutive sequence of
  actions supplied by the fallback policy that begins either at time
  zero or immediately after an action supplied by the base policy.
  The initial state of the maximal sequence is therefore within distance
  $\delta(d^{\circ})$ of $\G$ by
  Lemma~\ref{lem:lp_base_policy_states}.
  The maximal sequence contains at least $\ell(d^{\circ},d^{*})$
  actions.
  Lemma~\ref{lem:uniform_as_fallback_sequence} then gives
  \begin{equation*}
    \goaldist\bigl(
      \State_{\ell(d^{\circ},d^{*})}^{\fallbackpolicy}(\State_{\sigma})
    \bigr)
    \leq\beta\bigl(
      \delta(d^{\circ}),\ell(d^{\circ},d^{*})
    \bigr)
    \leq d^{*}
  \end{equation*}
  almost surely for the initial time $\sigma$ of this sequence.
  Consequently,
  \begin{equation*}
    \PP{
      T_{\mathrm{hit}}(\state_0,d^{*})
      \leq H(d^{\circ},t,\ell(d^{\circ},d^{*}))
    }
    \geq\PP{\mathbb{U}_t}
    =\mathcal P(t),
  \end{equation*}
  which proves \eqref{eq:hitting_time_distribution}.
  Since $\mathcal P(t)\to1$ as $t\to\infty$,
  \eqref{eq:uniform_as_finite_hitting_time} follows.

  Suppose that Assumption~\ref{ass:lp_fallback_locking} holds.
  Continuity of $\beta$ in its first argument and
  $\beta(0,0)=0$ imply that there exists
  $\underline{d}^{*}\in(0,d^{*}]$ satisfying
  $\beta(\underline{d}^{*},0)\leq d^{*}$.
  The construction preceding the theorem gives
  $\beta(\delta(d^{\circ}),
  \ell(d^{\circ},\underline{d}^{*}))\leq\underline{d}^{*}$.
  Applying the preceding argument with $\underline{d}^{*}$ in place
  of $d^{*}$ shows that, on $\mathbb{U}_t$, the trajectory enters the
  $\underline{d}^{*}$-neighborhood no later than
  $H(d^{\circ},t,\ell(d^{\circ},\underline{d}^{*}))$.
  After this entry,
  Algorithm~\ref{alg_calfwrapper} selects the fallback policy by
  Assumption~\ref{ass:lp_fallback_locking}.
  Assumption~\ref{ass:lp_uniform_fallback_as} gives, almost surely,
  \begin{equation*}
    \goaldist\bigl(
      \State_k^{\fallbackpolicy}(\State_{\sigma})
    \bigr)
    \leq\beta(\underline{d}^{*},k)
    \leq\beta(\underline{d}^{*},0)
    \leq d^{*}
  \end{equation*}
  for every subsequent time $k$.
  Therefore,
  \begin{equation*}
    \PP{
      T_{\mathrm{set}}(\state_0,d^{*})
      \leq H(d^{\circ},t,\ell(d^{\circ},\underline{d}^{*}))
    }
    \geq\mathcal P(t),
  \end{equation*}
  which proves \eqref{eq:reaching_time_distribution}.
  Letting $t\to\infty$ proves
  \eqref{eq:uniform_as_finite_reaching_time}.
\end{proof}
}

\section{Experimental Study}\label{sec:experiments}

\newcommand{\NabCartpoleConservativeGain}{286.2}
\newcommand{\NabCartpoleConservativeGainCI}{47.88}
\newcommand{\NabCartpoleConservativeGoal}{0.673}
\newcommand{\NabCartpoleGuardedGain}{512.9}
\newcommand{\NabCartpoleGuardedGainCI}{44.93}
\newcommand{\NabCartpoleGuardedGoal}{0.930}
\newcommand{\NabCartpoleModerateGain}{436.7}
\newcommand{\NabCartpoleModerateGainCI}{39.23}
\newcommand{\NabCartpoleModerateGoal}{0.911}
\newcommand{\NabCartpoleBalancedGain}{359.5}
\newcommand{\NabCartpoleBalancedGainCI}{33.96}
\newcommand{\NabCartpoleBalancedGoal}{0.839}
\newcommand{\NabCartpoleHighGain}{186.6}
\newcommand{\NabCartpoleHighGainCI}{23.82}
\newcommand{\NabCartpoleHighGoal}{0.637}
\newcommand{\NabCartpoleAlmostOpenGain}{22.0}
\newcommand{\NabCartpoleAlmostOpenGainCI}{10.18}
\newcommand{\NabCartpoleAlmostOpenGoal}{0.505}
\newcommand{\NabPendulumConservativeGain}{205.9}
\newcommand{\NabPendulumConservativeGainCI}{37.65}
\newcommand{\NabPendulumConservativeGoal}{0.983}
\newcommand{\NabPendulumGuardedGain}{193.8}
\newcommand{\NabPendulumGuardedGainCI}{33.49}
\newcommand{\NabPendulumGuardedGoal}{0.985}
\newcommand{\NabPendulumModerateGain}{116.8}
\newcommand{\NabPendulumModerateGainCI}{23.41}
\newcommand{\NabPendulumModerateGoal}{0.904}
\newcommand{\NabPendulumBalancedGain}{71.1}
\newcommand{\NabPendulumBalancedGainCI}{16.17}
\newcommand{\NabPendulumBalancedGoal}{0.762}
\newcommand{\NabPendulumHighGain}{32.2}
\newcommand{\NabPendulumHighGainCI}{7.05}
\newcommand{\NabPendulumHighGoal}{0.563}
\newcommand{\NabPendulumAlmostOpenGain}{5.2}
\newcommand{\NabPendulumAlmostOpenGainCI}{1.15}
\newcommand{\NabPendulumAlmostOpenGoal}{0.380}
\newcommand{\NabRobotConservativeGain}{-5.8}
\newcommand{\NabRobotConservativeGainCI}{4.25}
\newcommand{\NabRobotConservativeGoal}{1.000}
\newcommand{\NabRobotGuardedGain}{1.9}
\newcommand{\NabRobotGuardedGainCI}{3.30}
\newcommand{\NabRobotGuardedGoal}{1.000}
\newcommand{\NabRobotModerateGain}{1.5}
\newcommand{\NabRobotModerateGainCI}{2.15}
\newcommand{\NabRobotModerateGoal}{1.000}
\newcommand{\NabRobotBalancedGain}{1.0}
\newcommand{\NabRobotBalancedGainCI}{1.56}
\newcommand{\NabRobotBalancedGoal}{0.958}
\newcommand{\NabRobotHighGain}{0.7}
\newcommand{\NabRobotHighGainCI}{1.01}
\newcommand{\NabRobotHighGoal}{0.253}
\newcommand{\NabRobotAlmostOpenGain}{0.3}
\newcommand{\NabRobotAlmostOpenGainCI}{0.59}
\newcommand{\NabRobotAlmostOpenGoal}{0.213}
\newcommand{\NabUnderwaterConservativeGain}{565.1}
\newcommand{\NabUnderwaterConservativeGainCI}{101.59}
\newcommand{\NabUnderwaterConservativeGoal}{0.998}
\newcommand{\NabUnderwaterGuardedGain}{359.1}
\newcommand{\NabUnderwaterGuardedGainCI}{91.57}
\newcommand{\NabUnderwaterGuardedGoal}{0.998}
\newcommand{\NabUnderwaterModerateGain}{114.4}
\newcommand{\NabUnderwaterModerateGainCI}{57.54}
\newcommand{\NabUnderwaterModerateGoal}{0.983}
\newcommand{\NabUnderwaterBalancedGain}{-67.7}
\newcommand{\NabUnderwaterBalancedGainCI}{57.90}
\newcommand{\NabUnderwaterBalancedGoal}{0.887}
\newcommand{\NabUnderwaterHighGain}{28.3}
\newcommand{\NabUnderwaterHighGainCI}{29.26}
\newcommand{\NabUnderwaterHighGoal}{0.829}
\newcommand{\NabUnderwaterAlmostOpenGain}{5.2}
\newcommand{\NabUnderwaterAlmostOpenGainCI}{5.50}
\newcommand{\NabUnderwaterAlmostOpenGoal}{0.765}
\newcommand{\NabPendulumEarlyConservativeGain}{404.7}
\newcommand{\NabPendulumEarlyConservativeGainCI}{102.4}
\newcommand{\NabPendulumEarlyConservativeGoal}{0.960}
\newcommand{\NabPendulumEarlyConservativeNab}{0.139}
\newcommand{\NabPendulumEarlyConservativeFallback}{0.861}
\newcommand{\NabPendulumEarlyGuardedGain}{394.1}
\newcommand{\NabPendulumEarlyGuardedGainCI}{95.3}
\newcommand{\NabPendulumEarlyGuardedGoal}{0.977}
\newcommand{\NabPendulumEarlyGuardedNab}{0.205}
\newcommand{\NabPendulumEarlyGuardedFallback}{0.795}
\newcommand{\NabPendulumEarlyModerateGain}{262.0}
\newcommand{\NabPendulumEarlyModerateGainCI}{79.6}
\newcommand{\NabPendulumEarlyModerateGoal}{0.853}
\newcommand{\NabPendulumEarlyModerateNab}{0.410}
\newcommand{\NabPendulumEarlyModerateFallback}{0.590}
\newcommand{\NabPendulumEarlyBalancedGain}{147.3}
\newcommand{\NabPendulumEarlyBalancedGainCI}{56.2}
\newcommand{\NabPendulumEarlyBalancedGoal}{0.520}
\newcommand{\NabPendulumEarlyBalancedNab}{0.534}
\newcommand{\NabPendulumEarlyBalancedFallback}{0.466}
\newcommand{\NabPendulumEarlyHighGain}{45.8}
\newcommand{\NabPendulumEarlyHighGainCI}{32.6}
\newcommand{\NabPendulumEarlyHighGoal}{0.157}
\newcommand{\NabPendulumEarlyHighNab}{0.714}
\newcommand{\NabPendulumEarlyHighFallback}{0.286}
\newcommand{\NabPendulumEarlyAlmostOpenGain}{10.6}
\newcommand{\NabPendulumEarlyAlmostOpenGainCI}{3.8}
\newcommand{\NabPendulumEarlyAlmostOpenGoal}{0.080}
\newcommand{\NabPendulumEarlyAlmostOpenNab}{0.953}
\newcommand{\NabPendulumEarlyAlmostOpenFallback}{0.047}
\newcommand{\NabPendulumMidConservativeGain}{58.8}
\newcommand{\NabPendulumMidConservativeGainCI}{66.3}
\newcommand{\NabPendulumMidConservativeGoal}{0.997}
\newcommand{\NabPendulumMidConservativeNab}{0.151}
\newcommand{\NabPendulumMidConservativeFallback}{0.849}
\newcommand{\NabPendulumMidGuardedGain}{69.1}
\newcommand{\NabPendulumMidGuardedGainCI}{64.2}
\newcommand{\NabPendulumMidGuardedGoal}{1.000}
\newcommand{\NabPendulumMidGuardedNab}{0.211}
\newcommand{\NabPendulumMidGuardedFallback}{0.789}
\newcommand{\NabPendulumMidModerateGain}{38.6}
\newcommand{\NabPendulumMidModerateGainCI}{45.5}
\newcommand{\NabPendulumMidModerateGoal}{0.993}
\newcommand{\NabPendulumMidModerateNab}{0.407}
\newcommand{\NabPendulumMidModerateFallback}{0.593}
\newcommand{\NabPendulumMidBalancedGain}{17.0}
\newcommand{\NabPendulumMidBalancedGainCI}{27.1}
\newcommand{\NabPendulumMidBalancedGoal}{0.960}
\newcommand{\NabPendulumMidBalancedNab}{0.536}
\newcommand{\NabPendulumMidBalancedFallback}{0.464}
\newcommand{\NabPendulumMidHighGain}{8.0}
\newcommand{\NabPendulumMidHighGainCI}{15.1}
\newcommand{\NabPendulumMidHighGoal}{0.850}
\newcommand{\NabPendulumMidHighNab}{0.719}
\newcommand{\NabPendulumMidHighFallback}{0.281}
\newcommand{\NabPendulumMidAlmostOpenGain}{7.4}
\newcommand{\NabPendulumMidAlmostOpenGainCI}{6.9}
\newcommand{\NabPendulumMidAlmostOpenGoal}{0.543}
\newcommand{\NabPendulumMidAlmostOpenNab}{0.954}
\newcommand{\NabPendulumMidAlmostOpenFallback}{0.046}
\newcommand{\NabPendulumLateConservativeGain}{77.0}
\newcommand{\NabPendulumLateConservativeGainCI}{111.4}
\newcommand{\NabPendulumLateConservativeGoal}{0.983}
\newcommand{\NabPendulumLateConservativeNab}{0.134}
\newcommand{\NabPendulumLateConservativeFallback}{0.866}
\newcommand{\NabPendulumLateGuardedGain}{88.9}
\newcommand{\NabPendulumLateGuardedGainCI}{106.1}
\newcommand{\NabPendulumLateGuardedGoal}{0.970}
\newcommand{\NabPendulumLateGuardedNab}{0.187}
\newcommand{\NabPendulumLateGuardedFallback}{0.813}
\newcommand{\NabPendulumLateModerateGain}{70.8}
\newcommand{\NabPendulumLateModerateGainCI}{70.1}
\newcommand{\NabPendulumLateModerateGoal}{0.923}
\newcommand{\NabPendulumLateModerateNab}{0.399}
\newcommand{\NabPendulumLateModerateFallback}{0.601}
\newcommand{\NabPendulumLateBalancedGain}{52.6}
\newcommand{\NabPendulumLateBalancedGainCI}{47.7}
\newcommand{\NabPendulumLateBalancedGoal}{0.883}
\newcommand{\NabPendulumLateBalancedNab}{0.529}
\newcommand{\NabPendulumLateBalancedFallback}{0.471}
\newcommand{\NabPendulumLateHighGain}{33.3}
\newcommand{\NabPendulumLateHighGainCI}{32.1}
\newcommand{\NabPendulumLateHighGoal}{0.773}
\newcommand{\NabPendulumLateHighNab}{0.715}
\newcommand{\NabPendulumLateHighFallback}{0.285}
\newcommand{\NabPendulumLateAlmostOpenGain}{13.2}
\newcommand{\NabPendulumLateAlmostOpenGainCI}{11.0}
\newcommand{\NabPendulumLateAlmostOpenGoal}{0.410}
\newcommand{\NabPendulumLateAlmostOpenNab}{0.953}
\newcommand{\NabPendulumLateAlmostOpenFallback}{0.047}
\newcommand{\NabCartpoleEarlyConservativeGain}{-104.1}
\newcommand{\NabCartpoleEarlyConservativeGainCI}{371.9}
\newcommand{\NabCartpoleEarlyConservativeGoal}{0.694}
\newcommand{\NabCartpoleEarlyConservativeNab}{0.144}
\newcommand{\NabCartpoleEarlyConservativeFallback}{0.856}
\newcommand{\NabCartpoleEarlyGuardedGain}{164.5}
\newcommand{\NabCartpoleEarlyGuardedGainCI}{335.7}
\newcommand{\NabCartpoleEarlyGuardedGoal}{0.958}
\newcommand{\NabCartpoleEarlyGuardedNab}{0.174}
\newcommand{\NabCartpoleEarlyGuardedFallback}{0.826}
\newcommand{\NabCartpoleEarlyModerateGain}{190.4}
\newcommand{\NabCartpoleEarlyModerateGainCI}{308.5}
\newcommand{\NabCartpoleEarlyModerateGoal}{0.994}
\newcommand{\NabCartpoleEarlyModerateNab}{0.378}
\newcommand{\NabCartpoleEarlyModerateFallback}{0.622}
\newcommand{\NabCartpoleEarlyBalancedGain}{158.7}
\newcommand{\NabCartpoleEarlyBalancedGainCI}{238.1}
\newcommand{\NabCartpoleEarlyBalancedGoal}{0.973}
\newcommand{\NabCartpoleEarlyBalancedNab}{0.520}
\newcommand{\NabCartpoleEarlyBalancedFallback}{0.480}
\newcommand{\NabCartpoleEarlyHighGain}{104.7}
\newcommand{\NabCartpoleEarlyHighGainCI}{131.4}
\newcommand{\NabCartpoleEarlyHighGoal}{0.882}
\newcommand{\NabCartpoleEarlyHighNab}{0.711}
\newcommand{\NabCartpoleEarlyHighFallback}{0.289}
\newcommand{\NabCartpoleEarlyAlmostOpenGain}{32.3}
\newcommand{\NabCartpoleEarlyAlmostOpenGainCI}{31.7}
\newcommand{\NabCartpoleEarlyAlmostOpenGoal}{0.885}
\newcommand{\NabCartpoleEarlyAlmostOpenNab}{0.952}
\newcommand{\NabCartpoleEarlyAlmostOpenFallback}{0.048}
\newcommand{\NabCartpoleMidConservativeGain}{-294.8}
\newcommand{\NabCartpoleMidConservativeGainCI}{98.6}
\newcommand{\NabCartpoleMidConservativeGoal}{0.718}
\newcommand{\NabCartpoleMidConservativeNab}{0.150}
\newcommand{\NabCartpoleMidConservativeFallback}{0.850}
\newcommand{\NabCartpoleMidGuardedGain}{1.2}
\newcommand{\NabCartpoleMidGuardedGainCI}{89.5}
\newcommand{\NabCartpoleMidGuardedGoal}{0.988}
\newcommand{\NabCartpoleMidGuardedNab}{0.164}
\newcommand{\NabCartpoleMidGuardedFallback}{0.836}
\newcommand{\NabCartpoleMidModerateGain}{40.0}
\newcommand{\NabCartpoleMidModerateGainCI}{84.1}
\newcommand{\NabCartpoleMidModerateGoal}{0.997}
\newcommand{\NabCartpoleMidModerateNab}{0.372}
\newcommand{\NabCartpoleMidModerateFallback}{0.628}
\newcommand{\NabCartpoleMidBalancedGain}{55.8}
\newcommand{\NabCartpoleMidBalancedGainCI}{87.4}
\newcommand{\NabCartpoleMidBalancedGoal}{0.994}
\newcommand{\NabCartpoleMidBalancedNab}{0.516}
\newcommand{\NabCartpoleMidBalancedFallback}{0.484}
\newcommand{\NabCartpoleMidHighGain}{53.2}
\newcommand{\NabCartpoleMidHighGainCI}{84.9}
\newcommand{\NabCartpoleMidHighGoal}{0.936}
\newcommand{\NabCartpoleMidHighNab}{0.709}
\newcommand{\NabCartpoleMidHighFallback}{0.291}
\newcommand{\NabCartpoleMidAlmostOpenGain}{33.8}
\newcommand{\NabCartpoleMidAlmostOpenGainCI}{61.6}
\newcommand{\NabCartpoleMidAlmostOpenGoal}{0.897}
\newcommand{\NabCartpoleMidAlmostOpenNab}{0.952}
\newcommand{\NabCartpoleMidAlmostOpenFallback}{0.048}
\newcommand{\NabCartpoleLateConservativeGain}{508.0}
\newcommand{\NabCartpoleLateConservativeGainCI}{486.4}
\newcommand{\NabCartpoleLateConservativeGoal}{0.648}
\newcommand{\NabCartpoleLateConservativeNab}{0.140}
\newcommand{\NabCartpoleLateConservativeFallback}{0.860}
\newcommand{\NabCartpoleLateGuardedGain}{669.2}
\newcommand{\NabCartpoleLateGuardedGainCI}{416.6}
\newcommand{\NabCartpoleLateGuardedGoal}{0.879}
\newcommand{\NabCartpoleLateGuardedNab}{0.201}
\newcommand{\NabCartpoleLateGuardedFallback}{0.799}
\newcommand{\NabCartpoleLateModerateGain}{565.0}
\newcommand{\NabCartpoleLateModerateGainCI}{373.0}
\newcommand{\NabCartpoleLateModerateGoal}{0.812}
\newcommand{\NabCartpoleLateModerateNab}{0.403}
\newcommand{\NabCartpoleLateModerateFallback}{0.597}
\newcommand{\NabCartpoleLateBalancedGain}{516.6}
\newcommand{\NabCartpoleLateBalancedGainCI}{377.4}
\newcommand{\NabCartpoleLateBalancedGoal}{0.700}
\newcommand{\NabCartpoleLateBalancedNab}{0.539}
\newcommand{\NabCartpoleLateBalancedFallback}{0.461}
\newcommand{\NabCartpoleLateHighGain}{399.9}
\newcommand{\NabCartpoleLateHighGainCI}{340.3}
\newcommand{\NabCartpoleLateHighGoal}{0.509}
\newcommand{\NabCartpoleLateHighNab}{0.723}
\newcommand{\NabCartpoleLateHighFallback}{0.277}
\newcommand{\NabCartpoleLateAlmostOpenGain}{278.6}
\newcommand{\NabCartpoleLateAlmostOpenGainCI}{243.9}
\newcommand{\NabCartpoleLateAlmostOpenGoal}{0.276}
\newcommand{\NabCartpoleLateAlmostOpenNab}{0.953}
\newcommand{\NabCartpoleLateAlmostOpenFallback}{0.047}
\newcommand{\NabUnderwaterEarlyConservativeGain}{772.3}
\newcommand{\NabUnderwaterEarlyConservativeGainCI}{820.9}
\newcommand{\NabUnderwaterEarlyConservativeGoal}{1.000}
\newcommand{\NabUnderwaterEarlyConservativeNab}{0.132}
\newcommand{\NabUnderwaterEarlyConservativeFallback}{0.868}
\newcommand{\NabUnderwaterEarlyGuardedGain}{550.8}
\newcommand{\NabUnderwaterEarlyGuardedGainCI}{611.0}
\newcommand{\NabUnderwaterEarlyGuardedGoal}{1.000}
\newcommand{\NabUnderwaterEarlyGuardedNab}{0.198}
\newcommand{\NabUnderwaterEarlyGuardedFallback}{0.802}
\newcommand{\NabUnderwaterEarlyModerateGain}{-88.9}
\newcommand{\NabUnderwaterEarlyModerateGainCI}{106.4}
\newcommand{\NabUnderwaterEarlyModerateGoal}{0.989}
\newcommand{\NabUnderwaterEarlyModerateNab}{0.414}
\newcommand{\NabUnderwaterEarlyModerateFallback}{0.586}
\newcommand{\NabUnderwaterEarlyBalancedGain}{-496.4}
\newcommand{\NabUnderwaterEarlyBalancedGainCI}{501.9}
\newcommand{\NabUnderwaterEarlyBalancedGoal}{0.689}
\newcommand{\NabUnderwaterEarlyBalancedNab}{0.537}
\newcommand{\NabUnderwaterEarlyBalancedFallback}{0.463}
\newcommand{\NabUnderwaterEarlyHighGain}{-31.9}
\newcommand{\NabUnderwaterEarlyHighGainCI}{60.8}
\newcommand{\NabUnderwaterEarlyHighGoal}{0.633}
\newcommand{\NabUnderwaterEarlyHighNab}{0.720}
\newcommand{\NabUnderwaterEarlyHighFallback}{0.280}
\newcommand{\NabUnderwaterEarlyAlmostOpenGain}{-2.9}
\newcommand{\NabUnderwaterEarlyAlmostOpenGainCI}{6.2}
\newcommand{\NabUnderwaterEarlyAlmostOpenGoal}{0.644}
\newcommand{\NabUnderwaterEarlyAlmostOpenNab}{0.955}
\newcommand{\NabUnderwaterEarlyAlmostOpenFallback}{0.045}
\newcommand{\NabUnderwaterMidConservativeGain}{470.5}
\newcommand{\NabUnderwaterMidConservativeGainCI}{616.3}
\newcommand{\NabUnderwaterMidConservativeGoal}{1.000}
\newcommand{\NabUnderwaterMidConservativeNab}{0.139}
\newcommand{\NabUnderwaterMidConservativeFallback}{0.861}
\newcommand{\NabUnderwaterMidGuardedGain}{264.0}
\newcommand{\NabUnderwaterMidGuardedGainCI}{586.4}
\newcommand{\NabUnderwaterMidGuardedGoal}{1.000}
\newcommand{\NabUnderwaterMidGuardedNab}{0.210}
\newcommand{\NabUnderwaterMidGuardedFallback}{0.790}
\newcommand{\NabUnderwaterMidModerateGain}{160.3}
\newcommand{\NabUnderwaterMidModerateGainCI}{255.2}
\newcommand{\NabUnderwaterMidModerateGoal}{0.978}
\newcommand{\NabUnderwaterMidModerateNab}{0.423}
\newcommand{\NabUnderwaterMidModerateFallback}{0.577}
\newcommand{\NabUnderwaterMidBalancedGain}{-13.1}
\newcommand{\NabUnderwaterMidBalancedGainCI}{56.2}
\newcommand{\NabUnderwaterMidBalancedGoal}{0.889}
\newcommand{\NabUnderwaterMidBalancedNab}{0.551}
\newcommand{\NabUnderwaterMidBalancedFallback}{0.449}
\newcommand{\NabUnderwaterMidHighGain}{7.5}
\newcommand{\NabUnderwaterMidHighGainCI}{25.9}
\newcommand{\NabUnderwaterMidHighGoal}{0.867}
\newcommand{\NabUnderwaterMidHighNab}{0.725}
\newcommand{\NabUnderwaterMidHighFallback}{0.275}
\newcommand{\NabUnderwaterMidAlmostOpenGain}{19.4}
\newcommand{\NabUnderwaterMidAlmostOpenGainCI}{20.3}
\newcommand{\NabUnderwaterMidAlmostOpenGoal}{0.867}
\newcommand{\NabUnderwaterMidAlmostOpenNab}{0.955}
\newcommand{\NabUnderwaterMidAlmostOpenFallback}{0.045}
\newcommand{\NabUnderwaterLateConservativeGain}{583.7}
\newcommand{\NabUnderwaterLateConservativeGainCI}{616.5}
\newcommand{\NabUnderwaterLateConservativeGoal}{1.000}
\newcommand{\NabUnderwaterLateConservativeNab}{0.142}
\newcommand{\NabUnderwaterLateConservativeFallback}{0.858}
\newcommand{\NabUnderwaterLateGuardedGain}{359.8}
\newcommand{\NabUnderwaterLateGuardedGainCI}{561.9}
\newcommand{\NabUnderwaterLateGuardedGoal}{0.989}
\newcommand{\NabUnderwaterLateGuardedNab}{0.196}
\newcommand{\NabUnderwaterLateGuardedFallback}{0.804}
\newcommand{\NabUnderwaterLateModerateGain}{68.3}
\newcommand{\NabUnderwaterLateModerateGainCI}{338.9}
\newcommand{\NabUnderwaterLateModerateGoal}{1.000}
\newcommand{\NabUnderwaterLateModerateNab}{0.411}
\newcommand{\NabUnderwaterLateModerateFallback}{0.589}
\newcommand{\NabUnderwaterLateBalancedGain}{-114.8}
\newcommand{\NabUnderwaterLateBalancedGainCI}{607.5}
\newcommand{\NabUnderwaterLateBalancedGoal}{0.867}
\newcommand{\NabUnderwaterLateBalancedNab}{0.540}
\newcommand{\NabUnderwaterLateBalancedFallback}{0.460}
\newcommand{\NabUnderwaterLateHighGain}{100.1}
\newcommand{\NabUnderwaterLateHighGainCI}{117.5}
\newcommand{\NabUnderwaterLateHighGoal}{0.767}
\newcommand{\NabUnderwaterLateHighNab}{0.721}
\newcommand{\NabUnderwaterLateHighFallback}{0.279}
\newcommand{\NabUnderwaterLateAlmostOpenGain}{3.2}
\newcommand{\NabUnderwaterLateAlmostOpenGainCI}{8.2}
\newcommand{\NabUnderwaterLateAlmostOpenGoal}{0.644}
\newcommand{\NabUnderwaterLateAlmostOpenNab}{0.954}
\newcommand{\NabUnderwaterLateAlmostOpenFallback}{0.046}
\newcommand{\NabRobotEarlyConservativeGain}{-10.4}
\newcommand{\NabRobotEarlyConservativeGainCI}{2.4}
\newcommand{\NabRobotEarlyConservativeGoal}{1.000}
\newcommand{\NabRobotEarlyConservativeNab}{0.289}
\newcommand{\NabRobotEarlyConservativeFallback}{0.711}
\newcommand{\NabRobotEarlyGuardedGain}{0.0}
\newcommand{\NabRobotEarlyGuardedGainCI}{0.0}
\newcommand{\NabRobotEarlyGuardedGoal}{1.000}
\newcommand{\NabRobotEarlyGuardedNab}{0.438}
\newcommand{\NabRobotEarlyGuardedFallback}{0.562}
\newcommand{\NabRobotEarlyModerateGain}{0.0}
\newcommand{\NabRobotEarlyModerateGainCI}{0.0}
\newcommand{\NabRobotEarlyModerateGoal}{1.000}
\newcommand{\NabRobotEarlyModerateNab}{0.534}
\newcommand{\NabRobotEarlyModerateFallback}{0.466}
\newcommand{\NabRobotEarlyBalancedGain}{0.0}
\newcommand{\NabRobotEarlyBalancedGainCI}{0.0}
\newcommand{\NabRobotEarlyBalancedGoal}{1.000}
\newcommand{\NabRobotEarlyBalancedNab}{0.561}
\newcommand{\NabRobotEarlyBalancedFallback}{0.439}
\newcommand{\NabRobotEarlyHighGain}{0.0}
\newcommand{\NabRobotEarlyHighGainCI}{0.0}
\newcommand{\NabRobotEarlyHighGoal}{0.189}
\newcommand{\NabRobotEarlyHighNab}{0.712}
\newcommand{\NabRobotEarlyHighFallback}{0.288}
\newcommand{\NabRobotEarlyAlmostOpenGain}{0.0}
\newcommand{\NabRobotEarlyAlmostOpenGainCI}{0.0}
\newcommand{\NabRobotEarlyAlmostOpenGoal}{0.167}
\newcommand{\NabRobotEarlyAlmostOpenNab}{0.952}
\newcommand{\NabRobotEarlyAlmostOpenFallback}{0.048}
\newcommand{\NabRobotMidConservativeGain}{-9.3}
\newcommand{\NabRobotMidConservativeGainCI}{3.0}
\newcommand{\NabRobotMidConservativeGoal}{1.000}
\newcommand{\NabRobotMidConservativeNab}{0.307}
\newcommand{\NabRobotMidConservativeFallback}{0.693}
\newcommand{\NabRobotMidGuardedGain}{-1.2}
\newcommand{\NabRobotMidGuardedGainCI}{2.4}
\newcommand{\NabRobotMidGuardedGoal}{1.000}
\newcommand{\NabRobotMidGuardedNab}{0.440}
\newcommand{\NabRobotMidGuardedFallback}{0.560}
\newcommand{\NabRobotMidModerateGain}{0.0}
\newcommand{\NabRobotMidModerateGainCI}{0.0}
\newcommand{\NabRobotMidModerateGoal}{1.000}
\newcommand{\NabRobotMidModerateNab}{0.531}
\newcommand{\NabRobotMidModerateFallback}{0.469}
\newcommand{\NabRobotMidBalancedGain}{-0.0}
\newcommand{\NabRobotMidBalancedGainCI}{0.0}
\newcommand{\NabRobotMidBalancedGoal}{0.933}
\newcommand{\NabRobotMidBalancedNab}{0.557}
\newcommand{\NabRobotMidBalancedFallback}{0.443}
\newcommand{\NabRobotMidHighGain}{-0.0}
\newcommand{\NabRobotMidHighGainCI}{0.0}
\newcommand{\NabRobotMidHighGoal}{0.156}
\newcommand{\NabRobotMidHighNab}{0.710}
\newcommand{\NabRobotMidHighFallback}{0.290}
\newcommand{\NabRobotMidAlmostOpenGain}{-0.0}
\newcommand{\NabRobotMidAlmostOpenGainCI}{0.0}
\newcommand{\NabRobotMidAlmostOpenGoal}{0.156}
\newcommand{\NabRobotMidAlmostOpenNab}{0.952}
\newcommand{\NabRobotMidAlmostOpenFallback}{0.048}
\newcommand{\NabRobotLateConservativeGain}{-7.6}
\newcommand{\NabRobotLateConservativeGainCI}{1.9}
\newcommand{\NabRobotLateConservativeGoal}{1.000}
\newcommand{\NabRobotLateConservativeNab}{0.299}
\newcommand{\NabRobotLateConservativeFallback}{0.701}
\newcommand{\NabRobotLateGuardedGain}{0.0}
\newcommand{\NabRobotLateGuardedGainCI}{0.0}
\newcommand{\NabRobotLateGuardedGoal}{1.000}
\newcommand{\NabRobotLateGuardedNab}{0.467}
\newcommand{\NabRobotLateGuardedFallback}{0.533}
\newcommand{\NabRobotLateModerateGain}{0.0}
\newcommand{\NabRobotLateModerateGainCI}{0.0}
\newcommand{\NabRobotLateModerateGoal}{1.000}
\newcommand{\NabRobotLateModerateNab}{0.555}
\newcommand{\NabRobotLateModerateFallback}{0.445}
\newcommand{\NabRobotLateBalancedGain}{-0.0}
\newcommand{\NabRobotLateBalancedGainCI}{0.0}
\newcommand{\NabRobotLateBalancedGoal}{0.967}
\newcommand{\NabRobotLateBalancedNab}{0.574}
\newcommand{\NabRobotLateBalancedFallback}{0.426}
\newcommand{\NabRobotLateHighGain}{0.0}
\newcommand{\NabRobotLateHighGainCI}{0.0}
\newcommand{\NabRobotLateHighGoal}{0.400}
\newcommand{\NabRobotLateHighNab}{0.725}
\newcommand{\NabRobotLateHighFallback}{0.275}
\newcommand{\NabRobotLateAlmostOpenGain}{0.0}
\newcommand{\NabRobotLateAlmostOpenGainCI}{0.0}
\newcommand{\NabRobotLateAlmostOpenGoal}{0.389}
\newcommand{\NabRobotLateAlmostOpenNab}{0.955}
\newcommand{\NabRobotLateAlmostOpenFallback}{0.045}
\newcommand{\NabPendulumLowConservativeGain}{-0.02}
\newcommand{\NabPendulumLowConservativeGainCI}{0.02}
\newcommand{\NabPendulumLowConservativeGoal}{0.970}
\newcommand{\NabPendulumLowConservativeNab}{0.040}
\newcommand{\NabPendulumLowConservativeFallback}{0.960}
\newcommand{\NabPendulumLowGuardedGain}{-0.08}
\newcommand{\NabPendulumLowGuardedGainCI}{0.03}
\newcommand{\NabPendulumLowGuardedGoal}{0.980}
\newcommand{\NabPendulumLowGuardedNab}{0.134}
\newcommand{\NabPendulumLowGuardedFallback}{0.866}
\newcommand{\NabPendulumLowModerateGain}{-0.26}
\newcommand{\NabPendulumLowModerateGainCI}{0.06}
\newcommand{\NabPendulumLowModerateGoal}{0.807}
\newcommand{\NabPendulumLowModerateNab}{0.376}
\newcommand{\NabPendulumLowModerateFallback}{0.624}
\newcommand{\NabPendulumLowBalancedGain}{-0.38}
\newcommand{\NabPendulumLowBalancedGainCI}{0.06}
\newcommand{\NabPendulumLowBalancedGoal}{0.473}
\newcommand{\NabPendulumLowBalancedNab}{0.515}
\newcommand{\NabPendulumLowBalancedFallback}{0.485}
\newcommand{\NabPendulumLowHighGain}{-0.53}
\newcommand{\NabPendulumLowHighGainCI}{0.04}
\newcommand{\NabPendulumLowHighGoal}{0.067}
\newcommand{\NabPendulumLowHighNab}{0.706}
\newcommand{\NabPendulumLowHighFallback}{0.294}
\newcommand{\NabPendulumLowAlmostOpenGain}{-0.64}
\newcommand{\NabPendulumLowAlmostOpenGainCI}{0.03}
\newcommand{\NabPendulumLowAlmostOpenGoal}{0.000}
\newcommand{\NabPendulumLowAlmostOpenNab}{0.952}
\newcommand{\NabPendulumLowAlmostOpenFallback}{0.048}
\newcommand{\NabPendulumMidReturnConservativeGain}{0.04}
\newcommand{\NabPendulumMidReturnConservativeGainCI}{0.04}
\newcommand{\NabPendulumMidReturnConservativeGoal}{0.963}
\newcommand{\NabPendulumMidReturnConservativeNab}{0.127}
\newcommand{\NabPendulumMidReturnConservativeFallback}{0.873}
\newcommand{\NabPendulumMidReturnGuardedGain}{0.02}
\newcommand{\NabPendulumMidReturnGuardedGainCI}{0.04}
\newcommand{\NabPendulumMidReturnGuardedGoal}{0.977}
\newcommand{\NabPendulumMidReturnGuardedNab}{0.191}
\newcommand{\NabPendulumMidReturnGuardedFallback}{0.809}
\newcommand{\NabPendulumMidReturnModerateGain}{-0.10}
\newcommand{\NabPendulumMidReturnModerateGainCI}{0.04}
\newcommand{\NabPendulumMidReturnModerateGoal}{0.847}
\newcommand{\NabPendulumMidReturnModerateNab}{0.400}
\newcommand{\NabPendulumMidReturnModerateFallback}{0.600}
\newcommand{\NabPendulumMidReturnBalancedGain}{-0.20}
\newcommand{\NabPendulumMidReturnBalancedGainCI}{0.04}
\newcommand{\NabPendulumMidReturnBalancedGoal}{0.507}
\newcommand{\NabPendulumMidReturnBalancedNab}{0.530}
\newcommand{\NabPendulumMidReturnBalancedFallback}{0.470}
\newcommand{\NabPendulumMidReturnHighGain}{-0.28}
\newcommand{\NabPendulumMidReturnHighGainCI}{0.02}
\newcommand{\NabPendulumMidReturnHighGoal}{0.127}
\newcommand{\NabPendulumMidReturnHighNab}{0.714}
\newcommand{\NabPendulumMidReturnHighFallback}{0.286}
\newcommand{\NabPendulumMidReturnAlmostOpenGain}{-0.33}
\newcommand{\NabPendulumMidReturnAlmostOpenGainCI}{0.02}
\newcommand{\NabPendulumMidReturnAlmostOpenGoal}{0.000}
\newcommand{\NabPendulumMidReturnAlmostOpenNab}{0.953}
\newcommand{\NabPendulumMidReturnAlmostOpenFallback}{0.047}
\newcommand{\NabPendulumHighConservativeGain}{0.11}
\newcommand{\NabPendulumHighConservativeGainCI}{0.02}
\newcommand{\NabPendulumHighConservativeGoal}{1.000}
\newcommand{\NabPendulumHighConservativeNab}{0.128}
\newcommand{\NabPendulumHighConservativeFallback}{0.872}
\newcommand{\NabPendulumHighGuardedGain}{0.12}
\newcommand{\NabPendulumHighGuardedGainCI}{0.01}
\newcommand{\NabPendulumHighGuardedGoal}{1.000}
\newcommand{\NabPendulumHighGuardedNab}{0.181}
\newcommand{\NabPendulumHighGuardedFallback}{0.819}
\newcommand{\NabPendulumHighModerateGain}{0.10}
\newcommand{\NabPendulumHighModerateGainCI}{0.02}
\newcommand{\NabPendulumHighModerateGoal}{0.997}
\newcommand{\NabPendulumHighModerateNab}{0.394}
\newcommand{\NabPendulumHighModerateFallback}{0.606}
\newcommand{\NabPendulumHighBalancedGain}{0.08}
\newcommand{\NabPendulumHighBalancedGainCI}{0.03}
\newcommand{\NabPendulumHighBalancedGoal}{0.970}
\newcommand{\NabPendulumHighBalancedNab}{0.528}
\newcommand{\NabPendulumHighBalancedFallback}{0.472}
\newcommand{\NabPendulumHighHighGain}{0.07}
\newcommand{\NabPendulumHighHighGainCI}{0.03}
\newcommand{\NabPendulumHighHighGoal}{0.877}
\newcommand{\NabPendulumHighHighNab}{0.714}
\newcommand{\NabPendulumHighHighFallback}{0.286}
\newcommand{\NabPendulumHighAlmostOpenGain}{0.04}
\newcommand{\NabPendulumHighAlmostOpenGainCI}{0.03}
\newcommand{\NabPendulumHighAlmostOpenGoal}{0.230}
\newcommand{\NabPendulumHighAlmostOpenNab}{0.953}
\newcommand{\NabPendulumHighAlmostOpenFallback}{0.047}
\newcommand{\NabCartpoleLowConservativeGain}{-0.05}
\newcommand{\NabCartpoleLowConservativeGainCI}{0.03}
\newcommand{\NabCartpoleLowConservativeGoal}{0.594}
\newcommand{\NabCartpoleLowConservativeNab}{0.069}
\newcommand{\NabCartpoleLowConservativeFallback}{0.931}
\newcommand{\NabCartpoleLowGuardedGain}{-0.03}
\newcommand{\NabCartpoleLowGuardedGainCI}{0.04}
\newcommand{\NabCartpoleLowGuardedGoal}{0.818}
\newcommand{\NabCartpoleLowGuardedNab}{0.152}
\newcommand{\NabCartpoleLowGuardedFallback}{0.848}
\newcommand{\NabCartpoleLowModerateGain}{-0.12}
\newcommand{\NabCartpoleLowModerateGainCI}{0.06}
\newcommand{\NabCartpoleLowModerateGoal}{0.864}
\newcommand{\NabCartpoleLowModerateNab}{0.381}
\newcommand{\NabCartpoleLowModerateFallback}{0.619}
\newcommand{\NabCartpoleLowBalancedGain}{-0.21}
\newcommand{\NabCartpoleLowBalancedGainCI}{0.07}
\newcommand{\NabCartpoleLowBalancedGoal}{0.748}
\newcommand{\NabCartpoleLowBalancedNab}{0.533}
\newcommand{\NabCartpoleLowBalancedFallback}{0.467}
\newcommand{\NabCartpoleLowHighGain}{-0.42}
\newcommand{\NabCartpoleLowHighGainCI}{0.10}
\newcommand{\NabCartpoleLowHighGoal}{0.300}
\newcommand{\NabCartpoleLowHighNab}{0.731}
\newcommand{\NabCartpoleLowHighFallback}{0.269}
\newcommand{\NabCartpoleLowAlmostOpenGain}{-0.64}
\newcommand{\NabCartpoleLowAlmostOpenGainCI}{0.08}
\newcommand{\NabCartpoleLowAlmostOpenGoal}{0.000}
\newcommand{\NabCartpoleLowAlmostOpenNab}{0.957}
\newcommand{\NabCartpoleLowAlmostOpenFallback}{0.043}
\newcommand{\NabCartpoleMidReturnConservativeGain}{0.01}
\newcommand{\NabCartpoleMidReturnConservativeGainCI}{0.04}
\newcommand{\NabCartpoleMidReturnConservativeGoal}{0.627}
\newcommand{\NabCartpoleMidReturnConservativeNab}{0.113}
\newcommand{\NabCartpoleMidReturnConservativeFallback}{0.887}
\newcommand{\NabCartpoleMidReturnGuardedGain}{0.08}
\newcommand{\NabCartpoleMidReturnGuardedGainCI}{0.05}
\newcommand{\NabCartpoleMidReturnGuardedGoal}{0.894}
\newcommand{\NabCartpoleMidReturnGuardedNab}{0.169}
\newcommand{\NabCartpoleMidReturnGuardedFallback}{0.831}
\newcommand{\NabCartpoleMidReturnModerateGain}{0.02}
\newcommand{\NabCartpoleMidReturnModerateGainCI}{0.06}
\newcommand{\NabCartpoleMidReturnModerateGoal}{0.930}
\newcommand{\NabCartpoleMidReturnModerateNab}{0.386}
\newcommand{\NabCartpoleMidReturnModerateFallback}{0.614}
\newcommand{\NabCartpoleMidReturnBalancedGain}{-0.08}
\newcommand{\NabCartpoleMidReturnBalancedGainCI}{0.08}
\newcommand{\NabCartpoleMidReturnBalancedGoal}{0.839}
\newcommand{\NabCartpoleMidReturnBalancedNab}{0.527}
\newcommand{\NabCartpoleMidReturnBalancedFallback}{0.473}
\newcommand{\NabCartpoleMidReturnHighGain}{-0.24}
\newcommand{\NabCartpoleMidReturnHighGainCI}{0.14}
\newcommand{\NabCartpoleMidReturnHighGoal}{0.400}
\newcommand{\NabCartpoleMidReturnHighNab}{0.713}
\newcommand{\NabCartpoleMidReturnHighFallback}{0.287}
\newcommand{\NabCartpoleMidReturnAlmostOpenGain}{-0.30}
\newcommand{\NabCartpoleMidReturnAlmostOpenGainCI}{0.04}
\newcommand{\NabCartpoleMidReturnAlmostOpenGoal}{0.015}
\newcommand{\NabCartpoleMidReturnAlmostOpenNab}{0.953}
\newcommand{\NabCartpoleMidReturnAlmostOpenFallback}{0.047}
\newcommand{\NabCartpoleHighConservativeGain}{0.07}
\newcommand{\NabCartpoleHighConservativeGainCI}{0.01}
\newcommand{\NabCartpoleHighConservativeGoal}{0.661}
\newcommand{\NabCartpoleHighConservativeNab}{0.137}
\newcommand{\NabCartpoleHighConservativeFallback}{0.863}
\newcommand{\NabCartpoleHighGuardedGain}{0.15}
\newcommand{\NabCartpoleHighGuardedGainCI}{0.02}
\newcommand{\NabCartpoleHighGuardedGoal}{0.958}
\newcommand{\NabCartpoleHighGuardedNab}{0.178}
\newcommand{\NabCartpoleHighGuardedFallback}{0.822}
\newcommand{\NabCartpoleHighModerateGain}{0.13}
\newcommand{\NabCartpoleHighModerateGainCI}{0.02}
\newcommand{\NabCartpoleHighModerateGoal}{0.864}
\newcommand{\NabCartpoleHighModerateNab}{0.389}
\newcommand{\NabCartpoleHighModerateFallback}{0.611}
\newcommand{\NabCartpoleHighBalancedGain}{0.12}
\newcommand{\NabCartpoleHighBalancedGainCI}{0.03}
\newcommand{\NabCartpoleHighBalancedGoal}{0.791}
\newcommand{\NabCartpoleHighBalancedNab}{0.530}
\newcommand{\NabCartpoleHighBalancedFallback}{0.470}
\newcommand{\NabCartpoleHighHighGain}{0.10}
\newcommand{\NabCartpoleHighHighGainCI}{0.04}
\newcommand{\NabCartpoleHighHighGoal}{0.652}
\newcommand{\NabCartpoleHighHighNab}{0.716}
\newcommand{\NabCartpoleHighHighFallback}{0.284}
\newcommand{\NabCartpoleHighAlmostOpenGain}{0.07}
\newcommand{\NabCartpoleHighAlmostOpenGainCI}{0.02}
\newcommand{\NabCartpoleHighAlmostOpenGoal}{0.352}
\newcommand{\NabCartpoleHighAlmostOpenNab}{0.953}
\newcommand{\NabCartpoleHighAlmostOpenFallback}{0.047}
\newcommand{\NabUnderwaterLowConservativeGain}{-0.07}
\newcommand{\NabUnderwaterLowConservativeGainCI}{0.05}
\newcommand{\NabUnderwaterLowConservativeGoal}{1.000}
\newcommand{\NabUnderwaterLowConservativeNab}{0.125}
\newcommand{\NabUnderwaterLowConservativeFallback}{0.875}
\newcommand{\NabUnderwaterLowGuardedGain}{-0.16}
\newcommand{\NabUnderwaterLowGuardedGainCI}{0.06}
\newcommand{\NabUnderwaterLowGuardedGoal}{1.000}
\newcommand{\NabUnderwaterLowGuardedNab}{0.185}
\newcommand{\NabUnderwaterLowGuardedFallback}{0.815}
\newcommand{\NabUnderwaterLowModerateGain}{-0.40}
\newcommand{\NabUnderwaterLowModerateGainCI}{0.01}
\newcommand{\NabUnderwaterLowModerateGoal}{0.889}
\newcommand{\NabUnderwaterLowModerateNab}{0.412}
\newcommand{\NabUnderwaterLowModerateFallback}{0.588}
\newcommand{\NabUnderwaterLowBalancedGain}{-0.66}
\newcommand{\NabUnderwaterLowBalancedGainCI}{0.09}
\newcommand{\NabUnderwaterLowBalancedGoal}{0.378}
\newcommand{\NabUnderwaterLowBalancedNab}{0.545}
\newcommand{\NabUnderwaterLowBalancedFallback}{0.455}
\newcommand{\NabUnderwaterLowHighGain}{-0.70}
\newcommand{\NabUnderwaterLowHighGainCI}{0.09}
\newcommand{\NabUnderwaterLowHighGoal}{0.222}
\newcommand{\NabUnderwaterLowHighNab}{0.721}
\newcommand{\NabUnderwaterLowHighFallback}{0.279}
\newcommand{\NabUnderwaterLowAlmostOpenGain}{-0.74}
\newcommand{\NabUnderwaterLowAlmostOpenGainCI}{0.04}
\newcommand{\NabUnderwaterLowAlmostOpenGoal}{0.167}
\newcommand{\NabUnderwaterLowAlmostOpenNab}{0.953}
\newcommand{\NabUnderwaterLowAlmostOpenFallback}{0.047}
\newcommand{\NabUnderwaterMidReturnConservativeGain}{-0.03}
\newcommand{\NabUnderwaterMidReturnConservativeGainCI}{0.01}
\newcommand{\NabUnderwaterMidReturnConservativeGoal}{1.000}
\newcommand{\NabUnderwaterMidReturnConservativeNab}{0.112}
\newcommand{\NabUnderwaterMidReturnConservativeFallback}{0.888}
\newcommand{\NabUnderwaterMidReturnGuardedGain}{-0.07}
\newcommand{\NabUnderwaterMidReturnGuardedGainCI}{0.03}
\newcommand{\NabUnderwaterMidReturnGuardedGoal}{1.000}
\newcommand{\NabUnderwaterMidReturnGuardedNab}{0.176}
\newcommand{\NabUnderwaterMidReturnGuardedFallback}{0.824}
\newcommand{\NabUnderwaterMidReturnModerateGain}{-0.15}
\newcommand{\NabUnderwaterMidReturnModerateGainCI}{0.07}
\newcommand{\NabUnderwaterMidReturnModerateGoal}{0.933}
\newcommand{\NabUnderwaterMidReturnModerateNab}{0.399}
\newcommand{\NabUnderwaterMidReturnModerateFallback}{0.601}
\newcommand{\NabUnderwaterMidReturnBalancedGain}{-0.23}
\newcommand{\NabUnderwaterMidReturnBalancedGainCI}{0.15}
\newcommand{\NabUnderwaterMidReturnBalancedGoal}{0.867}
\newcommand{\NabUnderwaterMidReturnBalancedNab}{0.529}
\newcommand{\NabUnderwaterMidReturnBalancedFallback}{0.471}
\newcommand{\NabUnderwaterMidReturnHighGain}{-0.31}
\newcommand{\NabUnderwaterMidReturnHighGainCI}{0.12}
\newcommand{\NabUnderwaterMidReturnHighGoal}{0.533}
\newcommand{\NabUnderwaterMidReturnHighNab}{0.717}
\newcommand{\NabUnderwaterMidReturnHighFallback}{0.283}
\newcommand{\NabUnderwaterMidReturnAlmostOpenGain}{-0.37}
\newcommand{\NabUnderwaterMidReturnAlmostOpenGainCI}{0.10}
\newcommand{\NabUnderwaterMidReturnAlmostOpenGoal}{0.033}
\newcommand{\NabUnderwaterMidReturnAlmostOpenNab}{0.954}
\newcommand{\NabUnderwaterMidReturnAlmostOpenFallback}{0.046}
\newcommand{\NabUnderwaterHighConservativeGain}{-0.02}
\newcommand{\NabUnderwaterHighConservativeGainCI}{0.01}
\newcommand{\NabUnderwaterHighConservativeGoal}{1.000}
\newcommand{\NabUnderwaterHighConservativeNab}{0.124}
\newcommand{\NabUnderwaterHighConservativeFallback}{0.876}
\newcommand{\NabUnderwaterHighGuardedGain}{-0.08}
\newcommand{\NabUnderwaterHighGuardedGainCI}{0.06}
\newcommand{\NabUnderwaterHighGuardedGoal}{0.978}
\newcommand{\NabUnderwaterHighGuardedNab}{0.202}
\newcommand{\NabUnderwaterHighGuardedFallback}{0.798}
\newcommand{\NabUnderwaterHighModerateGain}{-0.13}
\newcommand{\NabUnderwaterHighModerateGainCI}{0.14}
\newcommand{\NabUnderwaterHighModerateGoal}{0.922}
\newcommand{\NabUnderwaterHighModerateNab}{0.427}
\newcommand{\NabUnderwaterHighModerateFallback}{0.573}
\newcommand{\NabUnderwaterHighBalancedGain}{-0.12}
\newcommand{\NabUnderwaterHighBalancedGainCI}{0.10}
\newcommand{\NabUnderwaterHighBalancedGoal}{0.978}
\newcommand{\NabUnderwaterHighBalancedNab}{0.561}
\newcommand{\NabUnderwaterHighBalancedFallback}{0.439}
\newcommand{\NabUnderwaterHighHighGain}{-0.12}
\newcommand{\NabUnderwaterHighHighGainCI}{0.08}
\newcommand{\NabUnderwaterHighHighGoal}{0.833}
\newcommand{\NabUnderwaterHighHighNab}{0.741}
\newcommand{\NabUnderwaterHighHighFallback}{0.259}
\newcommand{\NabUnderwaterHighAlmostOpenGain}{-0.17}
\newcommand{\NabUnderwaterHighAlmostOpenGainCI}{0.01}
\newcommand{\NabUnderwaterHighAlmostOpenGoal}{0.522}
\newcommand{\NabUnderwaterHighAlmostOpenNab}{0.958}
\newcommand{\NabUnderwaterHighAlmostOpenFallback}{0.042}
\newcommand{\NabRobotLowConservativeGain}{0.03}
\newcommand{\NabRobotLowConservativeGainCI}{0.01}
\newcommand{\NabRobotLowConservativeGoal}{1.000}
\newcommand{\NabRobotLowConservativeNab}{0.108}
\newcommand{\NabRobotLowConservativeFallback}{0.892}
\newcommand{\NabRobotLowGuardedGain}{-0.17}
\newcommand{\NabRobotLowGuardedGainCI}{0.12}
\newcommand{\NabRobotLowGuardedGoal}{1.000}
\newcommand{\NabRobotLowGuardedNab}{0.414}
\newcommand{\NabRobotLowGuardedFallback}{0.586}
\newcommand{\NabRobotLowModerateGain}{-0.45}
\newcommand{\NabRobotLowModerateGainCI}{0.22}
\newcommand{\NabRobotLowModerateGoal}{1.000}
\newcommand{\NabRobotLowModerateNab}{0.557}
\newcommand{\NabRobotLowModerateFallback}{0.443}
\newcommand{\NabRobotLowBalancedGain}{-0.62}
\newcommand{\NabRobotLowBalancedGainCI}{0.27}
\newcommand{\NabRobotLowBalancedGoal}{0.956}
\newcommand{\NabRobotLowBalancedNab}{0.597}
\newcommand{\NabRobotLowBalancedFallback}{0.403}
\newcommand{\NabRobotLowHighGain}{-0.70}
\newcommand{\NabRobotLowHighGainCI}{0.15}
\newcommand{\NabRobotLowHighGoal}{0.300}
\newcommand{\NabRobotLowHighNab}{0.719}
\newcommand{\NabRobotLowHighFallback}{0.281}
\newcommand{\NabRobotLowAlmostOpenGain}{-0.79}
\newcommand{\NabRobotLowAlmostOpenGainCI}{0.11}
\newcommand{\NabRobotLowAlmostOpenGoal}{0.033}
\newcommand{\NabRobotLowAlmostOpenNab}{0.952}
\newcommand{\NabRobotLowAlmostOpenFallback}{0.048}
\newcommand{\NabRobotMidReturnConservativeGain}{0.06}
\newcommand{\NabRobotMidReturnConservativeGainCI}{0.01}
\newcommand{\NabRobotMidReturnConservativeGoal}{1.000}
\newcommand{\NabRobotMidReturnConservativeNab}{0.220}
\newcommand{\NabRobotMidReturnConservativeFallback}{0.780}
\newcommand{\NabRobotMidReturnGuardedGain}{0.01}
\newcommand{\NabRobotMidReturnGuardedGainCI}{0.04}
\newcommand{\NabRobotMidReturnGuardedGoal}{1.000}
\newcommand{\NabRobotMidReturnGuardedNab}{0.460}
\newcommand{\NabRobotMidReturnGuardedFallback}{0.540}
\newcommand{\NabRobotMidReturnModerateGain}{0.00}
\newcommand{\NabRobotMidReturnModerateGainCI}{0.11}
\newcommand{\NabRobotMidReturnModerateGoal}{1.000}
\newcommand{\NabRobotMidReturnModerateNab}{0.556}
\newcommand{\NabRobotMidReturnModerateFallback}{0.444}
\newcommand{\NabRobotMidReturnBalancedGain}{0.02}
\newcommand{\NabRobotMidReturnBalancedGainCI}{0.11}
\newcommand{\NabRobotMidReturnBalancedGoal}{0.944}
\newcommand{\NabRobotMidReturnBalancedNab}{0.584}
\newcommand{\NabRobotMidReturnBalancedFallback}{0.416}
\newcommand{\NabRobotMidReturnHighGain}{0.02}
\newcommand{\NabRobotMidReturnHighGainCI}{0.10}
\newcommand{\NabRobotMidReturnHighGoal}{0.256}
\newcommand{\NabRobotMidReturnHighNab}{0.716}
\newcommand{\NabRobotMidReturnHighFallback}{0.284}
\newcommand{\NabRobotMidReturnAlmostOpenGain}{0.02}
\newcommand{\NabRobotMidReturnAlmostOpenGainCI}{0.12}
\newcommand{\NabRobotMidReturnAlmostOpenGoal}{0.189}
\newcommand{\NabRobotMidReturnAlmostOpenNab}{0.953}
\newcommand{\NabRobotMidReturnAlmostOpenFallback}{0.047}
\newcommand{\NabRobotHighConservativeGain}{0.09}
\newcommand{\NabRobotHighConservativeGainCI}{0.02}
\newcommand{\NabRobotHighConservativeGoal}{1.000}
\newcommand{\NabRobotHighConservativeNab}{0.260}
\newcommand{\NabRobotHighConservativeFallback}{0.740}
\newcommand{\NabRobotHighGuardedGain}{0.10}
\newcommand{\NabRobotHighGuardedGainCI}{0.03}
\newcommand{\NabRobotHighGuardedGoal}{1.000}
\newcommand{\NabRobotHighGuardedNab}{0.449}
\newcommand{\NabRobotHighGuardedFallback}{0.551}
\newcommand{\NabRobotHighModerateGain}{0.12}
\newcommand{\NabRobotHighModerateGainCI}{0.04}
\newcommand{\NabRobotHighModerateGoal}{1.000}
\newcommand{\NabRobotHighModerateNab}{0.543}
\newcommand{\NabRobotHighModerateFallback}{0.457}
\newcommand{\NabRobotHighBalancedGain}{0.12}
\newcommand{\NabRobotHighBalancedGainCI}{0.03}
\newcommand{\NabRobotHighBalancedGoal}{0.967}
\newcommand{\NabRobotHighBalancedNab}{0.573}
\newcommand{\NabRobotHighBalancedFallback}{0.427}
\newcommand{\NabRobotHighHighGain}{0.12}
\newcommand{\NabRobotHighHighGainCI}{0.03}
\newcommand{\NabRobotHighHighGoal}{0.200}
\newcommand{\NabRobotHighHighNab}{0.714}
\newcommand{\NabRobotHighHighFallback}{0.286}
\newcommand{\NabRobotHighAlmostOpenGain}{0.12}
\newcommand{\NabRobotHighAlmostOpenGainCI}{0.02}
\newcommand{\NabRobotHighAlmostOpenGoal}{0.144}
\newcommand{\NabRobotHighAlmostOpenNab}{0.952}
\newcommand{\NabRobotHighAlmostOpenFallback}{0.048}

The experimental study addresses the following questions:
\begin{itemize}
  \item How does CALF-Wrapper respond to changes in the quality of
    the supplied actor--critic checkpoint?
    Evaluating actor--critic checkpoints obtained at different stages
    of training and exhibiting low, intermediate, and high observed
    returns shows whether CALF-Wrapper can compensate for weak base
    policies through the fallback policy and benefit from
    better-trained base policies while using their corresponding
    critic estimates.
  \item How do different configurations of the policy-selection
    mechanism affect episode return, goal-reaching rate, and the
    frequency with which the fallback policy supplies the applied action?
  \item Does CALF-Wrapper exhibit consistent behavior across
    different environments and with base policies trained using
    different actor--critic algorithms, specifically PPO and TD3?
  \item How does CALF-Wrapper compare with the base policy, the
    fallback policy, and the PPO-Lagrangian and TD3-Lagrangian
    baselines implemented in this article?
  \item How sensitive is CALF-Wrapper to the critic threshold $\nu$
    and to perturbations of the critic estimate?
  \item What computational overhead does CALF-Wrapper introduce
    during deployment?
\end{itemize}

The remainder of this section is organized as follows:
\begin{itemize}
  \item \Cref{sec:environments} defines the four environments in
    which CALF-Wrapper is evaluated.
  \item \Cref{sec:return_checkpoint_selection} states which
    algorithms were used to train the base policies and defines the
    Low-, Mid-, and High-return checkpoints selected from the
    resulting sequences of actor--critic checkpoints.
  \item \Cref{sec:operating_modes} defines a broad range of operating
    modes for evaluating CALF-Wrapper.
    Each mode is specified by a target lower bound on the expected
    fraction of episode steps assigned to the base policy and by
    $p_{\mathrm{relax}}$ and $\lambda$.
    For each positive target lower bound, the reported modes set
    $p_{\mathrm{relax}}=1$ and determine $\lambda$ from that target.
    An additional sensitivity study varies $p_{\mathrm{relax}}$ while
    holding the target value fixed to support this choice.
  \item \Cref{sec:nu_calibration} examines the remaining
    hyperparameter, the critic threshold $\nu$.
    The sensitivity and ablation study relates $\nu$ to the reward
    scale of the problem and determines the values used in the main evaluation.
  \item \Cref{sec:experimental_results} reports the main evaluation
    across the Low-, Mid-, and High-return checkpoints and the six
    operating modes.
    This evaluation compares CALF-Wrapper with the base policy, the
    fallback policy, and the PPO-Lagrangian and TD3-Lagrangian baselines.
    The results show how the observed behavior of CALF-Wrapper
    changes with the quality of the supplied checkpoint and with the
    selected operating mode.
  \item Finally, \Cref{sec:critic_noise_robustness} evaluates
    CALF-Wrapper under perturbations of the critic estimate, and
    \Cref{sec:policy_selection_latency} measures its per-step latency
    and computational overhead during deployment.
\end{itemize}

\subsection{Environments}
\label{sec:environments}
The evaluation considers four control environments spanning classic
control and continuous robotic navigation.
Complete descriptions of their dynamics, observations, actions,
reward functions, goal sets, and initial conditions
are provided in Appendix~\ref{app:environment_descriptions}.
The implementations of all fallback policies are
available in the project repository \calfwrepo.
\begin{itemize}
  \item \textbf{Pendulum}, in which a control torque must stabilize
    the pendulum in the upright position, is described in
    Appendix~\ref{sec:pendulum_env} and \Cref{fig:gym_env_pendulum}.
  \item \textbf{CartPole Swing-Up}, in which a horizontal force must
    swing the pole to the upright position while keeping the cart
    near the center, is described in Appendix~\ref{sec:cartpole_env} and
    \Cref{fig:gym_env_cartpole}.
  \item \textbf{Contaminated-Zone Autonomous Underwater Vehicle
    Navigation}, in which an AUV must reach a surface opening while
    avoiding a low-reward region, is described in
    Appendix~\ref{sec:underwater_env} and
    \Cref{fig:envunderwaterdrone}.
  \item \textbf{Treasure-Collecting Robot}, in which a constant-speed
    mobile robot can improve its return by intercepting a moving
    treasure before reaching its goal, is described in
    Appendix~\ref{sec:robot_env} and
    \Cref{fig:treasure_robot}.
\end{itemize}
Pendulum and CartPole Swing-Up are classical-control benchmarks.
However, to make the CartPole task more challenging, the standard
task of balancing an initially upright pole was replaced with a swing-up task.
The controller must first raise the pole to the upright position and
then balance it while keeping the cart near the center.
Both Pendulum and CartPole Swing-Up have low-dimensional states and
scalar actions.
Therefore, the evaluation also includes Contaminated-Zone Autonomous
Underwater Vehicle Navigation and Treasure-Collecting Robot, which
have higher-dimensional states and more complex objectives.

\newcommand{\environmentdescriptionsappendix}{%
  \section{Environment Descriptions}
  \label{app:environment_descriptions}
  \subsection{Pendulum Environment}\label{sec:pendulum_env}
  \begin{figure}[H]
    \centering
    \includegraphics[width=0.8\linewidth]{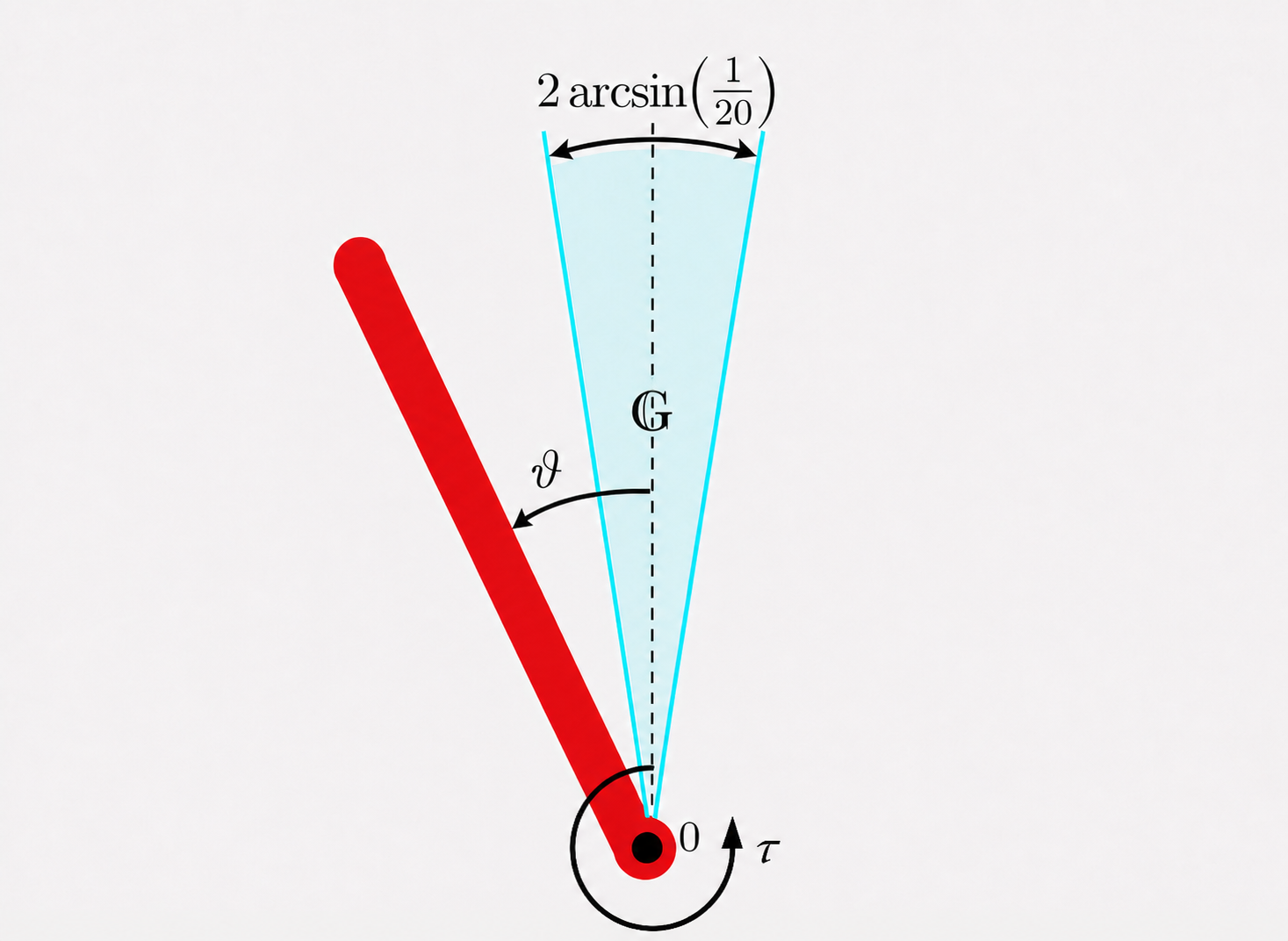}
    \caption{Pendulum environment.
      The state is $(\vartheta,\omega)$, where $\vartheta$ is the
      angle and $\omega$ is the angular velocity.
      The action is the control torque $\tau\in[-2,2]$ N$\cdot$m.
      The shaded sector shows the constraint on $\vartheta$ in the goal
    set $\G$ defined in~\eqref{eq:pendulum_goal_set}.
      Adapted from the official Gymnasium rendering and
      redrawn with an AI image-generation tool (see Acknowledgment).}
    \label{fig:gym_env_pendulum}
  \end{figure}

  The Pendulum~\cite{pendulum-v1} environment from Gymnasium has the
  following dynamics:
  \begin{equation*}
    \dot{\vartheta} = \omega, \quad \dot{\omega} =
    \frac{3g}{2l}\sin(\vartheta) + \frac{3}{ml^2}\tau
  \end{equation*}
  where $\vartheta$ is the angle, $\omega$ is the angular velocity,
  $\tau \in [-2,2]$ N$\cdot$m is the control torque (action),
  $m=1$~kg is the mass of the pendulum, $l=1$~m is the
  length of the pendulum, and $g=10$~m/s$^{2}$ is the
  gravitational acceleration.
  The initial state is randomly sampled from $(\vartheta, \omega)
  \sim \text{Uniform}([-\pi, \pi] \times [-1, 1])$.
  The observation is $[\cos(\vartheta), \sin(\vartheta), \omega]$.
  The goal is to stabilize the pendulum in the upright position,
  which corresponds to $\vartheta=0$.
  The goal set is
  \begin{equation}
    \label{eq:pendulum_goal_set}
    \G := \left\{\max\{|\cos \vartheta - 1|, |\sin\vartheta|\}
    \le\tfrac{1}{20} \land |\omega| \le \tfrac{3}{10}\right\}.
  \end{equation}

  The dynamics are integrated using the Euler scheme at 20 Hz.
  Episodes are truncated after 200 steps.
  The reward function is
  \[
    \begin{aligned}
      r(\vartheta,\omega,\tau)={}&-\arccos^2(\cos\vartheta)
      -0.1\omega^2\\
      &-0.001\tau^2.
    \end{aligned}
  \]
  \subsection{CartPole Swing-Up Environment}\label{sec:cartpole_env}
  \begin{figure}[H]
    \centering
    \includegraphics[width=\linewidth]{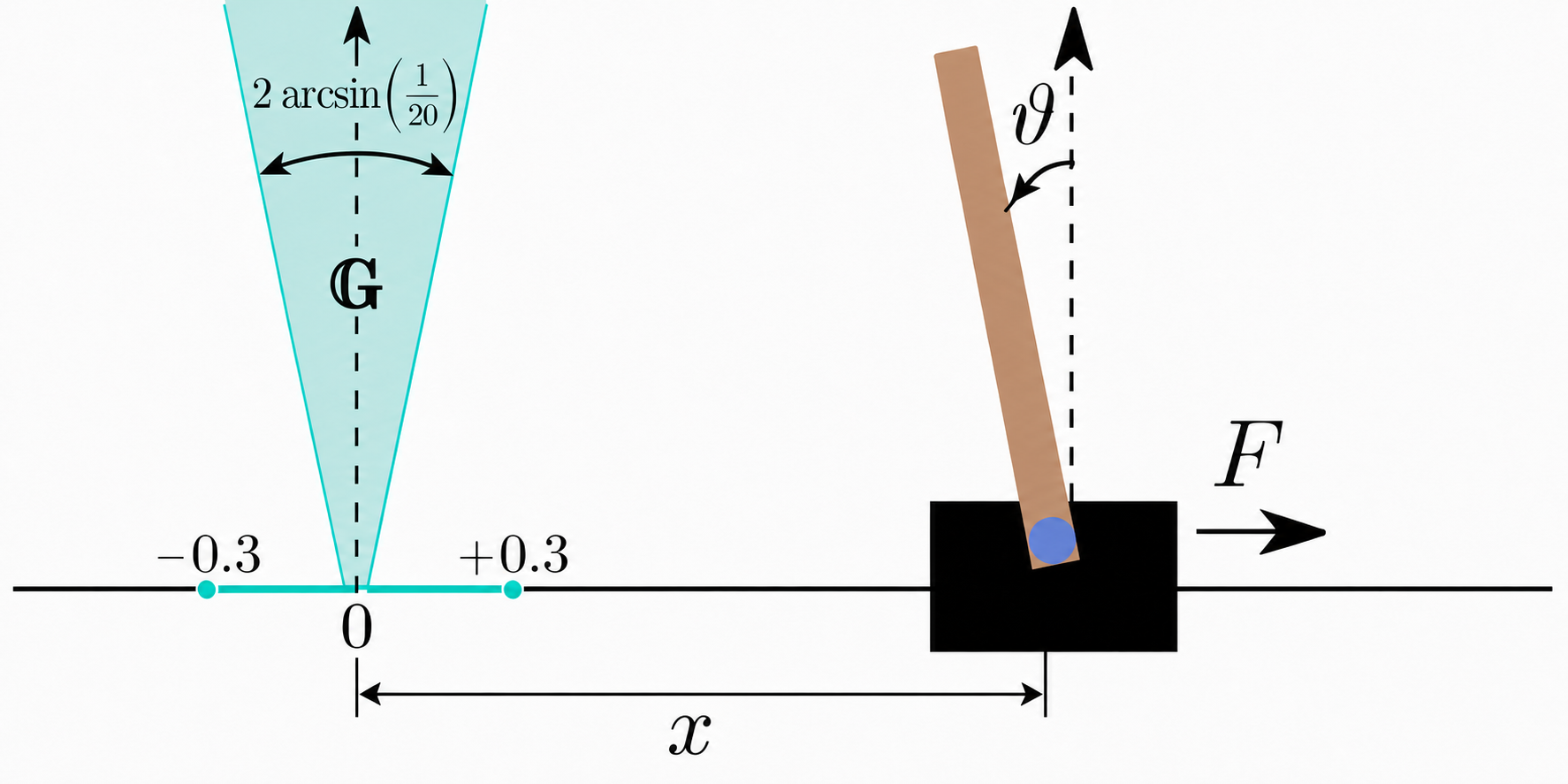}
    \caption{CartPole Swing-Up environment.
      The state is $(\vartheta,\omega,x,v)$, where $\vartheta$ is the
      pole angle, $\omega$ is the pole angular velocity, $x$ is the
      cart position, and $v$ is the cart velocity.
      The action is the force $F\in[-10,10]$ N applied to the cart.
      The shaded regions show the constraints on $x$ and $\vartheta$ in
    the goal set $\G$ defined in~\eqref{eq:cartpole_goal_set}.
      Adapted from the official Gymnasium rendering and
      redrawn with an AI image-generation tool (see Acknowledgment).}
    \label{fig:gym_env_cartpole}
  \end{figure}
  The system dynamics are described by
  \begin{equation*}
    \begin{aligned}
      \dot{x} &= v, &
      \dot{v} &= \frac{F+m_p l\omega^2\sin\vartheta
      -m_p l\dot{\omega}\cos\vartheta}{m_c+m_p}, \\
      \dot{\vartheta} &= \omega, &
      \dot{\omega} &=
      \frac{g\sin\vartheta
      -\cos\vartheta\,
      \dfrac{F+m_p l\omega^2\sin\vartheta}{m_c+m_p}}
      {l\left(\dfrac{4}{3}
      -\dfrac{m_p\cos^2\vartheta}{m_c+m_p}\right)}.
    \end{aligned}
  \end{equation*}
  Here, $x$ is the cart position, $v$ is the cart velocity,
  $\vartheta$ is the pole angle, $\omega$ is the pole angular
  velocity, $m_c=1.0$ kg is the cart mass, $m_p=0.1$ kg is the pole
  mass, $l=0.5$ m is half the pole length, $g=9.8$ m/s$^2$ is the
  gravitational acceleration, and $F \in [-10, 10]$ N is the force
  applied to the cart.
  The initial state is randomly sampled from \((\vartheta,
  \omega,x,v) \sim \text{Uniform}\left([0, 2 \pi] \times [-1, 1]^3\right)\).
  The goal is to swing up and balance the pole in the upright
  position, which corresponds to $\vartheta = 0$, while keeping the
  cart near $x = 0$.
  The goal set is
  \begin{equation}
    \label{eq:cartpole_goal_set}
    \begin{aligned}
      \G = \{\, & \max\{|\cos \vartheta - 1|, |\sin\vartheta|,
        |\omega|\} \le \tfrac{1}{20} \\
      & {} \land \max\{|v|, |x|\} \le \tfrac{3}{10}\,\}.
    \end{aligned}
  \end{equation}

  The system is integrated using the Euler scheme at 50 Hz.
  The PPO base policy and the PPO-Lagrangian baseline are trained with
  the same 1000-step episode limit used during evaluation.
  Episodes do not terminate when the cart position, cart velocity, or
  pole angular velocity reaches a prescribed bound.
  Instead, during training, after every integration
  step, $x$, $v$, and $\omega$ are
  clipped to $[-7.5,7.5]$, $[-12,12]$, and $[-15,15]$, respectively.
  If $x$ is clipped and $v$ points away from the interval, $v$ is set
  to zero.
  During evaluation, the state is not clipped.
  Every training episode and evaluation trial therefore continues
  until the 1000-step time limit.
  The 5D observation is $[x, v, \cos\vartheta, \sin\vartheta,
  \omega]$ with reward function
  \[
    \begin{aligned}
      r(x,\vartheta,v,\omega,F)
      ={}&-\tfrac{1}{2}\arccos^2(\cos\vartheta)
      -\tfrac{1}{2}\min\{x^2,25\}\\
         &-\tfrac{1}{20}\omega^2-\tfrac{1}{20}v^2.
    \end{aligned}
  \]
  The fallback policy applies the maximum force opposite to the cart
  velocity whenever $|v|>4.5$ and $|x|>0.75$ during these evaluations.

  \subsection{Contaminated-Zone Autonomous Underwater Vehicle Navigation}
  \label{sec:underwater_env}

  \begin{figure}[H]
    \centering
    \begin{minipage}{0.48\linewidth}
      \centering
      \includegraphics[width=\linewidth]{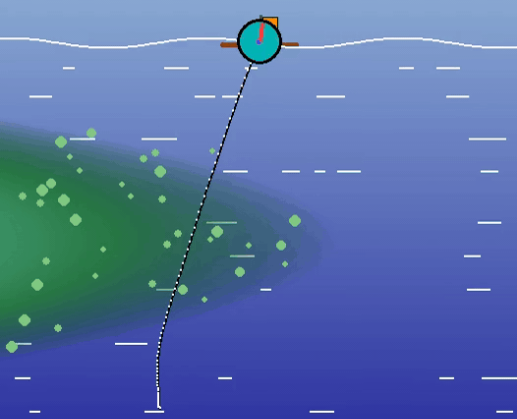}
    \end{minipage}\hfill
    \begin{minipage}{0.48\linewidth}
      \centering
      \includegraphics[width=\linewidth]{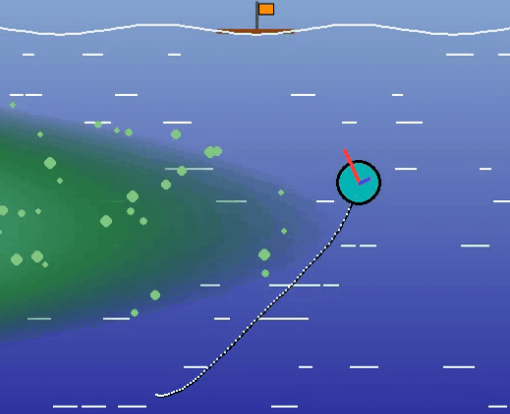}
    \end{minipage}
    \caption{Contaminated-Zone AUV Navigation.
      Left: a representative trajectory generated by the fallback policy.
      It reaches $\G$ along a direct path through the contaminated
      region, leaving scope for the base policy to achieve a higher return.
      Right: task layout and desired behavior;
      the green region is the contaminated low-reward region, and the
      flag~\raisebox{-0.25ex}{\includegraphics[height=1.35em]{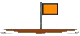}}
      marks the goal set $\G$ defined in~\eqref{eq:auv_goal_set}.
      The trajectory in the right panel bypasses the
      contaminated region on the way to $\G$.}
    \label{fig:envunderwaterdrone}
  \end{figure}

  \paragraph{Dynamics and interface}
  The task in \Cref{fig:envunderwaterdrone} represents an AUV moving
  in a vertical plane.
  Its continuous-time dynamics are
  \begin{equation}
    \begin{aligned}
      \dot{x} &=v_x, \qquad \dot{y}=v_y,\\
      \dot{\vartheta}&=\omega, \qquad
      \dot{\omega}=\frac{d_{\mathrm{offset}}F_{\mathrm{lat}}}{I},\\
      \dot{v}_x
      &=\frac{F_{\mathrm{long}}\cos\vartheta
      -F_{\mathrm{lat}}\sin\vartheta-C_d\lVert v\rVert v_x}{m},\\
      \dot{v}_y
      &=\frac{F_{\mathrm{long}}\sin\vartheta
      +F_{\mathrm{lat}}\cos\vartheta-C_d\lVert v\rVert v_y-mg}{m}.
    \end{aligned}
    \label{eq:auv_dynamics}
  \end{equation}
  The physical state
  $\state=(x,y,\vartheta,v_x,v_y,\omega)\in\mathbb{R}^{6}$ contains
  position, orientation, translational velocity, and angular velocity.
  The policy receives the angle-embedded observation
  $[x,y,\cos\vartheta,\sin\vartheta,v_x,v_y,\omega]$ and selects the
  body-frame thrusts
  $\action=(F_{\mathrm{long}},F_{\mathrm{lat}})\in[-1,1]\times[-0.5,0.5]$.
  The parameters are $m=1$, $I=0.1$, $g=0.5$, $C_d=0.05$, and
  $d_{\mathrm{offset}}=0.2$.
  Equation~\eqref{eq:auv_dynamics} is integrated by explicit Euler
  steps of $0.02$~s, and each episode contains at most 1500 steps.

  \paragraph{Objective, reward, and initialization}
  The goal is the surface opening
  \begin{equation}
    \label{eq:auv_goal_set}
    \G=\{\state\in\states: |x|\leq0.4\ \text{and}\ y\geq4\}.
  \end{equation}
  The green region in \Cref{fig:envunderwaterdrone} is the parabolic
  contaminated zone
  \begin{equation*}
    \mathcal{C}
    =\left\{(x,y)\in\mathbb{R}^{2}:
    \frac{x}{0.81}+\frac{(y-2)^{2}}{0.36}\leq1\right\}.
  \end{equation*}
  The per-step reward penalizes displacement from the surface
  opening, excessive translational and angular velocities, and entry
  into this zone:
  \begin{equation*}
    \reward(\state,\action)
    =-\frac{(y-4)^{2}+x^{2}}{4}
    -\frac{v_x^{2}+v_y^{2}}{20}-\frac{\omega^{2}}{100}
    -5\mathbf{1}_{\{(x,y)\in\mathcal{C}\}}.
  \end{equation*}
  Initial states are sampled as
  \begin{equation*}
    \begin{aligned}
      \state_0\sim\operatorname{Uniform}\big(&[-1.25,1.25]\times[0,4/3]
        \times[9\pi/20,11\pi/20]\\
      &\times[-0.2,0.2]^{3}\big),
    \end{aligned}
  \end{equation*}
  so that the vehicle starts in the lower part of the water column
  with small velocities and an approximately upward orientation.

  \paragraph{Fallback policy}
  The fallback is a closed-form PD controller that first computes the
  desired inertial-frame forces
  \begin{equation*}
    f_x=-1.5x-0.8v_x,
    \qquad
    f_y=mg+0.6(4.1-y)-1.2v_y,
  \end{equation*}
  and then transforms them to body-frame thrusts:
  \begin{equation*}
    \begin{bmatrix}F_{\mathrm{long}}\\F_{\mathrm{lat}}
    \end{bmatrix}
    =
    \begin{bmatrix}
      \cos\vartheta&\sin\vartheta\\
      -\sin\vartheta&\cos\vartheta
    \end{bmatrix}
    \begin{bmatrix}f_x\\f_y
    \end{bmatrix},
  \end{equation*}
  with each component clipped to the action bounds.
  It reliably directs the AUV to the surface opening but does not use
  the contaminated-zone geometry, which explains the suboptimal
  sample trajectory in the left panel of \Cref{fig:envunderwaterdrone}.

  \subsection{Treasure-Collecting Robot}\label{sec:robot_env}

  \begin{figure}[H]
    \centering
    \begin{minipage}{0.48\linewidth}
      \centering
      \includegraphics[width=\linewidth]{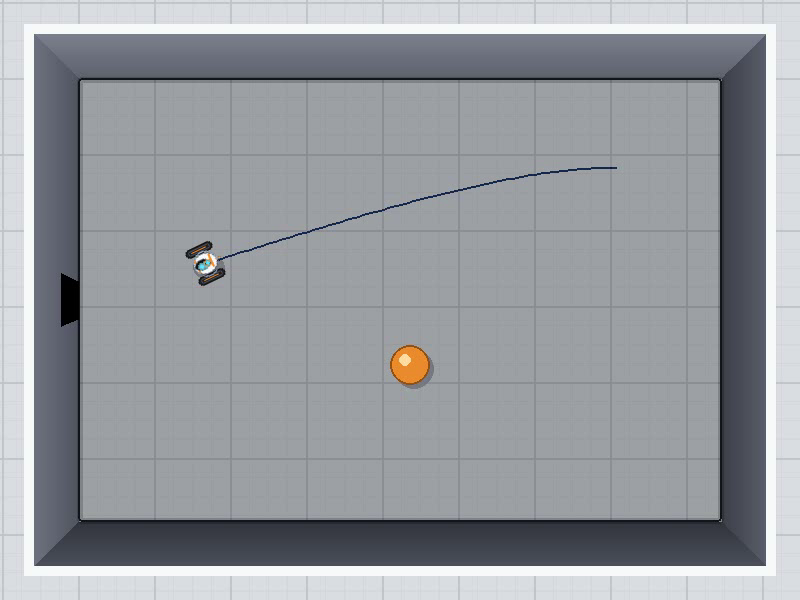}
    \end{minipage}\hfill
    \begin{minipage}{0.48\linewidth}
      \centering
      \includegraphics[width=\linewidth]{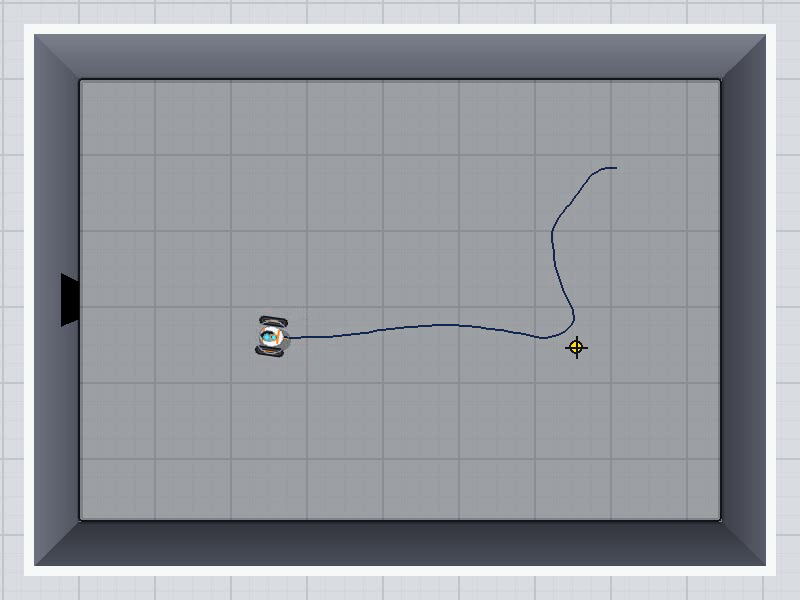}
    \end{minipage}
    \caption{Treasure-Collecting Robot.
      Left: representative behavior of the fallback policy.
      The black marker on the left boundary indicates the goal, and
      the orange disk denotes the collectible treasure.
      The fallback policy steers directly toward the goal and does
      not intercept the treasure.
      Right: desired behavior induced by the learned policy.
      The learned policy first deviates toward the collectible treasure
      and then returns toward the goal, instead of following a direct
    goal-seeking trajectory.}
    \label{fig:treasure_robot}
  \end{figure}

  \paragraph{Environment description}
  The robot moves with a constant linear speed.
  Its dynamics are given by
  \begin{equation*}
    \dot{x} = v \cos \vartheta, \quad \dot{y} = v \sin \vartheta, \quad
    \dot{\vartheta} = \omega,
  \end{equation*}
  where $(x, y) \in [0,1]^2$ are the robot's coordinates, $\vartheta
  \in [-\pi, \pi)$ is the robot's heading, $v\equiv 0.15$~m/s denotes
  the robot's constant linear speed, and $\omega \in [-\pi, \pi)$ is
  the angular velocity serving as the control input.

  During the episode, the robot must collect a treasure whose
  position $(x^{\mathrm{tr}}, y^{\mathrm{tr}}) \in [0,1]^2$ is known
  to the robot.
  The treasure is considered collected when the following condition
  holds:
  \begin{equation*}
    \left(x-x^{\mathrm{tr}}\right)^2
    +\left(y-y^{\mathrm{tr}}\right)^2 \le 0.08^2.
  \end{equation*}
  The treasure moves stochastically in the workspace.
  Let $(x_t^{\mathrm{tr}}, y_t^{\mathrm{tr}})$ and $(v_t^x, v_t^y)$
  denote its position and velocity at time $t$, and let $\Delta t=0.05$~s.
  Its initial velocity $(v_0^x,v_0^y)$ has a random heading and fixed
  speed $v_{\max}=0.12$.
  At each time step, the velocity is perturbed by random variables
  $\epsilon_t^x,\epsilon_t^y\sim\mathcal{N}(0,0.03^2)$, which are
  independent across components and time, and the resulting velocity
  is clipped to magnitude $v_{\max}$.
  The velocity and position updates are given by
  \begin{equation*}
    \begin{aligned}
      v^x_{t+1}
      &=\frac{v^x_t+\epsilon^x_t}
      {\max\!\left(1,
          \frac{\sqrt{(v^x_t+\epsilon^x_t)^2+(v^y_t+\epsilon^y_t)^2}}
      {v_{\max}}\right)},\\
      v^y_{t+1}
      &=\frac{v^y_t+\epsilon^y_t}
      {\max\!\left(1,
          \frac{\sqrt{(v^x_t+\epsilon^x_t)^2+(v^y_t+\epsilon^y_t)^2}}
      {v_{\max}}\right)},
    \end{aligned}
  \end{equation*}
  \begin{equation*}
    x^{\mathrm{tr}}_{t+1}=x^{\mathrm{tr}}_t+\Delta t\,v^x_t,
    \qquad
    y^{\mathrm{tr}}_{t+1}=y^{\mathrm{tr}}_t+\Delta t\,v^y_t.
  \end{equation*}
  The update is followed by reflection at the workspace boundaries
  for $x^{\mathrm{tr}}_{t+1}$ and $y^{\mathrm{tr}}_{t+1}$,
  implemented as component-wise velocity sign flips.

  After collecting the treasure, the robot must reach a goal region
  that is specified explicitly and is also favored by the
  distance-to-goal term in the reward.
  The goal set is defined as
  \begin{equation*}
    \G=\left\{(x,y)\in[0,1]^2\ \middle|\
    (x-x^{\mathrm{g}})^2+(y-y^{\mathrm{g}})^2<0.05^2\right\},
  \end{equation*}
  where $(x^{\mathrm{g}},y^{\mathrm{g}})=(0,0.5)$ is the goal position.

  The state vector is nine-dimensional and is defined as
  \begin{equation*}
    \state=(x,y,\cos\vartheta,\sin\vartheta,
    x^{\mathrm{tr}},y^{\mathrm{tr}},v^x,v^y,I^{\mathrm{tr}}),
  \end{equation*}
  where $I^{\mathrm{tr}}\in\{0,1\}$ is a binary indicator of treasure
  availability:
  \begin{equation*}
    I^{\mathrm{tr}}=
    \begin{cases}
      1, & \text{if the treasure is available for collection},\\
      0, & \text{if the treasure has already been collected}.
    \end{cases}
  \end{equation*}
    Although the treasure velocity $(v^x,v^y)$ is part of the state, the
  implementation excludes it from the inputs of all policies and critic
  estimates used in this environment.
  The dynamics are integrated using an explicit Euler scheme at 20~Hz.
  Each episode lasts at most 1000 steps (50~s) and terminates early
  if the goal set $\G$ is reached.
  Representative rollouts for this task are shown in \Cref{fig:treasure_robot}.

  \paragraph{Reward function and objective}
  The reward function is defined as
  \begin{multline*}
    \reward(\state,\action)
    =-\left\lVert(x^{\mathrm{g}},y^{\mathrm{g}})-(x,y)\right\rVert_2\\
    +50I^{\mathrm{tr}}\mathbf{1}_{\left\{
        \left\lVert(x,y)-(x^{\mathrm{tr}},y^{\mathrm{tr}})\right\rVert_2
    \le 0.08\right\}},
  \end{multline*}
  where $\lVert\bullet\rVert_2$ denotes the Euclidean norm.
  Thus, the task rewards both collecting the treasure and reaching
  the goal as quickly as possible.

  \paragraph{Initial conditions}
  The components of the initial state are sampled
  independently as follows:
  \begin{equation*}
    \begin{aligned}
      x_0&\sim\operatorname{Uniform}[0.7,0.9], &
      y_0&\sim\operatorname{Uniform}[0.1,0.9],\\
      x^{\mathrm{tr}}_0&\sim\operatorname{Uniform}[0.15,0.85], &
      y^{\mathrm{tr}}_0&\sim\operatorname{Uniform}[0.05,0.95],\\
      \vartheta_0&=\pi, &
      I^{\mathrm{tr}}_0&=1,\\
      v^x_0&=v_{\max}\cos\varphi, &
      v^y_0&=v_{\max}\sin\varphi,
    \end{aligned}
  \end{equation*}
  where $\varphi\sim\operatorname{Uniform}[0,2\pi)$.

  \paragraph{Fallback policy design}
  The fallback employs a simple geometric steering law that turns the
  robot toward the provided target point.
  From the observation $[x,y,\cos\vartheta,\sin\vartheta,\ldots]$,
  the controller reconstructs the current heading as
  $\vartheta=\operatorname{atan2}(\sin\vartheta,\cos\vartheta)$ and
  computes the desired heading to the goal as
  $\vartheta^\star=\operatorname{atan2}(y^{\mathrm{g}}-y, x^{\mathrm{g}}-x)$.
  The wrapped angle error is $e=(\vartheta^\star-\vartheta+\pi)\bmod
  2\pi-\pi$, and the action is the clipped proportional control
  \begin{equation*}
    \omega=\operatorname{clip}(k_{\mathrm{turn}}e,
    -\omega_{\max},\omega_{\max}).
  \end{equation*}
  The parameters are set to $k_{\mathrm{turn}}=1$ and $\omega_{\max}=\pi$.
  When the robot is numerically at the goal, $\vartheta^\star$ is set
  to $\vartheta$ to avoid an ill-defined direction.
  This fallback reliably reaches the goal but optimizes only the
  goal-reaching objective; it is not designed to capture the
  high-reward treasure, which results in suboptimal behavior with
  respect to the full reward function.
  This behavior is illustrated in the left panel of \Cref{fig:treasure_robot}.
}

\subsection{Low-, Mid-, and High-Return Checkpoints}
\label{sec:return_checkpoint_selection}

The subsequent experiments examine the behavior of CALF-Wrapper when
it is supplied with actor--critic checkpoints of different observed quality.
This comparison is intended to show how CALF-Wrapper behaves with
relatively poorly trained, intermediately trained, and well-trained
base policies.
Checkpoint quality is assessed using the observed return of the base
policy executed without CALF-Wrapper.
The terms Low-return, Mid-return, and High-return therefore refer to
the observed performance of the base policy and not directly to the
number of training steps.

Each training run produces a sequence of saved actor--critic checkpoints.
For each training run, the range of base-policy returns observed over
these checkpoints is divided into three equal-width intervals.
Low-, Mid-, and High-return refer to the lower, middle, and upper
intervals, respectively.
The checkpoints with base-policy returns closest to the centers of
their respective intervals are selected for evaluation.
The selected checkpoints are used as representative poorly trained,
intermediately trained, and well-trained actor--critic checkpoints in
terms of their observed base-policy returns.
Each selected checkpoint supplies both the base policy and its
corresponding critic estimate to CALF-Wrapper.

One evaluation campaign was conducted in each of the four
environments summarized in \Cref{sec:environments} and described in
Appendix~\ref{app:environment_descriptions}.
In every campaign, the base policy is the pretrained neural policy
supplied to CALF-Wrapper, whereas the fallback is the corresponding
environment-specific controller described in
Appendix~\ref{app:environment_descriptions}.
The base policies for Pendulum and CartPole Swing-Up were trained
using PPO, whereas the base policies for Contaminated-Zone AUV
Navigation and the Treasure-Collecting Robot were trained using TD3.

The critic output is clipped at the environment-specific upper value
specified in \Cref{tab:critic_upper_clipping_values}.
The evaluations set
$\states(\fallbackpolicy)=\{\state\in\states:
\goaldist(\state)\leq0.16\}$ for Contaminated-Zone AUV Navigation.
The boundary through which the AUV reaches $\G$ is the one-dimensional
surface opening $\{(x,y):|x|\leq0.4,\ y=4\}$ in the two-dimensional
position space.
The intersection of the $0.16$-neighborhood with the underwater
region $\{(x,y):y<4\}$ forms a strip of nonzero width in which the
fallback policy is selected before the AUV reaches $\G$.
Thus, $d^{*}>0$ is a natural choice for this environment.
The goal sets in Pendulum, CartPole Swing-Up, and the
Treasure-Collecting Robot already define regions of nonzero width, so
this geometric issue does not arise.
The evaluations therefore use $d^{*}=0$ and set
$\states(\fallbackpolicy)=\G$ in these three environments.

\subsection{Operating Modes}
\label{sec:operating_modes}

CALF-Wrapper determines at every time step whether to apply the
action proposed by the pretrained base policy or the action supplied
by the fallback policy.
A fixed configuration of this selection rule is called an operating mode.

For the evaluations in this paper, the majorant of the random
acceptance probabilities is chosen as
\begin{equation*}
  \bar{\rho}_t:=p_{\mathrm{relax}}\lambda^t,
  \qquad
  p_{\mathrm{relax}}\in[0,1],
  \quad
  \lambda\in[0,1).
\end{equation*}
For a simple interpretation of the operating modes, consider the
setting in which
$\states(\fallbackpolicy)=\varnothing$ and
$\rho_t(\state)=\bar{\rho}_t$ for every state.
The hyperparameters for the operating modes are selected using this setting.

For a finite episode of $T$ steps, the normalized acceptance budget
(NAB) is defined as the expected fraction of time steps on which
CALF-Wrapper selects the base policy:
\begin{equation}
  \mathrm{NAB}_T
  :=\frac{1}{T}
  \E{
    \sum_{t=0}^{T-1}
  \mathbf{1}\{\basepolicy\text{ is selected at time }t\}}.
  \label{eq:nab_definition}
\end{equation}
For this schedule, the probability of selecting the base policy at
time $t$ satisfies
\begin{equation}
  \PP{\basepolicy\text{ is selected at time }t}
  \geq p_{\mathrm{relax}}\lambda^t.
  \label{eq:base_policy_probability_lower_bound}
\end{equation}
Averaging this inequality over the $T$ episode steps gives the
following lower bound for NAB:
\begin{equation}
  \mathrm{NAB}_T
  \geq \underline{\mathrm{NAB}}_T(p_{\mathrm{relax}},\lambda)
  := \frac{p_{\mathrm{relax}}}{T}
  \sum_{t=0}^{T-1}\lambda^t.
  \label{eq:normalized_nab}
\end{equation}
In the simplified setting above, a target
$\underline{\mathrm{NAB}}_T$ specifies a lower bound on the expected
fraction of episode steps assigned to the base policy.
The evaluation considers six operating modes: Conservative, Guarded,
Moderate, Balanced, High, and Almost Open.
Their target $\underline{\mathrm{NAB}}_T$ values increase progressively.
The target $\underline{\mathrm{NAB}}_T$
provides a common scale for the random acceptance probability
schedules of the six modes.
The mode names describe progressively greater reliance on the base policy:
\begin{itemize}
  \item \textbf{Conservative} places no reliance on the random
    acceptance condition.
    Outside $\states(\fallbackpolicy)$, the
    fallback policy is selected unless the critic condition permits
    the base policy.
  \item \textbf{Guarded} places limited reliance on the base policy
    and is intended to retain predominant use of the fallback policy
    in the absence of selections based on the critic condition.
  \item \textbf{Moderate} places greater reliance on the base policy
    than Guarded while keeping the target
    $\underline{\mathrm{NAB}}_T$ below one half.
  \item \textbf{Balanced} places the target
    $\underline{\mathrm{NAB}}_T$ at the midpoint value of $0.5$.
  \item \textbf{High} places greater reliance on the base policy by
    assigning most of the target acceptance budget to it.
  \item \textbf{Almost Open} places near-complete reliance on the
    base policy through the random acceptance condition outside
    $\states(\fallbackpolicy)$.
\end{itemize}
Their exact parameter settings are listed in \Cref{tab:calfw_operating_modes}.
\begin{table}[H]
  \centering
  \caption{CALF-Wrapper parameters used in the reported evaluations.
    The horizons are $T=200$ for Pendulum, $T=1000$ for CartPole and
  the Treasure-Collecting Robot, and $T=1500$ for the AUV.}
  \label{tab:calfw_operating_modes}
  \scriptsize \setlength{\tabcolsep}{2.2pt}
  \begin{tabular}{@{}lccccc@{}}
    \toprule Mode & $p_{\mathrm{relax}}$ &
    \shortstack{Target\\$\underline{\mathrm{NAB}}_T$} &
    \shortstack{$\lambda$\\$T=200$} &
    \shortstack{$\lambda$\\$T=1000$} &
    \shortstack{$\lambda$\\$T=1500$} \\ \midrule Conservative & 0 &
    0.00 & 0 & 0 & 0 \\ Guarded & 1 & 0.10 & 0.950002 & 0.990000 &
    0.993334 \\ Moderate & 1 & 0.35 & 0.986695 & 0.997343 & 0.998229
    \\ Balanced & 1 & 0.50 & 0.992010 & 0.998406 & 0.998937 \\ High &
    1 & 0.70 & 0.996178 & 0.999238 & 0.999492 \\ Almost Open & 1 &
    0.95 & 0.999480 & 0.999896 & 0.999931 \\ \bottomrule
  \end{tabular}
\end{table}
For the Conservative mode, $\lambda=0$.
For each positive-budget mode, $p_{\mathrm{relax}}=1$, and $\lambda$
is obtained numerically as the unique solution in $[0,1)$ of
\begin{equation*}
  \frac{1-\lambda^T}{T(1-\lambda)}
  =\text{Target }\underline{\mathrm{NAB}}_T.
\end{equation*}
For each operating mode, the value of $\text{Target
}\underline{\mathrm{NAB}}_T$ is taken from the corresponding column
of \Cref{tab:calfw_operating_modes}.

A larger target
$\underline{\mathrm{NAB}}_T$ increases the influence of the base
policy outside $\states(\fallbackpolicy)$ and generally leaves fewer
steps for the fallback policy.
For the Balanced mode, the expected number of base-policy selections
in the simplified setting is at least the expected number of
fallback-policy selections.
The High and Almost Open modes can select the base policy more frequently.
The Moderate mode also gives the base policy substantial influence,
which the critic condition can increase further.
If the base policy stabilizes the system poorly, the fallback policy
may not receive enough steps to reach the goal set within the
episode, and the goal-reaching rate can be below $100\%$ from the
Moderate mode onward.
The Conservative and Guarded modes, and sometimes the Moderate mode,
can give the fallback policy enough influence to produce a higher
goal-reaching rate.
Conversely, larger targets allow a well-performing base policy to
influence more of the episode.
\Cref{sec:experimental_results} reports this trade-off across all
modes and checkpoint levels.

\begin{figure}[H]
  \centering
  \includegraphics[width=\columnwidth]{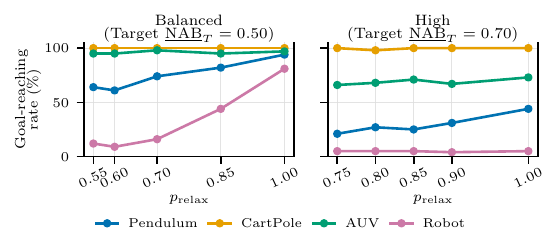}
  \caption{Goal-reaching rates for the Mid-return checkpoints as
    $p_{\mathrm{relax}}$ varies at fixed target
    $\underline{\mathrm{NAB}}_T=0.50$ for the Balanced mode and
    $0.70$ for the High mode.
    Each point is computed from 100 evaluation trials.
    AUV and Robot denote Contaminated-Zone AUV Navigation and the
  Treasure-Collecting Robot, respectively.}
  \label{fig:prelax_fixed_nab_mid_ggr}
\end{figure}

The target $\underline{\mathrm{NAB}}_T$ can equivalently be adjusted
by fixing $\lambda$ and varying $p_{\mathrm{relax}}$.
To assess the choice $p_{\mathrm{relax}}=1$,
\Cref{fig:prelax_fixed_nab_mid_ggr} varies $p_{\mathrm{relax}}$ at
fixed $\underline{\mathrm{NAB}}_T$ for the Mid-return checkpoints in
the Balanced and High modes.
These settings give base policies that are not yet well trained a
substantial influence and make the effect of the fallback policy on
the observed goal-reaching rate visible.
Exact paired McNemar tests with Holm correction identify no
statistically significant reduction in the goal-reaching rate at
$p_{\mathrm{relax}}=1$ and identify statistically significant
increases in several comparisons.
These results support the common choice $p_{\mathrm{relax}}=1$ used
in \Cref{tab:calfw_operating_modes}.

\begin{remark}
    The operating modes concern the majorant $\bar{\rho}_t$ of the random
  acceptance probabilities and not the functions $\rho_t(\state)$
  themselves.
  The dependence of $\rho_t(\state)$ on the state allows the assumptions
  of the theoretical results, such as Assumption~\ref{ass:lp_acceptance}
  of Theorem~\ref{thm:lp_first_hitting}, to be satisfied.
  The feasibility of all assumptions of
  Theorems~\ref{thm:etagoalreaching} and~\ref{thm:lp_first_hitting} is
  discussed separately in
  \Cref{sec:finite_critic_supremum_discussion,sec:uniform_lp_assumptions}.
\end{remark}

\subsection{Choice and Ablation of the Critic Threshold}
\label{sec:nu_calibration}

To isolate the effect of $\nu$, consider the Conservative mode with
$p_{\mathrm{relax}}=0$ and $\lambda=0$.
For simplicity of the following reasoning, suppose that
$\states(\fallbackpolicy)=\varnothing$.
This supposition is used only in the explanation below and does not
change the experimental configuration stated above.

CALF-Wrapper then selects the base policy precisely
when $\Valuebase(\State_t)\geq\Value_t^\dagger+\nu$.
Otherwise, CALF-Wrapper selects the fallback policy.
The threshold $\nu$ therefore determines the smallest increase above
the stored critic value $\Value_t^\dagger$ that is sufficient for
selecting the base policy.
For CALF-Wrapper to generate a trajectory that both reaches the goal
set and attains a high accumulated return, the base policy must be
sufficiently effective for the control task.
The trained base policy nevertheless has no established goal-reaching
property and may fail to reach the goal set from some initial states.
CALF-Wrapper compensates for this possibility by using the fallback
policy when it is needed.

The base policy has the greatest potential to improve the trajectory
near the beginning of an episode, when the state can be far from the
goal set and a substantial part of the trajectory remains.
The value of $\nu$ should therefore be small enough to accept
meaningful increases of the critic in this part of the episode.
After the state has approached the goal set, most of the motion from
the initial state toward the goal set has already been completed, and
the remaining time must still be sufficient to reach the goal set.
In this region, the possible additional improvement from selecting
the base policy is smaller, whereas, in practice, there is no
evidence that selecting it will continue to move the state toward the goal set.
The value of $\nu$ should therefore be large enough that ordinary
local variations of the critic do not repeatedly cause CALF-Wrapper
to select the base policy near the goal set.

The preceding discussion leads to the following idealized reference for $\nu$.
Suppose that the reward design is consistent with the control task,
the base policy approximates an optimal policy that reaches the goal
set, and the critic accurately approximates the corresponding optimal
value function $\Value^{*}$.
Under these assumptions, a natural candidate for a rule of thumb is
to set $\nu$ equal to the maximum variation of $\Value^{*}$ in a
prescribed $d^{*}$-neighborhood of the goal set:
\begin{equation*}
  \begin{gathered}
    \nu
    =
    \max_{\state,\state'\in\G(d^{*})}
    \left|\Value^{*}(\state)-\Value^{*}(\state')\right|,
    \\
    \G(d^{*})
    :=
    \{\state\in\states:\goaldist(\state)\leq d^{*}\},
    \qquad d^{*}\geq0.
  \end{gathered}
\end{equation*}
\par \noindent Computing this variation directly is generally intractable.
However, this direct computation is not required for the sensitivity
study and ablation considered in this subsection.

It is sufficient to select an appropriate environment-specific scale
for the range of values over which $\nu$ is varied.
Remark~\ref{rem:local_value_reward_scale} in
Appendix~\ref{sec:nu_scaling_interpretation} provides a formal
motivation for such a scale.
The remark is not a central result of the paper, and its assumptions
are not required by CALF-Wrapper or by the theoretical results in
Sections~\ref{sec:goal_reaching_property} and~\ref{sec:uniform_lp_bounds}.
For the sensitivity and ablation study,
\eqref{eq:local_value_reward_scale} from
Remark~\ref{rem:local_value_reward_scale} provides the insight that
$\nu$ can be reparameterized as follows:
\begin{equation}
  \nu
  :=
  \eta\left(r_{\G}^{+}-r_{\G}^{-}\right),
  \qquad \eta>0,
  \label{eq:nu_reward_calibration}
\end{equation}
where $r_{\G}^{-}$ and $r_{\G}^{+}$ are the minimum and maximum of
$\reward(\state,\action)$ over $\G(d^{*})\times\actions$, respectively.

\begin{figure}[t]
  \centering
  \includegraphics[width=0.85\columnwidth]{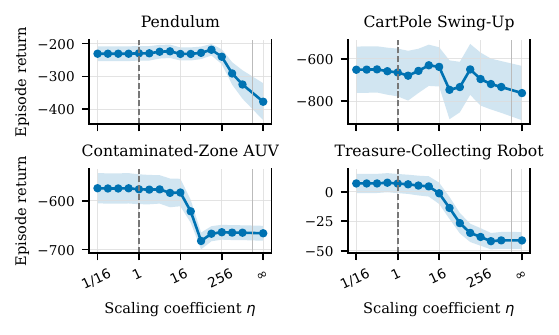}
  \caption{Sensitivity of Conservative CALF-Wrapper to the critic
    threshold $\nu$ at the High-return actor--critic checkpoint for
    each environment.
    The curves show mean episode return, and the shaded regions show
    95\% confidence intervals over 100 trials.
    The goal-reaching rate is 100\% in every evaluation shown.
    The vertical dashed line marks the reward-based reference
    $\eta=1$, for which $\nu=r_{\G}^{+}-r_{\G}^{-}$.
    The point $\nu=+\infty$ is the ablation in which selection of the
    base policy based on the critic is removed, and CALF-Wrapper
  selects the fallback policy at every time step.}
  \label{fig:nu_threshold_sweep}
\end{figure}

$\nu$ is evaluated in all four environments by varying $\eta$ over
\begin{equation}
  \eta\in\left\{2^{-4},2^{-3},\ldots,2^{10},+\infty\right\}.
  \label{eq:nu_ablation_grid}
\end{equation}
Each setting is evaluated over 100 trials from the same initial conditions.
The value $\eta=+\infty$ corresponds to $\nu=+\infty$ and makes
CALF-Wrapper select the fallback policy at every time step.
This setting provides the ablation in which selection of the base
policy based on the critic is removed.
The results for $d^{*}=0$ are shown in \Cref{fig:nu_threshold_sweep}.

\Cref{fig:nu_threshold_sweep} supports $\eta=1$ as a practical rule
of thumb based on the behavior observed across the four environments.
\begin{equation}
  \boxed{\nu=r_{\G}^{+}-r_{\G}^{-}.}
  \label{eq:nu_one_step_calibration}
\end{equation}
For Pendulum, increasing $\eta$ above one initially produces only a
slow upward change in the mean return, and the magnitude of this
change is small before the return begins to decrease at approximately
$\eta=256$.
For the Contaminated-Zone AUV and the Treasure-Collecting Robot,
$\eta=1$ lies well within a broad range of stable, near-maximum mean
returns, while a pronounced deterioration begins around $\eta=16$.
For CartPole Swing-Up, the mean return has an overall downward
tendency as $\eta$ increases.
After $\eta=1$, this tendency is accompanied by irregular upward and
downward changes, including both temporary recoveries and pronounced decreases.
Thus, increasing $\eta$ beyond one provides only a small improvement
for Pendulum, no material improvement before eventual deterioration
for the AUV and the Robot, and unstable changes with an overall
downward tendency for CartPole Swing-Up.
The sensitivity and ablation study yields the rule-of-thumb values of
$\nu$ reported in \Cref{tab:nu_reward_calibration}.
\begin{table}[H]
  \centering
  \caption{Reward-based reference values for the critic threshold.}
  \label{tab:nu_reward_calibration}
  \begin{tabular}{lrrr}
    \toprule Environment & $r_{\G}^{-}$ & $r_{\G}^{+}$ & $\nu$
    \\ \midrule Pendulum & $-0.015502$ & $0$ & $0.015502$ \\ CartPole
    Swing-Up & $-0.050876$ & $0$ & $0.050876$ \\ Contaminated-Zone
    AUV & $-0.040000$ & $0$ & $0.040000$ \\ Treasure-Collecting Robot
    & $-0.050000$ & $0$ & $0.050000$ \\ \bottomrule
  \end{tabular}
\end{table}

\newcommand{\nurewardscaleappendix}{%
  \section{A Bound on the Value-Function Difference}
  \label{sec:nu_scaling_interpretation}

  This appendix states and proves a bound on the difference between
  the values of two states near the goal set.
  The assumptions of the remark are used only to give a formal
  interpretation of the scale used for the sensitivity study.
  They are not assumptions required by CALF-Wrapper.

  \begin{remark}
    \label{rem:local_value_reward_scale}
    Consider deterministic dynamics described by a transition map
    $f:\states\times\actions\to\states$ and a deterministic base policy.
    For every $\state\in\states$ and $\action\in\actions$,
    $f(\state,\action)$ is the next state obtained after applying
    $\action$ at $\state$.
    For every initial state $\state$, define the resulting states by
    \begin{equation*}
      \state_{0}^{\basepolicy}(\state):=\state,
      \qquad
      \state_{k+1}^{\basepolicy}(\state)
      :=f\!\left(
        \state_{k}^{\basepolicy}(\state),
        \basepolicy(\state_{k}^{\basepolicy}(\state))
      \right).
    \end{equation*}
    Thus, $\state_{k}^{\basepolicy}(\state)$ is the state obtained
    after $k$ consecutive applications of the base policy from the
    initial state $\state$.
    Let the goal set consist of one point, $\G=\{\state^{*}\}$.
    Assume that the reward under the base policy has a local maximum at
    $\state^{*}$ and is locally quadratic around this point, and that
    the base policy locally exponentially stabilizes $\state^{*}$.

    Specifically, these assumptions mean that there exist $d^{*}>0$,
    $M\geq1$, $\alpha\in(0,1)$, and $0<c_{1}\leq c_{2}$ such that the
    relations
    \eqref{eq:local_reward_maximum}--\eqref{eq:local_exponential_stability}
    hold for every $\state\in \G(d^{*}):=\{\state\in\states:
    \|\state-\state^{*}\|\leq d^{*}\}$ and every $k\in\mathbb
    Z_{\geq0}$: \par\vspace{-0.5\baselineskip} {\footnotesize
      \begin{eqnarray}
        r_{\G}^{+} &:=& \reward(\state^{*},\basepolicy(\state^{*})) =
        \max_{\substack{\tilde{\state}\in\G(d^{*}),\,
        \action\in\actions}} \reward(\tilde{\state},\action),
        \label{eq:local_reward_maximum}
        \\ c_{1}\|\state-\state^{*}\|^{2} &\leq&
        r_{\G}^{+}-\reward(\state,\basepolicy(\state)) \leq
        c_{2}\|\state-\state^{*}\|^{2},
        \label{eq:local_quadratic_reward_bounds}
        \\ \left\|\state_k^{\basepolicy}(\state)-\state^{*}\right\|
        &\leq& M\alpha^{k}\left\|\state-\state^{*}\right\|.
        \label{eq:local_exponential_stability}
      \end{eqnarray}
    }

    {\normalsize
      \setlength{\baselineskip}{13pt}%
      Then, for every $\state,\state'\in\G(\tfrac{d^{*}}{M})$, \par
    }
    {\small
      \begin{eqnarray}
        \left| \Value^{\basepolicy}(\state) -
        \Value^{\basepolicy}(\state') \right| &\leq&
        \frac{c_{2}M^{2}}{c_{1}(1-\gamma\alpha^{2})}
        \left(r_{\G}^{+}-r_{\G}^{-}\right)
        \label{eq:local_value_reward_scale}
      \end{eqnarray}
    }
    \par \noindent where $r_{\G}^{-}:=
    \min_{\state\in\G(d^{*}),\,\action\in\actions}
    \reward(\state,\action)$.
  \end{remark}

  \begin{proof}
    For every $\state\in\G(\tfrac{d^{*}}{M})$,
    \eqref{eq:local_exponential_stability} gives
    \begin{equation*}
      \left\|
      \state_k^{\basepolicy}(\state)-\state^{*}
      \right\|
      \leq
      M\alpha^{k}\frac{d^{*}}{M}
      \leq d^{*}
    \end{equation*}
    for every $k\in\mathbb Z_{\geq0}$.
    Thus, every state on the trajectory generated by the base policy
    from $\state\in\G(\tfrac{d^{*}}{M})$ remains in $\G(d^{*})$.

    For every $\state\in\G(\tfrac{d^{*}}{M})$,
    \eqref{eq:local_exponential_stability} and
    \eqref{eq:local_quadratic_reward_bounds} give {\scriptsize
      \begin{multline*}
        r_{\G}^{+}
        -
        \reward\!\left(
          \state_k^{\basepolicy}(\state),
          \basepolicy(\state_k^{\basepolicy}(\state))
        \right)
        \leq
        c_{2}M^{2}\alpha^{2k}\|\state-\state^{*}\|^{2}
        \\
        \leq
        \frac{c_{2}M^{2}}{c_{1}}\alpha^{2k}
        \left(
          r_{\G}^{+}-\reward(\state,\basepolicy(\state))
        \right)
        \leq
        \frac{c_{2}M^{2}}{c_{1}}\alpha^{2k}
        \left(r_{\G}^{+}-r_{\G}^{-}\right).
      \end{multline*}
    }
    \par \noindent Hence, for every
    $\state,\state'\in\G(\tfrac{d^{*}}{M})$, both rewards lie in the
    same interval with upper endpoint $r_{\G}^{+}$ and width
    $\frac{c_{2}M^{2}}{c_{1}}\alpha^{2k} (r_{\G}^{+}-r_{\G}^{-})$.
    Therefore,
    \begin{equation*}
      \begin{split}
        &\left|
        \reward\!\left(
          \state_k^{\basepolicy}(\state),
          \basepolicy(\state_k^{\basepolicy}(\state))
        \right)
        -
        \reward\!\left(
          \state_k^{\basepolicy}(\state'),
          \basepolicy(\state_k^{\basepolicy}(\state'))
        \right)
        \right|
        \\
        &\qquad\leq
        \frac{c_{2}M^{2}}{c_{1}}
        \alpha^{2k}
        \left(r_{\G}^{+}-r_{\G}^{-}\right).
      \end{split}
    \end{equation*}
    Multiplication by $\gamma^{k}$ and summation over $k\in\mathbb
    Z_{\geq0}$ give {\small
      \begin{equation*}
        \left|
        \Value^{\basepolicy}(\state)-\Value^{\basepolicy}(\state')
        \right|
        \leq
        \frac{c_{2}M^{2}}{c_{1}(1-\gamma\alpha^{2})}
        \left(r_{\G}^{+}-r_{\G}^{-}\right).
      \end{equation*}
    }
  \end{proof}
}

\begin{table*}[!b]
  \centering
  \caption{Episode returns and goal-reaching rates at the Low-, Mid-, and High-return checkpoints in each environment. The PPO-Lagrangian and TD3-Lagrangian rows report checkpoints from the corresponding stages of their training runs. Episode return is shown as mean $\pm$ 95\% confidence-interval half-width over 100 evaluation trials. Goal-reaching rate is shown as a percentage. The Fallback actions columns report the mean percentage of episode time steps at which the fallback policy supplied the applied action. At each checkpoint, the episode-return and goal-reaching-rate cells of the first-, second-, and third-ranked methods are shown in bold black, bold \textcolor{secondbestblue}{dark blue}, and bold \textcolor{thirdbestorange}{dark orange}, respectively. Methods are ordered first by goal-reaching rate and then by episode return; ties receive the same marking.}
  \label{tab:nab_reward_stage_results}
  \scriptsize
  \renewcommand{\arraystretch}{0.90}
  \setlength{\tabcolsep}{3pt}
  \resizebox{0.88\textwidth}{!}{%
    \begin{tabular}{lccccccccc}
      \toprule
      Method & \multicolumn{3}{c}{Low-return checkpoint} & \multicolumn{3}{c}{Mid-return checkpoint} & \multicolumn{3}{c}{High-return checkpoint} \\
      \cmidrule(lr){2-4}\cmidrule(lr){5-7}\cmidrule(lr){8-10}
      & \begin{tabular}[c]{@{}c@{}}Episode\\return\end{tabular} & \begin{tabular}[c]{@{}c@{}}Goal-\\reaching\\rate (\%)\end{tabular} & \begin{tabular}[c]{@{}c@{}}Fallback\\actions (\%)\end{tabular} & \begin{tabular}[c]{@{}c@{}}Episode\\return\end{tabular} & \begin{tabular}[c]{@{}c@{}}Goal-\\reaching\\rate (\%)\end{tabular} & \begin{tabular}[c]{@{}c@{}}Fallback\\actions (\%)\end{tabular} & \begin{tabular}[c]{@{}c@{}}Episode\\return\end{tabular} & \begin{tabular}[c]{@{}c@{}}Goal-\\reaching\\rate (\%)\end{tabular} & \begin{tabular}[c]{@{}c@{}}Fallback\\actions (\%)\end{tabular} \\
      \midrule
      \multicolumn{10}{l}{\textit{Pendulum-v1} (PPO)} \\
      \specialrule{0.5pt}{1.5pt}{1pt}
      Fallback & \textcolor{secondbestblue}{$\bm{-373.28\pm52.71}$} & \textcolor{secondbestblue}{$\bm{100.0}$} & $100.0$ & \textcolor{thirdbestorange}{$\bm{-373.28\pm52.71}$} & \textcolor{thirdbestorange}{$\bm{100.0}$} & $100.0$ & $-373.28\pm52.71$ & $100.0$ & $100.0$ \\
      \cmidrule(lr){1-10}
      Conservative & $\bm{-372.17\pm52.80}$ & $\bm{100.0}$ & $99.5$ & $\bm{-254.28\pm28.94}$ & $\bm{100.0}$ & $92.0$ & \textcolor{secondbestblue}{$\bm{-214.91\pm22.60}$} & \textcolor{secondbestblue}{$\bm{100.0}$} & $88.3$ \\
      Guarded & \textcolor{thirdbestorange}{$\bm{-375.69\pm48.25}$} & \textcolor{thirdbestorange}{$\bm{100.0}$} & $90.4$ & \textcolor{secondbestblue}{$\bm{-257.72\pm22.54}$} & \textcolor{secondbestblue}{$\bm{100.0}$} & $85.1$ & $\bm{-203.10\pm24.56}$ & $\bm{100.0}$ & $84.7$ \\
      \cmidrule(lr){1-10}
      Moderate & $-526.50\pm32.97$ & $99.0$ & $69.9$ & $-387.25\pm25.89$ & $99.0$ & $68.8$ & \textcolor{thirdbestorange}{$\bm{-215.49\pm28.04}$} & \textcolor{thirdbestorange}{$\bm{100.0}$} & $76.8$ \\
      Balanced & $-674.58\pm38.25$ & $72.0$ & $55.5$ & $-481.76\pm28.96$ & $94.0$ & $56.1$ & $-219.53\pm30.41$ & $100.0$ & $73.5$ \\
      High & $-889.25\pm34.06$ & $12.0$ & $32.3$ & $-592.44\pm28.24$ & $44.0$ & $34.5$ & $-274.33\pm47.65$ & $92.0$ & $66.7$ \\
      Almost Open & $-1050.67\pm31.78$ & $2.0$ & $6.7$ & $-674.34\pm25.42$ & $3.0$ & $7.3$ & $-305.74\pm58.93$ & $84.0$ & $59.6$ \\
      \cmidrule(lr){1-10}
      Base policy & $-1096.20\pm12.02$ & $0.0$ & $0.0$ & $-703.05\pm12.81$ & $0.0$ & $0.0$ & $-341.05\pm68.16$ & $77.0$ & $0.0$ \\
      PPO-Lagrangian & $-1147.65\pm28.27$ & $0.0$ & -- & $-360.21\pm57.33$ & $70.0$ & -- & $-273.77\pm39.58$ & $100.0$ & -- \\
      \addlinespace[4pt]
      \multicolumn{10}{l}{\textit{CartPole Swing-Up} (PPO)} \\
      \specialrule{0.5pt}{1.5pt}{1pt}
      Fallback & \textcolor{thirdbestorange}{$\bm{-762.58\pm128.49}$} & \textcolor{thirdbestorange}{$\bm{100.0}$} & $100.0$ & $-762.58\pm128.49$ & $100.0$ & $100.0$ & $-762.58\pm128.49$ & $100.0$ & $100.0$ \\
      \cmidrule(lr){1-10}
      Conservative & \textcolor{secondbestblue}{$\bm{-752.70\pm118.38}$} & \textcolor{secondbestblue}{$\bm{100.0}$} & $90.8$ & $-759.51\pm131.32$ & $100.0$ & $87.4$ & $-665.18\pm113.04$ & $100.0$ & $89.5$ \\
      Guarded & $\bm{-614.14\pm93.33}$ & $\bm{100.0}$ & $83.0$ & $\bm{-464.05\pm73.14}$ & $\bm{100.0}$ & $82.7$ & $-481.28\pm70.69$ & $100.0$ & $84.8$ \\
      \cmidrule(lr){1-10}
      Moderate & $-845.76\pm131.69$ & $100.0$ & $71.2$ & $-471.22\pm86.32$ & $99.0$ & $75.8$ & $\bm{-410.80\pm42.70}$ & $\bm{100.0}$ & $78.7$ \\
      Balanced & $-784.91\pm130.06$ & $96.0$ & $64.9$ & \textcolor{secondbestblue}{$\bm{-503.33\pm72.93}$} & \textcolor{secondbestblue}{$\bm{100.0}$} & $72.7$ & \textcolor{thirdbestorange}{$\bm{-460.12\pm65.71}$} & \textcolor{thirdbestorange}{$\bm{100.0}$} & $75.8$ \\
      High & $-719.52\pm100.53$ & $77.0$ & $56.2$ & \textcolor{thirdbestorange}{$\bm{-594.43\pm113.03}$} & \textcolor{thirdbestorange}{$\bm{100.0}$} & $67.4$ & \textcolor{secondbestblue}{$\bm{-440.06\pm52.51}$} & \textcolor{secondbestblue}{$\bm{100.0}$} & $73.1$ \\
      Almost Open & $-853.83\pm54.30$ & $75.0$ & $26.7$ & $-661.57\pm127.63$ & $98.0$ & $60.2$ & $-518.43\pm92.81$ & $99.0$ & $67.3$ \\
      \cmidrule(lr){1-10}
      Base policy & $-1156.17\pm56.37$ & $8.0$ & $0.0$ & $-883.55\pm251.92$ & $90.0$ & $0.0$ & $-625.10\pm144.78$ & $92.0$ & $0.0$ \\
      PPO-Lagrangian & $-1118.55\pm271.26$ & $97.0$ & -- & $-873.94\pm171.32$ & $96.0$ & -- & $-557.28\pm63.60$ & $100.0$ & -- \\
      \addlinespace[4pt]
      \multicolumn{10}{l}{\textit{Contaminated-Zone AUV Navigation} (TD3)} \\
      \specialrule{0.5pt}{1.5pt}{1pt}
      Fallback & $\bm{-660.17\pm15.57}$ & $\bm{100.0}$ & $100.0$ & $\bm{-660.17\pm15.57}$ & $\bm{100.0}$ & $100.0$ & $-660.17\pm15.57$ & $100.0$ & $100.0$ \\
      \cmidrule(lr){1-10}
      Conservative & \textcolor{secondbestblue}{$\bm{-1148.02\pm101.25}$} & \textcolor{secondbestblue}{$\bm{100.0}$} & $88.8$ & \textcolor{secondbestblue}{$\bm{-824.54\pm41.11}$} & \textcolor{secondbestblue}{$\bm{100.0}$} & $88.0$ & $-559.83\pm32.95$ & $100.0$ & $87.9$ \\
      Guarded & \textcolor{thirdbestorange}{$\bm{-1888.85\pm92.40}$} & \textcolor{thirdbestorange}{$\bm{100.0}$} & $82.9$ & $-1293.43\pm115.10$ & $100.0$ & $82.0$ & $\bm{-498.23\pm25.71}$ & $\bm{100.0}$ & $83.4$ \\
      \cmidrule(lr){1-10}
      Moderate & $-3478.32\pm206.15$ & $77.0$ & $63.1$ & $-1555.73\pm125.43$ & $98.0$ & $63.3$ & \textcolor{secondbestblue}{$\bm{-518.23\pm26.91}$} & \textcolor{secondbestblue}{$\bm{100.0}$} & $80.7$ \\
      Balanced & $-4041.75\pm251.20$ & $45.0$ & $50.6$ & $-1643.67\pm122.66$ & $97.0$ & $55.3$ & \textcolor{thirdbestorange}{$\bm{-520.85\pm27.06}$} & \textcolor{thirdbestorange}{$\bm{100.0}$} & $80.2$ \\
      High & $-4862.36\pm210.37$ & $18.0$ & $31.8$ & $-2166.47\pm184.10$ & $73.0$ & $38.3$ & $-522.27\pm27.10$ & $100.0$ & $79.7$ \\
      Almost Open & $-5647.09\pm221.84$ & $10.0$ & $9.2$ & $-2932.75\pm129.47$ & $2.0$ & $5.6$ & $-536.52\pm39.42$ & $100.0$ & $79.1$ \\
      \cmidrule(lr){1-10}
      Base policy & $-5823.79\pm158.87$ & $5.0$ & $0.0$ & $-2983.37\pm120.32$ & $1.0$ & $0.0$ & $-539.18\pm40.56$ & $100.0$ & $0.0$ \\
      TD3-Lagrangian & $-6459.84\pm102.57$ & $1.0$ & -- & \textcolor{thirdbestorange}{$\bm{-995.06\pm105.04}$} & \textcolor{thirdbestorange}{$\bm{100.0}$} & -- & $-577.19\pm31.20$ & $100.0$ & -- \\
      \addlinespace[4pt]
      \multicolumn{10}{l}{\textit{Treasure-Collecting Robot} (TD3)} \\
      \specialrule{0.5pt}{1.5pt}{1pt}
      Fallback & \textcolor{secondbestblue}{$\bm{-30.15\pm8.40}$} & \textcolor{secondbestblue}{$\bm{100.0}$} & $100.0$ & $-30.15\pm8.40$ & $100.0$ & $100.0$ & $-30.146\pm8.395$ & $100.0$ & $100.0$ \\
      \cmidrule(lr){1-10}
      Conservative & $\bm{-23.89\pm8.94}$ & $\bm{100.0}$ & $92.1$ & \textcolor{secondbestblue}{$\bm{-8.15\pm9.41}$} & \textcolor{secondbestblue}{$\bm{100.0}$} & $78.1$ & \textcolor{thirdbestorange}{$\bm{23.392\pm6.030}$} & \textcolor{thirdbestorange}{$\bm{100.0}$} & $69.8$ \\
      Guarded & \textcolor{thirdbestorange}{$\bm{-116.27\pm18.72}$} & \textcolor{thirdbestorange}{$\bm{100.0}$} & $61.1$ & $\bm{-7.61\pm13.53}$ & $\bm{100.0}$ & $55.5$ & \textcolor{secondbestblue}{$\bm{33.348\pm3.387}$} & \textcolor{secondbestblue}{$\bm{100.0}$} & $55.7$ \\
      \cmidrule(lr){1-10}
      Moderate & $-236.59\pm38.74$ & $100.0$ & $45.9$ & \textcolor{thirdbestorange}{$\bm{-16.90\pm22.05}$} & \textcolor{thirdbestorange}{$\bm{100.0}$} & $47.4$ & $\bm{34.277\pm2.638}$ & $\bm{100.0}$ & $47.5$ \\
      Balanced & $-305.47\pm53.81$ & $97.0$ & $41.4$ & $-8.19\pm20.32$ & $81.0$ & $44.2$ & $34.286\pm2.636$ & $98.0$ & $43.9$ \\
      High & $-377.47\pm62.79$ & $17.0$ & $29.0$ & $-0.63\pm15.87$ & $5.0$ & $28.8$ & $34.025\pm2.705$ & $6.0$ & $28.3$ \\
      Almost Open & $-426.14\pm68.75$ & $0.0$ & $4.8$ & $2.10\pm15.51$ & $3.0$ & $4.8$ & $34.290\pm2.636$ & $5.0$ & $4.7$ \\
      \cmidrule(lr){1-10}
      Base policy & $-455.38\pm68.54$ & $0.0$ & $0.0$ & $2.08\pm15.50$ & $3.0$ & $0.0$ & $34.290\pm2.636$ & $5.0$ & $0.0$ \\
      TD3-Lagrangian & $-187.11\pm60.23$ & $79.0$ & -- & $14.15\pm8.79$ & $95.0$ & -- & $29.83\pm2.85$ & $99.0$ & -- \\
      \bottomrule
    \end{tabular}%
  }
\end{table*}

\begin{figure*}[!t]
  \centering
  \includegraphics[width=0.88\textwidth]{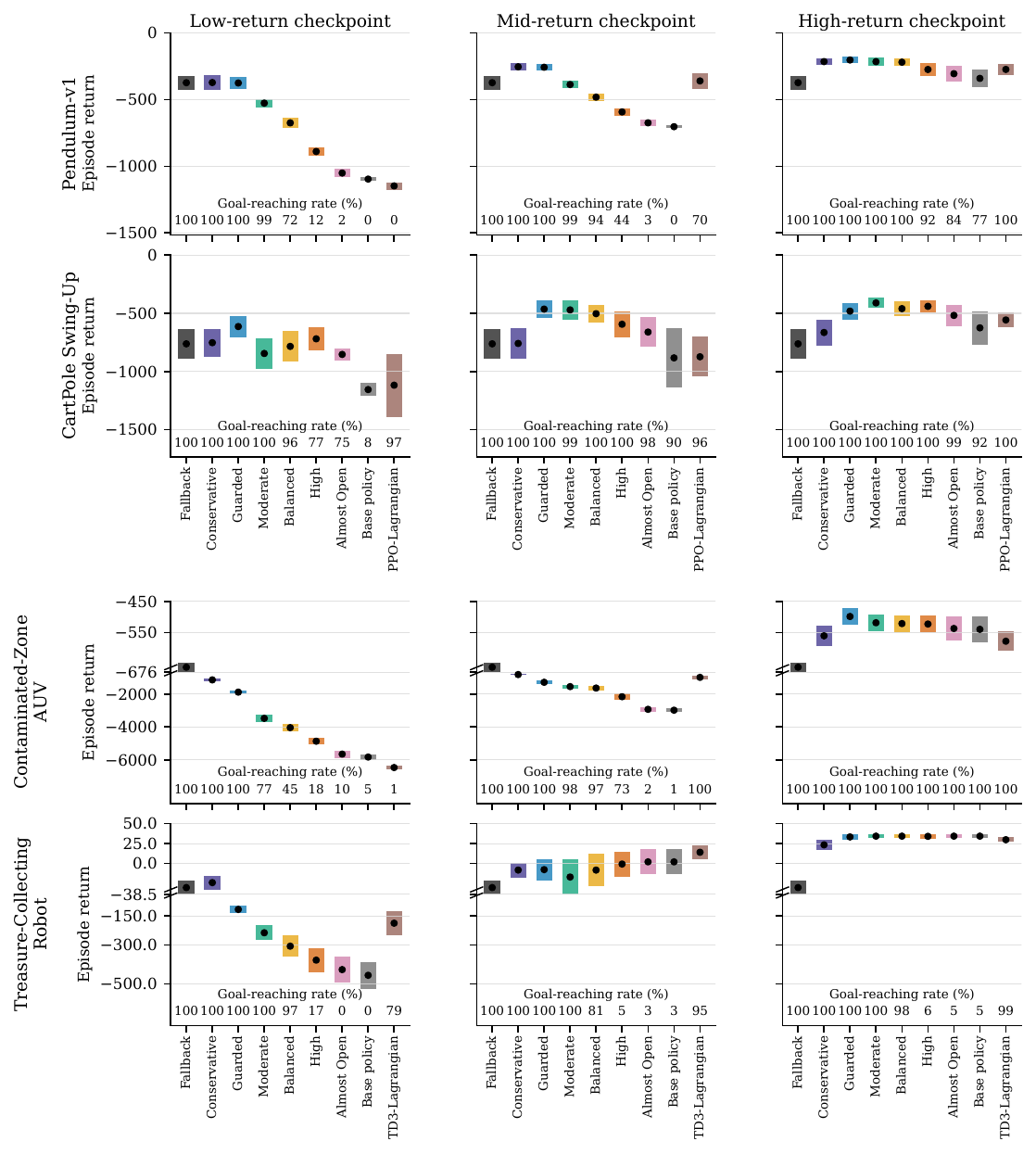}
  \caption{Episode returns and goal-reaching rates for the base policy,
    fallback policy, six CALF-Wrapper operating modes, and the
    corresponding PPO-Lagrangian or TD3-Lagrangian baseline at the
    Low-, Mid-, and High-return checkpoints reported in
    \Cref{tab:nab_reward_stage_results}.
    Each black point shows the mean episode return, and each colored
    rectangle shows its 95\% confidence interval over 100 evaluation trials.
    The numbers inside each panel show the goal-reaching rates in percent.
    The episode-return axes for the AUV and the Treasure-Collecting
    Robot change scale at $-675.74$ and $-38.541$, respectively, to
  show the differences above these values.}
  \label{fig:original_style_checkpoint_results}
\end{figure*}

\subsection{Results}
\label{sec:experimental_results}

The experimental evaluation examines whether CALF-Wrapper attains
both a high episode return and a high observed goal-reaching rate.
The evaluation uses four environments and base policies trained with PPO or TD3.
\mbox{Low-,} \mbox{Mid-,} and High-return actor--critic checkpoints
are selected from different parts of each training run to evaluate
CALF-Wrapper with base policies of different observed quality.

At every checkpoint, the evaluation includes the following comparisons:
\begin{itemize}
  \item The six CALF-Wrapper operating modes are compared with the
    corresponding base policy and fallback policy.
  \item PPO-Lagrangian and TD3-Lagrangian are included as constrained
    reinforcement-learning baselines formulated for the same
    goal-reaching objective.
    These baselines provide a comparison with policies trained to
    balance episode return and goal reaching through a constraint.
    Their formulation and implementation are described in
    Appendix~\ref{sec:lagrangian_baseline_implementation}.
\end{itemize}

The central results are reported in \Cref{tab:nab_reward_stage_results}.
A graphical representation of these results is provided in
\Cref{fig:original_style_checkpoint_results}.
For every environment and checkpoint, the table reports the mean
episode return, the observed goal-reaching rate, and the percentage
of actions supplied by the fallback policy for each CALF-Wrapper operating mode.
The same table reports the corresponding results for the base policy,
the fallback policy, and the PPO-Lagrangian or TD3-Lagrangian baseline.

Each reported entry is computed from 100 evaluation trials.
The goal-reaching rate is the percentage of evaluation trials in which
the state at the end of the episode belongs to $\G$.
The base policy, fallback policy, and CALF-Wrapper modes use the same
initial conditions within each environment and checkpoint.

\begin{table*}[!b]
  \begin{minipage}[t]{0.49\textwidth}
  \centering
  \caption{Paired return comparisons with the fallback policy at the Low-, Mid-, and High-return checkpoints.
    The paired evaluations use the same initial conditions within each environment and checkpoint.
    Each row reports the result of two-sided paired Wilcoxon signed-rank tests over 100 evaluation trials after Holm correction across the 24 environment and operating-mode combinations within the stated checkpoint stage.
    The paired return difference is the CALF-Wrapper return minus the fallback-policy return.
    For rows with a significant test, the table reports whether the observed mean paired return difference is positive or negative.}
  \label{tab:checkpoint_stage_fallback_comparisons}
  \scriptsize
  \setlength{\tabcolsep}{3pt}
  \renewcommand{\arraystretch}{1.05}
  \begin{tabularx}{\linewidth}{l>{\raggedright\arraybackslash}p{0.29\linewidth}c>{\raggedright\arraybackslash}X}
    \toprule
    Return & Statistical result & Count & Environment and operating modes \\
    \midrule
    Low & Significant difference; observed mean paired difference positive & $1/24$ & Pendulum: Conservative. \\
    Low & Significant difference; observed mean paired difference negative & $15/24$ & Pendulum: Moderate, Balanced, High, Almost Open; AUV: Conservative, Guarded, Moderate, Balanced, High, Almost Open; Robot: Guarded, Moderate, Balanced, High, Almost Open. \\
    Low & No significant return difference from the fallback policy & $8/24$ & Pendulum: Guarded; CartPole: Conservative, Guarded, Moderate, Balanced, High, Almost Open; Robot: Conservative. \\
    \midrule
    Mid & Significant difference; observed mean paired difference positive & $11/24$ & Pendulum: Conservative, Guarded; CartPole: Guarded, Moderate, Balanced; Robot: Conservative, Guarded, Moderate, Balanced, High, Almost Open. \\
    Mid & Significant difference; observed mean paired difference negative & $9/24$ & Pendulum: Balanced, High, Almost Open; AUV: Conservative, Guarded, Moderate, Balanced, High, Almost Open. \\
    Mid & No significant return difference from the fallback policy & $4/24$ & Pendulum: Moderate; CartPole: Conservative, High, Almost Open. \\
    \midrule
    High & Significant difference; observed mean paired difference positive & $23/24$ & Pendulum: Conservative, Guarded, Moderate, Balanced, High, Almost Open; CartPole: Guarded, Moderate, Balanced, High, Almost Open; AUV: Conservative, Guarded, Moderate, Balanced, High, Almost Open; Robot: Conservative, Guarded, Moderate, Balanced, High, Almost Open. \\
    High & Significant difference; observed mean paired difference negative & $0/24$ & None. \\
    High & No significant return difference from the fallback policy & $1/24$ & CartPole: Conservative. \\
    \bottomrule
  \end{tabularx}
  \end{minipage}
  \hfill
  \begin{minipage}[t]{0.49\textwidth}
  \centering
  \caption{Paired return comparisons with the base policy at the Low-, Mid-, and High-return checkpoints.
    The paired evaluations use the same initial conditions within each environment and checkpoint.
    Each row reports the result of two-sided paired Wilcoxon signed-rank tests over 100 paired evaluation trials after Holm correction across the 24 environment and operating-mode combinations within the stated checkpoint stage.
    The paired return difference is the CALF-Wrapper return minus the base-policy return.
    For rows with a significant test, the table separately reports whether the observed mean paired return difference is positive or negative.}
  \label{tab:checkpoint_stage_base_comparisons}
  \scriptsize
  \setlength{\tabcolsep}{3pt}
  \renewcommand{\arraystretch}{1.05}
  \begin{tabularx}{\linewidth}{l>{\raggedright\arraybackslash}p{0.29\linewidth}c>{\raggedright\arraybackslash}X}
    \toprule
    Return & Statistical result & Count & Environment and operating modes \\
    \midrule
    Low & Significant difference; observed mean paired difference positive & $22/24$ & Pendulum: Conservative, Guarded, Moderate, Balanced, High, Almost Open; CartPole: Conservative, Guarded, Moderate, Balanced, High, Almost Open; AUV: Conservative, Guarded, Moderate, Balanced, High; Robot: Conservative, Guarded, Moderate, Balanced, High. \\
    Low & Significant difference; observed mean paired difference negative & $0/24$ & None. \\
    Low & No significant return difference from the base policy & $2/24$ & AUV: Almost Open; Robot: Almost Open. \\
    \midrule
    Mid & Significant difference; observed mean paired difference positive & $14/24$ & Pendulum: Conservative, Guarded, Moderate, Balanced, High, Almost Open; CartPole: Guarded, Moderate; AUV: Conservative, Guarded, Moderate, Balanced, High, Almost Open. \\
    Mid & Significant difference; observed mean paired difference negative & $3/24$ & Robot: Conservative, Guarded, Balanced. \\
    Mid & No significant return difference from the base policy & $7/24$ & CartPole: Conservative, Balanced, High, Almost Open; Robot: Moderate, High, Almost Open. \\
    \midrule
    High & Significant difference; observed mean paired difference positive & $11/24$ & Pendulum: Guarded, Moderate, Balanced, High, Almost Open; CartPole: Almost Open; AUV: Guarded, Moderate, Balanced, High, Almost Open. \\
    High & Significant difference; observed mean paired difference negative & $1/24$ & Robot: Conservative. \\
    High & No significant return difference from the base policy & $12/24$ & Pendulum: Conservative; CartPole: Conservative, Guarded, Moderate, Balanced, High; AUV: Conservative; Robot: Guarded, Moderate, Balanced, High, Almost Open. \\
    \bottomrule
  \end{tabularx}
\end{minipage}

\end{table*}

\subsubsection{Episode Return at the High-Return Checkpoints}
\label{sec:results_high_return}

At the High-return checkpoints, CALF-Wrapper produces strong
episode-return results across the six operating modes.
Relative to the fallback policy, 23 of the 24 environment and
operating-mode comparisons have a significant return difference and
a higher observed mean return for CALF-Wrapper.
No significant difference is detected in the remaining comparison.
Relative to the base policy, 11 comparisons have a significant return
difference and a higher observed mean return for CALF-Wrapper, and no
significant difference is detected in 12 comparisons.
Only Conservative on the Treasure-Collecting Robot has a significant
return difference and a lower observed mean return for CALF-Wrapper.
The complete comparisons are reported in
\Cref{tab:checkpoint_stage_fallback_comparisons,tab:checkpoint_stage_base_comparisons}.
The complete statistical design and all adjusted $p$-values are
reported in Appendix~\ref{sec:checkpoint_statistical_design}.

CALF-Wrapper can also produce a return higher than the returns of
both constituent policies.
At the High-return checkpoints, nine operating-mode comparisons are
significantly higher than both the base policy and the fallback
policy according to the intersection--union tests.
The corresponding comparisons with each constituent policy are
reported in
\Cref{tab:checkpoint_stage_fallback_comparisons,tab:checkpoint_stage_base_comparisons}.
All adjusted $p$-values for the intersection--union tests are
reported in \Cref{tab:checkpoint_stage_iut_holm_pvalues}.

\subsubsection{Episode Return at the Low- and Mid-Return Checkpoints}
\label{sec:results_low_mid_return}

The Low-return checkpoints provide a less favorable but expected result.
Relative to the fallback policy, only one of the 24 CALF-Wrapper
comparisons has a significant return difference and a higher observed
mean return for CALF-Wrapper, 15 have a significant return difference
and a lower observed mean return for CALF-Wrapper, and eight have no
detected significant difference.
These comparisons are reported in
\Cref{tab:checkpoint_stage_fallback_comparisons}.
In 22 of the 24 comparisons with the Low-return base policy, a
significant return difference and a higher observed mean return for
CALF-Wrapper are obtained.
This improvement primarily shows that the fallback policy compensates
for the poor behavior of the weak base policy.
CALF-Wrapper cannot be expected to create a strong reward-seeking
policy when the supplied base policy contains little useful behavior.

The Mid-return checkpoints show the transition between these two cases.
On Pendulum, Conservative and Guarded have a significantly higher
return than both the base policy and the fallback policy.
On CartPole, the same result is obtained for Guarded and Moderate.
No operating mode has a significantly higher return than both
policies on the AUV or the Treasure-Collecting Robot at the
Mid-return checkpoints.
The adjusted $p$-values for these joint comparisons are reported in
\Cref{tab:checkpoint_stage_iut_holm_pvalues}.
Useful combined behavior can therefore appear before the base policy
reaches the High-return checkpoint, but the most consistent return
results are obtained with the High-return checkpoints.

Overall, CALF-Wrapper can compensate for a weak base policy, but its
strongest return results are obtained when the base policy is already
well trained.
Across the 48 separate comparisons with the base policy and the
fallback policy at the High-return checkpoints, all but one comparison
has either a significant return difference and a higher observed mean
return for CALF-Wrapper or no detected significant difference.
In several operating modes, the intersection--union tests identify a
CALF-Wrapper return significantly higher than the returns of both policies.

\subsubsection{Goal-Reaching Rate}
\label{sec:results_goal_reaching}

CALF-Wrapper possesses the goal-reaching property established by the
theoretical analysis.
Nevertheless, the observed goal-reaching rate is not 100\% in every
operating mode.
There is no contradiction between these results.
The goal-reaching property concerns the eventual behavior of the
system, whereas each experimental episode terminates after a
prescribed number of steps.
In the more open operating modes, the base policy supplies more
actions and can move the system farther from the goal set.
The influence of the fallback policy may then become sufficient only
when too few steps remain to bring the system to the goal set.
Thus, the existence of the goal-reaching property does not imply that
every choice of $\lambda$ and $p_{\mathrm{relax}}$ produces an
observed goal-reaching rate of 100\%.
The Low- and Mid-return checkpoints serve as a stress test that makes
the effect of the operating-mode parameters visible.

This dependence does not limit the practical applicability of CALF-Wrapper.
The wrapper can be configured so that the fallback policy receives
sufficient influence.
Conservative and Guarded attained a goal-reaching rate of 100\% in
every reported evaluation, including all Low-, Mid-, and High-return
checkpoints.

Conservative provides the most restrictive configuration, whereas
Guarded allows the base policy to make a greater contribution to the
episode return.
At the High-return checkpoints, Guarded was among the three
highest-ranked methods in \Cref{tab:nab_reward_stage_results} on Pendulum, the AUV,
and the Treasure-Collecting Robot.
On CartPole, its observed mean return was higher than the observed
mean returns of both the base policy and the fallback policy.

\emph{Guarded therefore provides the strongest default operating mode
  among those evaluated.
  It attained a goal-reaching rate of 100\% at every checkpoint while
retaining a useful contribution from the base policy to the episode return.}

\subsubsection{Sensitivity to the Operating Mode}
\label{sec:results_mode_sensitivity}

Except for Conservative, the reported operating modes use
$p_{\mathrm{relax}}=1$ and differ in the value of $\lambda$
selected for the corresponding target $\underline{\mathrm{NAB}}_T$.
The evaluation across Guarded, Moderate, Balanced, High, and Almost
Open therefore also evaluates the sensitivity of the episode return
and goal-reaching rate to $\lambda$ and the corresponding acceptance budget.
Conservative supplies the boundary case with zero probabilistic
acceptance budget.
Since the Conservative mode has no random acceptance, the difference
between its results and those of the fallback policy in
\Cref{tab:nab_reward_stage_results} is due only to selection of the
base policy based on the critic.

\subsubsection{Comparison with the Lagrangian Baselines}
\label{sec:results_lagrangian_comparison}

Guarded is selected for comparison with PPO-Lagrangian and
TD3-Lagrangian because it provides the strongest default operating
mode among those evaluated.
Guarded attained an observed goal-reaching rate of 100\% at every
Low-, Mid-, and High-return checkpoint while retaining a useful
contribution from the base policy to the episode return.
The comparison is reported in \Cref{tab:nab_reward_stage_results}.

The Lagrangian baselines produce strong results after sufficient training.
At the High-return checkpoints, their observed goal-reaching rates
are 100\% on Pendulum, CartPole, and the AUV and 99\% on the
Treasure-Collecting Robot.
Their episode returns also show that the constrained formulation can
produce standalone neural policies that combine a high return with a
high observed goal-reaching rate.
Strong goal-reaching behavior is already present at the Mid-return
checkpoints, where the corresponding rates are 70\%, 96\%, 100\%, and 95\%.
These rates exceed the corresponding rates of 0\%, 90\%, 1\%, and 3\%
produced by the ordinary PPO and TD3 base policies.

However, this behavior must first be acquired through training.
At the Low-return checkpoints, the observed goal-reaching rates of
the Lagrangian baselines are 0\%, 97\%, 1\%, and 79\%.
The resulting standalone neural policy has no deployment-time
mechanism that directly supports goal-reaching behavior after training.
Its observed goal-reaching rate therefore depends on the behavior
acquired during constrained training.

Guarded CALF-Wrapper differs in that its goal-reaching behavior is
supported by a deployment-time policy-selection mechanism rather than
being acquired solely through training.
At the Low-return checkpoints, Guarded attains an observed
goal-reaching rate of 100\% and a higher observed mean return than
the corresponding Lagrangian baseline in all four environments.
At the Mid-return checkpoints, Guarded retains a goal-reaching rate
of 100\% in every environment.
Its observed mean return is higher than that of the Lagrangian
baseline on Pendulum and CartPole and lower on the AUV and the
Treasure-Collecting Robot.

At the High-return checkpoints, Guarded has a higher observed mean
return than the corresponding Lagrangian baseline in all four environments.
Both methods attain an observed goal-reaching rate of 100\% on
Pendulum, CartPole, and the AUV, while Guarded attains 100\% and
TD3-Lagrangian attains 99\% on the Treasure-Collecting Robot.
The Lagrangian baselines therefore show that constrained training can
produce standalone neural policies with high returns and high
observed goal-reaching rates.
Guarded CALF-Wrapper addresses a different deployment-time case by
supporting goal-reaching behavior at every evaluated checkpoint,
including checkpoints from the early parts of actor--critic training.

\subsection{Robustness to Critic Noise}
\label{sec:critic_noise_robustness}
\begin{figure}[H]
  \centering
  \includegraphics[width=\columnwidth]{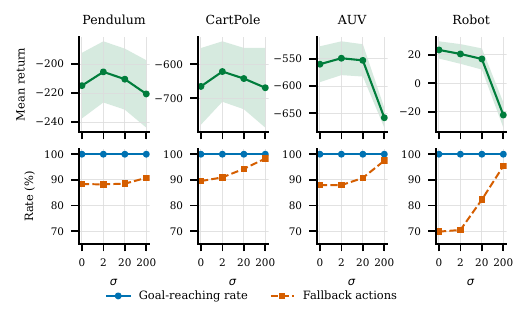}
  \caption{Conservative CALF-Wrapper under additive critic noise
    $\sigma\nu Z$ at the High-return checkpoints.
    The upper row shows the mean episode return, with shaded 95\%
    confidence intervals, and the lower row shows the goal-reaching
    rate and the percentage of time steps at which the fallback
    policy supplied the applied action.
  Every point is computed from 100 paired evaluation trials.}
  \label{fig:critic_noise_robustness}
\end{figure}

The effect of critic error is evaluated by replacing the critic value
$\Valuebase(\state)$ used by Conservative CALF-Wrapper with
$\Valuebase(\state)+\sigma\nu Z$, where $Z\sim\mathcal{N}(0,1)$,
before applying the upper clipping value.
The Conservative mode is used because it selects the base policy only
through the critic condition, so that the effect of critic error is
not combined with that of random acceptance.
The values $\sigma\in\{0,2,20,200\}$ are evaluated at the High-return
checkpoint in each environment.
Each value is evaluated over the same 100 initial conditions, and the
same sequence of standard-normal draws is used for corresponding
trials at every value of $\sigma$.
The condition $\sigma=0$ reproduces the corresponding Conservative
entry in \Cref{tab:nab_reward_stage_results} to the displayed precision.
The results are shown in \Cref{fig:critic_noise_robustness}.

The goal-reaching rate remains $100\%$ across all 16 combinations of
environments and noise levels in \Cref{fig:critic_noise_robustness}.
For $\sigma=2$ and $\sigma=20$, the changes in mean return remain
limited relative to the results at $\sigma=0$.
A pronounced reduction in mean return appears only at $\sigma=200$,
where the AUV mean return decreases from $-559.8$ to $-658.3$ and the
Robot mean return decreases from $23.4$ to $-22.3$.
The Pendulum and CartPole mean returns remain close to their values
at $\sigma=0$ even at $\sigma=200$.
The fallback-action percentage generally increases with $\sigma$ as
CALF-Wrapper relies more frequently on the fallback policy.
These results demonstrate substantial empirical robustness of
CALF-Wrapper to additive noise in the critic estimate: the observed
goal-reaching rate remains $100\%$ at every evaluated noise level,
and a pronounced performance degradation occurs only when the
critic-noise scale is 200 times the critic threshold.

\subsection{CALF-Wrapper Per-Step Latency}
\label{sec:policy_selection_latency}
To exclude overhead introduced by Python and NumPy, the experiment
used a corresponding native Rust implementation of CALF-Wrapper.
The implementation processes one environment and one observation at a
time and does not use a vectorized environment or batched observations.
Rust evaluates the base policy and critic through ONNX Runtime 1.26.0
and its CUDA execution provider on an NVIDIA GeForce RTX 3090 GPU.

A fixed history of 1000 observations was recorded for each environment.
One measurement block processes these observations sequentially and
reports the elapsed time divided by 1000 as the mean per-step latency.
The same observation history was replayed for the base policy and all
six CALF-Wrapper operating modes.
The measurement excludes the environment transition, model export,
training, and construction of the recorded observation history.

The ``Base-policy latency'' column in \Cref{tab:runtime_overhead}
includes one base-policy evaluation.
The ``Critic latency'' column includes one critic evaluation.
For TD3, this evaluation uses the previously computed base-policy
action to report the critic cost separately from the base-policy cost.
The ``Other operations latency'' column includes clipping the critic
value, handling the start of an episode, evaluating the
goal-neighborhood condition, applying the conditions in
Algorithm~\ref{alg_calfwrapper}, updating $\Value_t^{\dagger}$, and
computing the fallback action when the fallback policy is selected.
The ``Fallback policy actions'' column reports the percentage of the
1000 observations for which the fallback policy supplied the applied
action during the runtime measurement.
The ``CALF-Wrapper latency'' column reports the complete CALF-Wrapper
computation.
The ``Added latency'' column reports the CALF-Wrapper latency minus
the base-policy latency within the corresponding paired measurement.

\begin{table}[t]
  \centering
  \caption{Measured mean per-step latency of the native Rust implementation on an NVIDIA GeForce RTX 3090 GPU. The base-policy and critic latencies are means over all 180 measurement blocks for each environment. The mode-dependent values are computed from 30 paired blocks of 1000 sequential observations processed one at a time. Added latency is the CALF-Wrapper latency minus the corresponding base-policy latency; its intervals are paired block-bootstrap 95\% confidence intervals.}
  \label{tab:runtime_overhead}
  \scriptsize
  \setlength{\tabcolsep}{1.3pt}
  \resizebox{\columnwidth}{!}{%
    \begin{tabular}{llrrrrrr}
      \toprule
      Env. & Mode & \shortstack{Fallback\\policy\\actions\\(\%)} & \shortstack{Base-\\policy\\latency\\($\upmu$s)} & \shortstack{Critic\\latency\\($\upmu$s)} & \shortstack{Other\\operations\\latency\\($\upmu$s)} & \shortstack{CALF-\\Wrapper\\latency\\($\upmu$s)} & \shortstack{Added\\latency\\($\upmu$s)\\(95\% CI)} \\
      \midrule
      \multirow{6}{*}{\shortstack{Pendulum\\(PPO)}} & Conserv. & 89.8 & \multirow{6}{*}{73.6} & \multirow{6}{*}{64.9} & 0.013 & 68.9 & $-1.3$ [$-2.2$, $-0.4$] \\
       & Guarded & 84.6 &  &  & 0.014 & 76.3 & $3.4$ [$2.0$, $4.8$] \\
       & Moderate & 76.0 &  &  & 0.016 & 84.6 & $10.0$ [$8.1$, $11.8$] \\
       & Balanced & 71.3 &  &  & 0.014 & 85.4 & $13.5$ [$12.6$, $14.5$] \\
       & High & 65.2 &  &  & 0.015 & 95.2 & $18.6$ [$17.0$, $20.2$] \\
       & Almost Open & 58.8 &  &  & 0.014 & 98.7 & $23.2$ [$21.5$, $24.9$] \\
      \midrule
      \multirow{6}{*}{\shortstack{CartPole\\(PPO)}} & Conserv. & 91.7 & \multirow{6}{*}{73.6} & \multirow{6}{*}{64.1} & 0.027 & 69.6 & $-2.0$ [$-3.0$, $-0.9$] \\
       & Guarded & 85.8 &  &  & 0.027 & 75.5 & $1.6$ [$0.4$, $2.7$] \\
       & Moderate & 80.1 &  &  & 0.027 & 81.8 & $5.7$ [$3.9$, $7.3$] \\
       & Balanced & 79.0 &  &  & 0.025 & 79.9 & $6.8$ [$5.5$, $8.1$] \\
       & High & 78.2 &  &  & 0.024 & 81.7 & $7.9$ [$6.7$, $9.2$] \\
       & Almost Open & 77.2 &  &  & 0.024 & 81.2 & $8.2$ [$7.0$, $9.4$] \\
      \midrule
      \multirow{6}{*}{\shortstack{AUV\\(TD3)}} & Conserv. & 70.0 & \multirow{6}{*}{83.3} & \multirow{6}{*}{173.6} & 0.014 & 260.6 & $178.6$ [$175.8$, $181.4$] \\
       & Guarded & 69.3 &  &  & 0.015 & 265.4 & $181.4$ [$177.8$, $184.8$] \\
       & Moderate & 67.0 &  &  & 0.015 & 265.4 & $180.3$ [$177.6$, $183.0$] \\
       & Balanced & 66.3 &  &  & 0.015 & 264.5 & $181.1$ [$178.2$, $184.3$] \\
       & High & 66.0 &  &  & 0.015 & 263.3 & $180.3$ [$177.7$, $182.8$] \\
       & Almost Open & 65.6 &  &  & 0.014 & 262.8 & $180.4$ [$177.6$, $183.3$] \\
      \midrule
      \multirow{6}{*}{\shortstack{Robot\\(TD3)}} & Conserv. & 95.4 & \multirow{6}{*}{76.0} & \multirow{6}{*}{174.6} & 0.053 & 262.2 & $184.2$ [$181.3$, $187.0$] \\
       & Guarded & 89.0 &  &  & 0.050 & 251.5 & $178.5$ [$176.3$, $180.9$] \\
       & Moderate & 65.8 &  &  & 0.044 & 254.6 & $180.2$ [$176.8$, $183.8$] \\
       & Balanced & 50.0 &  &  & 0.040 & 262.8 & $185.6$ [$183.1$, $188.3$] \\
       & High & 30.4 &  &  & 0.033 & 260.1 & $182.2$ [$178.9$, $185.1$] \\
       & Almost Open & 5.1 &  &  & 0.016 & 260.6 & $185.2$ [$182.3$, $188.0$] \\
      \bottomrule
    \end{tabular}%
  }
\end{table}

The largest measured mean per-step latency in the ``CALF-Wrapper
latency'' column is $265.4\,\upmu\mathrm{s}$, or $0.2654\,\mathrm{ms}$.
The largest value in the ``Added latency'' column is
$185.6\,\upmu\mathrm{s}$, or $0.1856\,\mathrm{ms}$.
Thus, the complete CALF-Wrapper computation remains below
$0.3\,\mathrm{ms}$ per step, and its additional latency relative to
the base policy remains below $0.2\,\mathrm{ms}$ in all reported configurations.

The PPO checkpoints provide a state-value critic directly, as
discussed in Remark~\ref{rem:value_estimate_construction} and
\Cref{sec:computational_complexity}.
The PPO critic can therefore be evaluated without first evaluating
the base policy.
The PPO base policy and critic used in the experiment have comparable
network sizes, and their measured evaluation times are also comparable.
The base policy is evaluated only when
Algorithm~\ref{alg_calfwrapper} selects it, and the fallback policies
in these environments require only simple scalar computations.
Consequently, ``Added latency'' ranges from $-1.3$ to
$23.2\,\upmu\mathrm{s}$ for Pendulum and from $-2.0$ to
$8.2\,\upmu\mathrm{s}$ for CartPole across the six operating modes.

The TD3 checkpoints instead provide two action-value critics.
The implementation first evaluates the base policy and then evaluates
both action-value critics at the resulting action; their smaller
output is used as $\Valuebase$.
This construction accounts for the larger critic cost reported for TD3.
A less expensive construction could evaluate one action-value critic
at a fixed representative action, such as the zero action, as
described in Remark~\ref{rem:value_estimate_construction}.
The reported experiment retains the more expensive construction to
measure its actual overhead.
``Added latency'' nevertheless remains between $178.6$ and
$181.4\,\upmu\mathrm{s}$ for the AUV and between $178.5$ and
$185.6\,\upmu\mathrm{s}$ for the Robot.

Across all six operating modes, ``Other operations latency'' is only
$0.013$--$0.053\,\upmu\mathrm{s}$ per observation.
The scalar operations performed by CALF-Wrapper and the simple
fallback policies therefore contribute negligibly to the measured
latency compared with the neural-network evaluations, which is
consistent with the computational complexity analysis in
\Cref{sec:computational_complexity}.

\FloatBarrier

\section{Conclusion, Limitations, and Future Work}

\subsection{Conclusion}
CALF-Wrapper is designed for control tasks in which an agent must
reach a prescribed goal set while obtaining a high accumulated reward.
CALF-Wrapper combines a base policy intended to obtain high reward
with a predefined fallback policy.

The article provides the following results:
\begin{itemize}
  \item The goal-reaching capabilities of CALF-Wrapper were analyzed
    theoretically and evaluated on two classical-control environments
    and two more challenging environments with higher-dimensional
    state spaces and more complex dynamics.
  \item The experiments provide a comparison with related
    constrained safe-reinforcement-learning methods formulated for
    the same objective.
  \item Sensitivity and ablation studies provide straightforward
    recommendations for selecting the hyperparameters of CALF-Wrapper
    without an additional hyperparameter search.
    The main experiments show how these recommendations combine
    effective task performance with a high observed goal-reaching rate.
  \item The computational complexity analysis and hardware
    measurements show that CALF-Wrapper has a comparatively simple
    deployment-time implementation and low per-step latency in the
    evaluated configurations.
    Recommendations are also provided for reducing this overhead.
\end{itemize}
Taken together, these results provide the theoretical and practical
basis for applying CALF-Wrapper to the stated class of control tasks.

\subsection{Limitations}
CALF-Wrapper addresses whether the system reaches a prescribed goal set.
It does not include a mechanism for detecting or preventing
violations of state or action constraints, which are widely studied
in safe reinforcement learning.
Therefore, the theoretical results and experiments do not establish
constraint satisfaction, collision avoidance, or other safety
properties, and the reported goal-reaching rates should not be
interpreted as constraint-violation rates.

CALF-Wrapper requires a base policy with a corresponding critic
estimate and a predefined fallback policy that satisfies the
goal-reaching property under the stated assumptions.
It therefore belongs to the broader class of methods
that assume access to a separately designed backup or fallback policy
with the required properties, including the methods discussed in
\Cref{sec:related_work}~\cite{seto1998,wagener2021sailr}.
Within this scope, CALF-Wrapper has a direct implementation, requires
limited additional computation, and provides explicit parameters for
controlling the influence of the base policy and the fallback policy.

\subsection{Future Work}
The ideas developed for CALF-Wrapper can be extended in several directions.
\begin{itemize}
  \item Evaluation on physical systems is a natural next step.
    The theoretical results, simulation study, and implementation
    guidance provide a foundation for addressing the additional
    requirements of practical deployment.
  \item CALF-Wrapper can be extended to address the limitation
    discussed above by developing a method that establishes both the
    goal-reaching property and satisfaction of specified state and
    action constraints.
  \item Another direction is to develop a reinforcement-learning
    method that attains high goal-reaching rates throughout training.
    The resulting method should ultimately produce a standalone
    neural policy that does not use the fallback policy during
    deployment, attains a high goal-reaching rate, and maximizes the
    accumulated reward.
\end{itemize}

\appendix
\theoremoneproofappendix \uniformlqproofappendix \uniformasappendix
\klfunctiondecompositionappendix
\environmentdescriptionsappendix
\boundedsuperlevelsetsappendix
\nurewardscaleappendix

\section{PPO-Lagrangian and TD3-Lagrangian Baselines}
  \label{sec:lagrangian_baseline_implementation}

  \setcounter{table}{9}
  \begin{table*}[!b]
  \centering
  \caption{Holm-adjusted $p$-values for the two-sided paired Wilcoxon signed-rank comparisons of the CALF-Wrapper episode return with the fallback-policy return and the base-policy return.
    Each checkpoint-stage and reference-policy column is a separate family of 24 comparisons.
    Bold values are below $0.05$.}
  \label{tab:checkpoint_stage_all_holm_pvalues}
  \scriptsize
  \setlength{\tabcolsep}{4pt}
  \resizebox{0.96\textwidth}{!}{%
    \begin{tabular}{llcccccc}
      \toprule
      & & \multicolumn{3}{c}{Comparison with fallback policy} & \multicolumn{3}{c}{Comparison with base policy} \\
      \cmidrule(lr){3-5}\cmidrule(lr){6-8}
      Environment & Mode & Low & Mid & High & Low & Mid & High \\
      \midrule
      Pendulum & Conservative & $\mathbf{1.48\times10^{-4}}$ & $\mathbf{3.71\times10^{-5}}$ & $\mathbf{7.01\times10^{-6}}$ & $\mathbf{9.35\times10^{-17}}$ & $\mathbf{9.35\times10^{-17}}$ & $0.172$ \\
       & Guarded & $0.989$ & $\mathbf{0.002}$ & $\mathbf{1.97\times10^{-6}}$ & $\mathbf{9.35\times10^{-17}}$ & $\mathbf{9.35\times10^{-17}}$ & $\mathbf{2.12\times10^{-9}}$ \\
       & Moderate & $\mathbf{1.61\times10^{-9}}$ & $0.616$ & $\mathbf{2.69\times10^{-5}}$ & $\mathbf{9.35\times10^{-17}}$ & $\mathbf{9.35\times10^{-17}}$ & $\mathbf{6.33\times10^{-12}}$ \\
       & Balanced & $\mathbf{6.84\times10^{-15}}$ & $\mathbf{0.002}$ & $\mathbf{4.87\times10^{-5}}$ & $\mathbf{9.35\times10^{-17}}$ & $\mathbf{1.14\times10^{-16}}$ & $\mathbf{2.45\times10^{-14}}$ \\
       & High & $\mathbf{1.86\times10^{-16}}$ & $\mathbf{1.57\times10^{-9}}$ & $\mathbf{0.019}$ & $\mathbf{9.35\times10^{-17}}$ & $\mathbf{5.40\times10^{-15}}$ & $\mathbf{1.46\times10^{-15}}$ \\
       & Almost Open & $\mathbf{1.58\times10^{-16}}$ & $\mathbf{5.08\times10^{-13}}$ & $\mathbf{0.04}$ & $\mathbf{7.17\times10^{-14}}$ & $\mathbf{1.07\times10^{-5}}$ & $\mathbf{5.76\times10^{-12}}$ \\
      \midrule
      CartPole & Conservative & $1$ & $0.886$ & $0.168$ & $\mathbf{1.20\times10^{-6}}$ & $1$ & $0.066$ \\
       & Guarded & $0.603$ & $\mathbf{0.006}$ & $\mathbf{0.011}$ & $\mathbf{6.69\times10^{-11}}$ & $\mathbf{1.17\times10^{-5}}$ & $1$ \\
       & Moderate & $0.603$ & $\mathbf{0.007}$ & $\mathbf{5.80\times10^{-4}}$ & $\mathbf{4.52\times10^{-5}}$ & $\mathbf{1.75\times10^{-4}}$ & $0.111$ \\
       & Balanced & $1$ & $\mathbf{0.045}$ & $\mathbf{0.005}$ & $\mathbf{9.61\times10^{-8}}$ & $0.053$ & $1$ \\
       & High & $1$ & $0.387$ & $\mathbf{0.003}$ & $\mathbf{1.04\times10^{-11}}$ & $1$ & $1$ \\
       & Almost Open & $0.088$ & $0.886$ & $\mathbf{0.023}$ & $\mathbf{9.55\times10^{-14}}$ & $0.063$ & $\mathbf{0.009}$ \\
      \midrule
      AUV & Conservative & $\mathbf{8.42\times10^{-15}}$ & $\mathbf{6.91\times10^{-14}}$ & $\mathbf{5.79\times10^{-9}}$ & $\mathbf{9.35\times10^{-17}}$ & $\mathbf{9.35\times10^{-17}}$ & $1$ \\
       & Guarded & $\mathbf{9.35\times10^{-17}}$ & $\mathbf{1.47\times10^{-15}}$ & $\mathbf{2.68\times10^{-16}}$ & $\mathbf{9.35\times10^{-17}}$ & $\mathbf{9.35\times10^{-17}}$ & $\mathbf{7.11\times10^{-16}}$ \\
       & Moderate & $\mathbf{9.35\times10^{-17}}$ & $\mathbf{9.35\times10^{-17}}$ & $\mathbf{1.11\times10^{-15}}$ & $\mathbf{1.07\times10^{-16}}$ & $\mathbf{9.60\times10^{-16}}$ & $\mathbf{3.24\times10^{-15}}$ \\
       & Balanced & $\mathbf{9.35\times10^{-17}}$ & $\mathbf{9.35\times10^{-17}}$ & $\mathbf{2.10\times10^{-15}}$ & $\mathbf{5.71\times10^{-15}}$ & $\mathbf{9.35\times10^{-17}}$ & $\mathbf{5.22\times10^{-14}}$ \\
       & High & $\mathbf{9.35\times10^{-17}}$ & $\mathbf{9.35\times10^{-17}}$ & $\mathbf{2.49\times10^{-15}}$ & $\mathbf{1.41\times10^{-11}}$ & $\mathbf{2.40\times10^{-12}}$ & $\mathbf{1.13\times10^{-8}}$ \\
       & Almost Open & $\mathbf{9.35\times10^{-17}}$ & $\mathbf{9.35\times10^{-17}}$ & $\mathbf{4.89\times10^{-14}}$ & $0.918$ & $\mathbf{4.58\times10^{-8}}$ & $\mathbf{7.31\times10^{-5}}$ \\
      \midrule
      Robot & Conservative & $1$ & $\mathbf{9.78\times10^{-5}}$ & $\mathbf{1.25\times10^{-13}}$ & $\mathbf{4.44\times10^{-13}}$ & $\mathbf{0.003}$ & $\mathbf{9.15\times10^{-13}}$ \\
       & Guarded & $\mathbf{2.94\times10^{-10}}$ & $\mathbf{0.012}$ & $\mathbf{1.62\times10^{-15}}$ & $\mathbf{9.39\times10^{-12}}$ & $\mathbf{0.007}$ & $0.517$ \\
       & Moderate & $\mathbf{2.01\times10^{-11}}$ & $\mathbf{5.23\times10^{-4}}$ & $\mathbf{9.99\times10^{-16}}$ & $\mathbf{5.82\times10^{-8}}$ & $0.063$ & $0.203$ \\
       & Balanced & $\mathbf{4.60\times10^{-12}}$ & $\mathbf{2.07\times10^{-5}}$ & $\mathbf{9.99\times10^{-16}}$ & $\mathbf{6.20\times10^{-5}}$ & $\mathbf{0.007}$ & $0.512$ \\
       & High & $\mathbf{2.36\times10^{-12}}$ & $\mathbf{2.74\times10^{-6}}$ & $\mathbf{9.99\times10^{-16}}$ & $\mathbf{0.047}$ & $0.063$ & $1$ \\
       & Almost Open & $\mathbf{1.04\times10^{-12}}$ & $\mathbf{3.62\times10^{-7}}$ & $\mathbf{9.99\times10^{-16}}$ & $0.155$ & $1$ & $1$ \\
      \bottomrule
    \end{tabular}%
  }
\end{table*}

  \setcounter{table}{10}

  PPO-Lagrangian is evaluated on Pendulum and CartPole Swing-Up, and
  TD3-Lagrangian is evaluated on Contaminated-Zone AUV Navigation and
  the Treasure-Collecting Robot.
  These algorithms were used because PPO and TD3 are also used to
  train the corresponding base policies.
  The comparison therefore retains the policy class used in each
  environment and changes whether the goal-reaching objective is
  addressed by constrained training or by CALF-Wrapper after training.
  Unlike shielding and predictive safety filtering methods, these
  baselines require neither a dynamics model nor an additional
  safe-set specification that is absent from the four experimental tasks.

  Let $T$ denote the episode horizon, and let $t_{\mathrm{end}}$ be the smaller
  of $T$ and the first environment termination time.
  For Pendulum and CartPole Swing-Up, the terminal cost is
  \begin{equation}
    C_{t_{\mathrm{end}}}=\mathbf{1}\{\State_{t_{\mathrm{end}}}\notin\G\}.
    \label{eq:lagrangian_terminal_cost_classical}
  \end{equation}
  For Contaminated-Zone AUV Navigation and the Treasure-Collecting
  Robot, the terminal cost is
  \begin{equation}
    C_{t_{\mathrm{end}}}=\mathbf{1}\{\text{the trajectory has not reached }\G
    \text{ by }t_{\mathrm{end}}\}.
    \label{eq:lagrangian_terminal_cost_navigation}
  \end{equation}
  The two definitions agree with the goal-reaching-rate event used
  for each environment.
  The constrained training problem is
  \begin{equation}
    \begin{aligned}
      \max_\pi\quad
      &J_R(\pi)=\E{\sum_{t=0}^{T-1}\gamma_R^t
      \reward(\State_t,\Action_t)},\\
      \text{subject to}\quad
      &J_C(\pi)=\E{C_{t_{\mathrm{end}}}}\leq\varepsilon,
    \end{aligned}
    \label{eq:lagrangian_baseline_cmdp}
  \end{equation}
  where $\varepsilon=0.05$.
  Since $C_{t_{\mathrm{end}}}$ is binary, the constraint in
  \eqref{eq:lagrangian_baseline_cmdp} requires a goal-reaching
  probability of at least $0.95$ under the environment's
  initial-state distribution and the finite evaluation horizon.
  This finite-horizon expected constraint is not the uniform
  almost-sure goal-reaching condition in
  Assumption~\ref{ass:lp_uniform_fallback_as}.

  Both baselines use the Lagrangian
  \begin{equation}
    \mathcal{L}(\pi,\mu)
    =J_R(\pi)-\mu\bigl(J_C(\pi)-\varepsilon\bigr),
    \qquad \mu\geq0,
    \label{eq:lagrangian_baseline_objective}
  \end{equation}
  and update the multiplier according to
  \begin{equation}
    \mu\leftarrow
    \Pi_{[0,\mu_{\max}]}
    \left(\mu+\alpha_\mu(\widehat J_C-\varepsilon)\right).
    \label{eq:lagrangian_baseline_dual_update}
  \end{equation}

  The physical observation is augmented with $t/T$ so that one
  stationary actor and critic can represent the finite-horizon problem.
  The cost discount is $\gamma_C=1$.
  Thus, the sum of costs in a completed episode equals the binary
  event in \eqref{eq:lagrangian_terminal_cost_classical} or
  \eqref{eq:lagrangian_terminal_cost_navigation}.
  Reward and cost targets stop at both environment termination and
  time-limit truncation, and the terminal observation before
  automatic environment reset is used to evaluate membership in $\G$.

  PPO-Lagrangian adds a cost value function to PPO and computes
  separate reward and cost generalized-advantage estimates.
  The policy update uses their Lagrangian combination and a bounded
  tanh-squashed Gaussian policy.
  The multiplier estimate is the mean binary cost over complete newly
  finished episodes.

  TD3-Lagrangian retains the twin reward critics and delayed actor
  updates of TD3 and adds a bounded cost critic that estimates the
  probability of terminal failure.
  Its actor maximizes the reward critic minus the multiplier times
  the cost critic.
  Its multiplier estimate is computed from the current deterministic
  actor and the cost critic at newly sampled initial states.
  Failures generated by exploratory behavior actions are recorded but
  are not used for this update.

\section{Statistical Design for the Episode-Return Comparisons}
  \label{sec:checkpoint_statistical_design}

  The statistical comparisons summarized in
  \Cref{tab:checkpoint_stage_fallback_comparisons,tab:checkpoint_stage_base_comparisons}
  condition on the selected actor--critic checkpoints.
  For each environment, checkpoint stage, and CALF-Wrapper operating
  mode, the 100 CALF-Wrapper episode returns are paired separately
  with the 100 base-policy returns and the 100 fallback-policy returns.
  Each pair uses the same initial condition.

  A paired difference is computed as the CALF-Wrapper episode return
  minus the episode return of the policy used for comparison.
  Each comparison uses a two-sided paired Wilcoxon signed-rank test.
  Zero paired differences are excluded from the ranking according to
  the Wilcoxon convention.
  The $p$-values are computed using the automatic exact or asymptotic
  method selection provided by the SciPy implementation.
  When an adjusted $p$-value is below $0.05$, the direction reported
  in
  \Cref{tab:checkpoint_stage_fallback_comparisons,tab:checkpoint_stage_base_comparisons}
  is determined by the sign of the observed mean paired difference.

  The multiplicity correction is applied separately for every
  checkpoint stage and every policy used for comparison.
  Each family therefore contains the 24 comparisons formed by the
  four environments and six CALF-Wrapper operating modes.
  The 24 unadjusted $p$-values in each family are adjusted using the
  Holm procedure.
  Statistical significance is assessed at the adjusted level of $0.05$.
  All 144 adjusted $p$-values from these six families are reported in
  \Cref{tab:checkpoint_stage_all_holm_pvalues}.

  The claim that a CALF-Wrapper return is higher than both the
  base-policy return and the fallback-policy return uses an
  intersection--union test.
  Two one-sided paired Wilcoxon signed-rank tests compare
  CALF-Wrapper separately with the base policy and the fallback policy.
  The unadjusted $p$-value for the intersection--union test is the
  larger of these two one-sided $p$-values.
  For each checkpoint stage, the 24 intersection--union $p$-values
  are adjusted together using the Holm procedure.
  All 72 adjusted $p$-values are reported in
  \Cref{tab:checkpoint_stage_iut_holm_pvalues}.

    All tests are conditional on the reported actor--critic checkpoints
  and describe variation over initial conditions, which corresponds to
  the deployment of CALF-Wrapper with a fixed base policy, critic
  estimate, and fallback policy.
  The Low-, Mid-, and High-return checkpoints are taken from different
  stages of the actor--critic training run in each environment and
  supply base policies and critic estimates of substantially different
  quality.
  The training scripts with the configurations of the reported training
  runs, all evaluated checkpoints, including those of
  the Lagrangian baselines, the evaluation code and data, and the code
  that reproduces the reported figures and tables are available in the
  project repository \calfwrepo.
  PPO-Lagrangian and TD3-Lagrangian modify policy training rather than
  deployment and are therefore compared descriptively using the 95\%
  confidence intervals in \Cref{tab:nab_reward_stage_results}.
  Failure to reject the null hypothesis is not interpreted as
  evidence of equivalence or noninferiority.

  \begin{table}[!t]
  \centering
  \caption{Holm-adjusted $p$-values for the intersection--union tests of whether the CALF-Wrapper episode return is higher than both the base-policy return and the fallback-policy return.
    Each checkpoint-stage column is a separate family of 24 comparisons.
    Bold values are below $0.05$.}
  \label{tab:checkpoint_stage_iut_holm_pvalues}
  \scriptsize
  \setlength{\tabcolsep}{4pt}
  \begin{tabular}{llccc}
    \toprule
    Environment & Mode & Low & Mid & High \\
    \midrule
    Pendulum & Conservative & $1$ & $\mathbf{3.42\times10^{-5}}$ & $0.12$ \\
     & Guarded & $1$ & $\mathbf{0.003}$ & $\mathbf{1.64\times10^{-6}}$ \\
     & Moderate & $1$ & $1$ & $\mathbf{2.55\times10^{-5}}$ \\
     & Balanced & $1$ & $1$ & $\mathbf{4.70\times10^{-5}}$ \\
     & High & $1$ & $1$ & $\mathbf{0.038}$ \\
     & Almost Open & $1$ & $1$ & $0.129$ \\
    \midrule
    CartPole & Conservative & $1$ & $1$ & $1$ \\
     & Guarded & $1$ & $\mathbf{0.009}$ & $1$ \\
     & Moderate & $1$ & $\mathbf{0.011}$ & $0.076$ \\
     & Balanced & $1$ & $0.091$ & $1$ \\
     & High & $1$ & $1$ & $1$ \\
     & Almost Open & $1$ & $1$ & $1$ \\
    \midrule
    AUV & Conservative & $1$ & $1$ & $1$ \\
     & Guarded & $1$ & $1$ & $\mathbf{3.56\times10^{-16}}$ \\
     & Moderate & $1$ & $1$ & $\mathbf{1.69\times10^{-15}}$ \\
     & Balanced & $1$ & $1$ & $\mathbf{2.87\times10^{-14}}$ \\
     & High & $1$ & $1$ & $\mathbf{7.90\times10^{-9}}$ \\
     & Almost Open & $1$ & $1$ & $\mathbf{4.70\times10^{-5}}$ \\
    \midrule
    Robot & Conservative & $1$ & $1$ & $1$ \\
     & Guarded & $1$ & $1$ & $1$ \\
     & Moderate & $1$ & $1$ & $1$ \\
     & Balanced & $1$ & $1$ & $1$ \\
     & High & $1$ & $1$ & $1$ \\
     & Almost Open & $1$ & $1$ & $1$ \\
    \bottomrule
  \end{tabular}
\end{table}

  \setcounter{table}{11}

  \FloatBarrier

\section*{Conflict of Interest}
The authors declare no conflict of interest.

\section*{Data Availability}

All simulation code required to reproduce the experimental runs and
the resulting run data are available in the project repository: \calfwrepo.

\section*{Acknowledgment}
The illustrations of the Pendulum and CartPole Swing-Up environments in
\Cref{fig:gym_env_pendulum,fig:gym_env_cartpole} are based on
the official Gymnasium renderings of these environments and were
redrawn with an image-generation model accessed through OpenAI Codex.
AI coding assistants (Claude by Anthropic and Codex by
OpenAI) were used
to improve the visual rendering of the simulations of the
Contaminated-Zone AUV Navigation and Treasure-Collecting Robot
environments shown in \Cref{fig:envunderwaterdrone,fig:treasure_robot},
to refactor and document the code in the project repository, and to
prepare it as a ready-to-use reproduction package.
The CALF-Wrapper algorithm, its original implementation, the
theoretical results, and the experimental design were developed by the
authors, and AI assistance was limited to polishing the existing code,
refining the presentation of the derivations, and improving the
exposition.
AI tools were also used to survey and verify the related literature,
for language editing and grammar checking, and to verify
bibliographic details.
The authors reviewed all AI-assisted changes and take full
responsibility for the content of this article.

\bibliographystyle{IEEEtran}
\bibliography{bib/CALFW,bib/calf-enhance}

\end{document}